\documentclass{article}

    \PassOptionsToPackage{numbers,sort&compress}{natbib}

\usepackage[dandb, final]{Styles/neurips_2025}

\usepackage{caption}
\usepackage{subcaption}
\usepackage[utf8]{inputenc} %
\usepackage[T1]{fontenc}    %
\usepackage{hyperref}       %
\usepackage{url}            %
\usepackage{booktabs}       %
\usepackage{amsfonts}       %
\usepackage{nicefrac}       %
\usepackage{microtype}      %
\usepackage{xcolor}         %
\usepackage{graphicx}
\usepackage{svg}            %
\usepackage{multirow}
\title{BEDLAM2.0:\\ Synthetic Humans and Cameras in Motion}

\usepackage{epsfig}
\usepackage{graphicx}
\usepackage{amsmath}
\usepackage{amssymb}
\usepackage{xspace}

\author{%
Joachim Tesch$^{1}$
\quad Giorgio Becherini$^{1}$
\quad Prerana Achar$^{1}$
\quad Anastasios Yiannakidis$^{1}$\\
\textbf{
\quad Muhammed Kocabas$^{2}$
\quad Priyanka Patel$^{2}$
\quad Michael J.~Black$^{1}$}\\
 $^1$Max Planck Institute for Intelligent Systems, T{\"u}bingen, Germany\\
 $^2$Meshcapade GmbH
}

\newcommand{\btwo}{BEDLAM2.0\xspace}
\newcommand{\bedlam}{BEDLAM\xspace}
\newcommand{\supmat}{Appendix\xspace}

\begin{document}

\maketitle

\begin{abstract}
Inferring 3D human motion from video remains a challenging problem with many applications.
While traditional methods estimate the human in image coordinates, many applications require human motion to be estimated in world coordinates.
This is particularly challenging when there is both human and camera motion.
Progress on this topic has been limited by the lack of rich video data with ground truth human and camera movement.
We address this with \btwo, a new dataset that goes beyond the popular
\bedlam dataset in important ways.
In addition to introducing more diverse and realistic cameras and
camera motions, \btwo increases diversity and realism of body shape,
motions, clothing, hair, and 3D environments. Additionally, it adds
shoes, which were missing in \bedlam.
\bedlam has become a key resource for training 3D human pose and
motion regressors today and we show that \btwo is significantly
better, particularly for training methods that estimate humans in
world coordinates.
We compare state-of-the art methods trained on
\bedlam and \btwo, and find that \btwo significantly improves accuracy over \bedlam. 
For research purposes, we provide the rendered videos, ground truth body parameters, and camera motions. We also provide the 3D assets to which we have rights and links to those from third parties.

\end{abstract}

\section{Introduction}

The BEDLAM dataset \cite{bedlam} was the first synthetic dataset of 3D
human motion of sufficient realism and complexity that synthetic data
alone could be used to train a state-of-the-art (SOTA) method for
estimating 3D human shape and pose (HPS) from images.
Since its introduction, BEDLAM has become a standard dataset for supervised training of HPS regression methods.
Despite this success, BEDLAM has several key limitations that hold back the field.
Key among these is that BEDLAM uses limited camera focal lengths and camera motions.
Here we address these limitations  and provide a significantly richer
dataset appropriate for end-to-end training of HPS methods that estimate humans in
world coordinates.
Beyond richer camera motions, the BEDLAM2.0 dataset addresses
several other important limitations that improve its diversity and
realism.

Beyond the camera motions, \btwo goes beyond B1 in the following ways:
\begin{itemize}
    \item We significantly expand the range of body shapes with more high-BMI
      bodies.  %
    \item We provide more varied and realistic 3D hair that is adapted to
      individual head shapes. %
    \item We add widely varied shoes, which are completely missing in BEDLAM. This includes defining the sole thickness, making foot-ground contact more realistic.
    \item We include more 3D clothing outfits and grade many outfits
      into standard sizes. In comparison to B1, this lets us dress
      diverse body shapes in realistic clothing.
     \item B2 also includes more 3D scenes, more complex human
       motions, and longer motions, increasing the diversity of the dataset.
\end{itemize}

The most significant upgrade involves the cameras.
Specifically, we define cameras that cover the range of focal lengths seen in real images and videos,  including dynamically changing focal lengths.
We also define several types of camera motions: panning, zooming, orbiting, tracking, etc. and add realistic motion noise to these.  This is similar to prior work \cite{wang2023zolly,pace2024kocabas,wang2024blade}.
We go further, however, to capture real camera motions using hand-held phone and tablet devices as well as an Apple Vision Pro headset for ego-centric camera motion captures. 
For these captures, we place 3D humans in a 3D scene and users move around the scene to view the virtual subject(s).  
This induces natural movements with realistic camera shake from both hand-held and ego-centric views.
The resulting dataset contains over 27K image sequences with over 8M frames, over 4K diverse body shapes, resulting in 13.3M bounding boxes.

To evaluate the dataset, we train several SOTA HPS regressors using \bedlam (B1) and \btwo (B2)
and evaluate their accuracy. 
Using B2 results in a significant improvement in accuracy compared with B1 across all standard metrics and the combination of B1 and B2 leads to SOTA performance on human pose estimation in world coordinates.

Similar to \bedlam, the released dataset includes the videos, ground truth body parameters in SMPL-X format \cite{SMPL-X:2019}, camera motions and focal lengths, clothing assets, 3D hair assets, and depth maps.
For assets that we cannot distribute, we provide links to the sources. 
As with \bedlam, the assets will enable others to render their own versions of the dataset for specific tasks like egocentric vision.
We also provide separate training and testing splits.
\btwo is available for research purposes.

\section{Related Work}

Prior to \bedlam, numerous synthetic datasets were proposed for training human pose and shape (HPS) regressors \cite{moulding,varol2017learning,Ionescu14PAMI,Mehta17TDV,cgf.13125,liang2019shape,JiangZHLLB20,liu2019temporally,pumarola20193dpeople,bazavan2021hspace,playingpami,Patel21CVPR}; see \cite{bedlam} for a review.
Due to limited realism and/or diversity, none of these are able to replace training data extracted from real images.
These methods also focus on human pose and not camera motion.
While \bedlam includes a few sequences with moving cameras,  the diversity of camera motions and focal lengths is limited and most of the sequences have static cameras.

While early work on HPS estimation focuses on estimating the 3D human pose in camera coordinates, many methods require bodies in world coordinates. 
Recent methods focus on this problem \cite{sun2023trace,pace2024kocabas,CameraHMR2025,wham:cvpr:2024,SLAHMR,promptHMR,wang2024tram,shen2024gvhmr} but are limited by the lack of training data with ground truth camera motions and 3D humans together.
\btwo is designed to support this direction so, here, we focus on synthetic datasets published since \bedlam that include varied cameras and camera motions.

Contemporaneous with \bedlam, SynBody \cite{yang2023synbody} is a synthetic dataset in which each sequence is rendered from 8 static views. While similar to \bedlam, it is less effective for training HPS regressors \cite{cai2023smplerx}.
Microsoft's SynthBody dataset \cite{hewitt2024look} uses SMPL-H \cite{mano} with a different head compared to SMPL-X and provides only static poses. They demonstrate how it can be used to detect dense 2D keypoints and they use these to fit a 3D body using optimization. They do not use the dataset to train a 3D regressor or show results on standard benchmarks.
STAGE \cite{stage} takes a different approach, using generative AI to take a 3D body and produce realistic images matching the body pose but with varied visual attributes such as BMI and clothing.
They do not use this for training a regressor but, rather, use this to generate benchmarks for evaluation.

PDHuman \cite{wang2023zolly} and BEDLAM-CC \cite{wang2024blade} tackle the problem of varied focal lengths, which correlate with the depth of the person from the camera; e.g.~long focal lengths are used for people far away and short ones for people close to the camera. 
They generate synthetic training images with widely varied views and focal lengths and show that training on such data improves robustness to real-world cameras. 
They do not, however, address camera motion.

Recent methods have introduced richer camera motions than those in \bedlam.
For example, EgoGen \cite{egogen} builds on the \bedlam assets and re-purposes them for tasks in egocentric vision.
They provide a generative process of an agent's motion in a 3D scene. This enables automated collection of video sequences from an egocentric viewpoint. 
The HumanVid dataset \cite{HumanVidSyn} also leverages \bedlam and adds camera motions using simple rules.
They sample multiple random camera locations in space at specific keyframes and point the camera at the subject's face. They then smoothly connect the cameras to create the camera motions.
WHAC-A-Mole \cite{yin2024whac} renders dancing sequences, with pairs of dancers, with varied camera motions including Arc, Pan, Push, Pull, and Tracking shots plus combinations of these.
Unfortunately, the synthetic data lacks realism.
PACE \cite{pace2024kocabas} renders 3D characters in scenes with a moving camera but the amount and diversity of the data is limited. Consequently, they use it only for evaluation of human and camera motion estimates.

\section{Dataset: Methods}

Here, we describe the key improvements of \btwo that increase its diversity and realism as compared with \bedlam.

\begin{figure}
    \centerline{
    \includegraphics[height=0.9in]
    {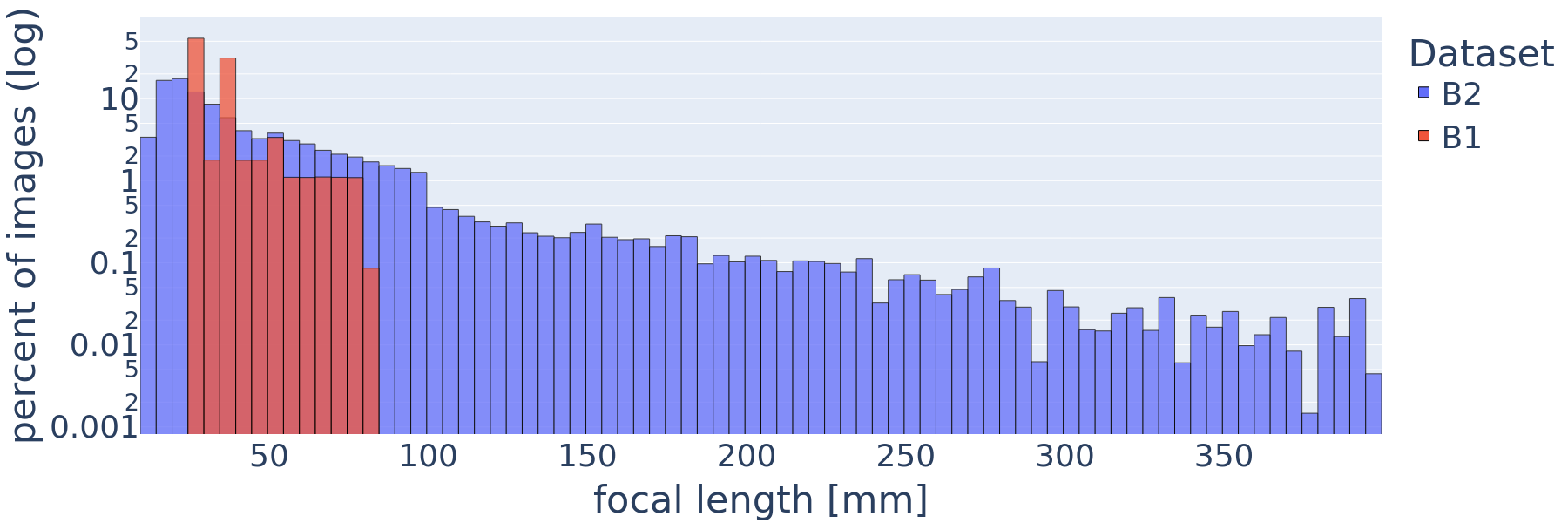}
    \includegraphics[height=0.9in]{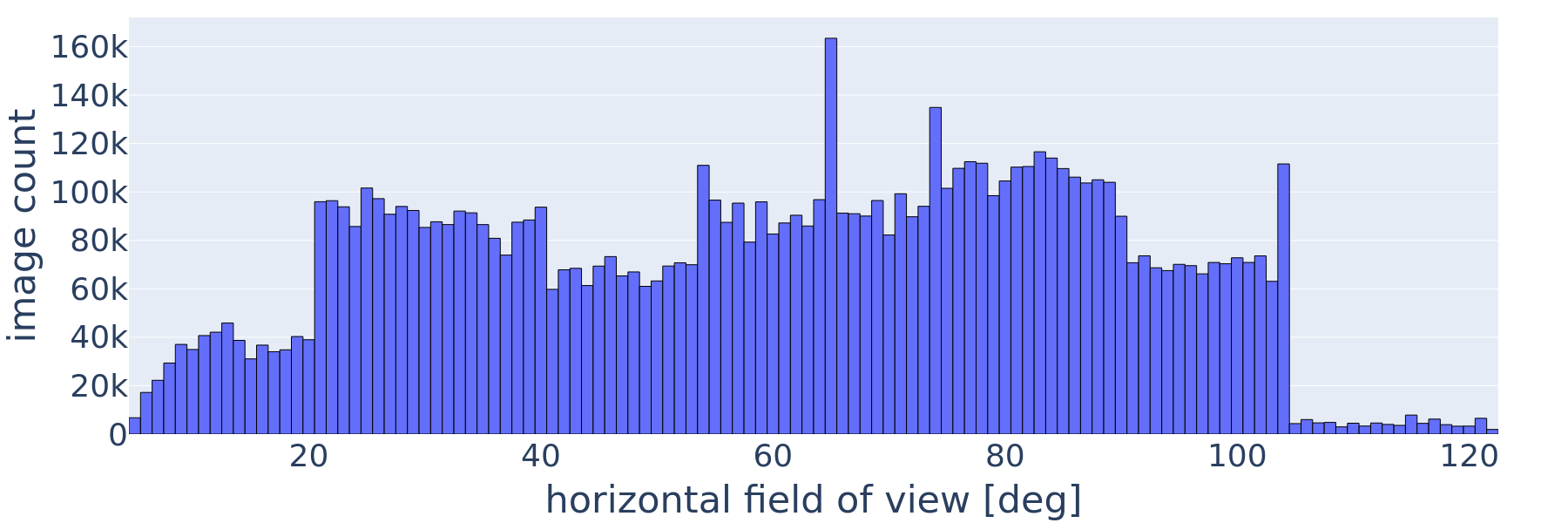}  
    }
 \caption{{\bf Camera intrinsics.} (left) Log frequency of focal lengths. Red: \bedlam; Blue \btwo. (right) Histogram of the horizontal field of view (HFOV).}
    \label{fig:cameraintrinsics}
\end{figure}

\begin{figure}
\centering
\begin{subfigure}{3.2in}
\centerline{
    \includegraphics[height=0.8in]{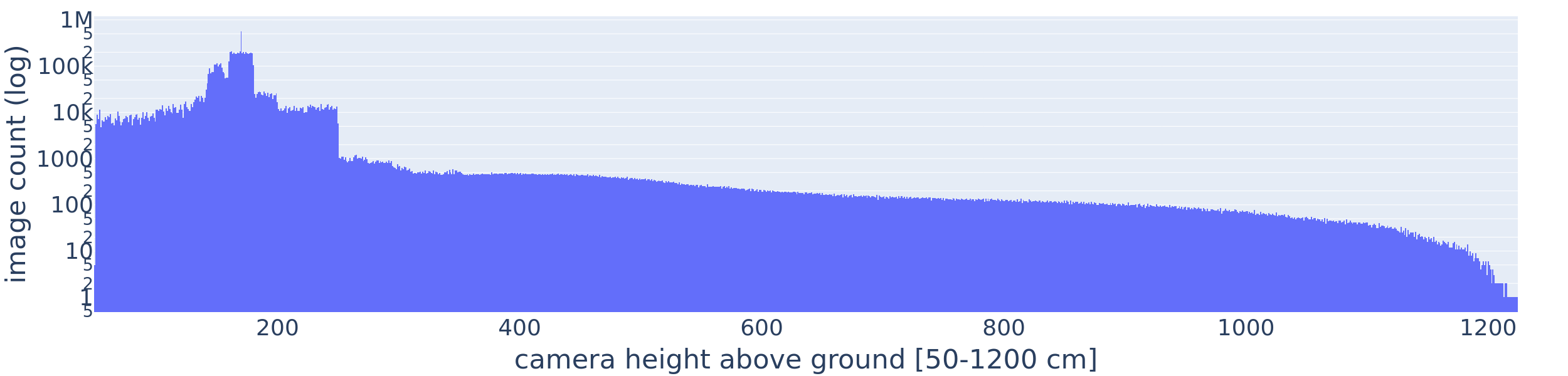}
}
 \centerline{
    \includegraphics[height=0.8in]{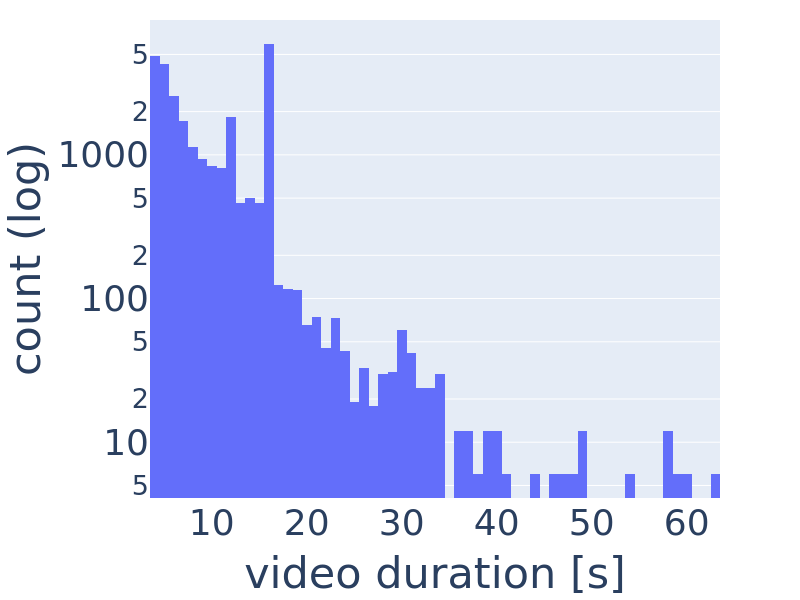}    
    \includegraphics[height=0.8in]{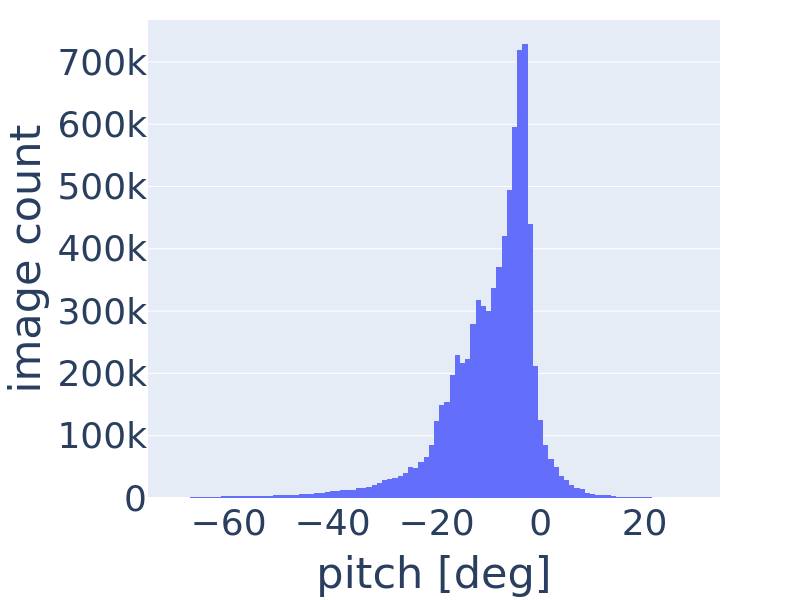}
    \includegraphics[height=0.8in]{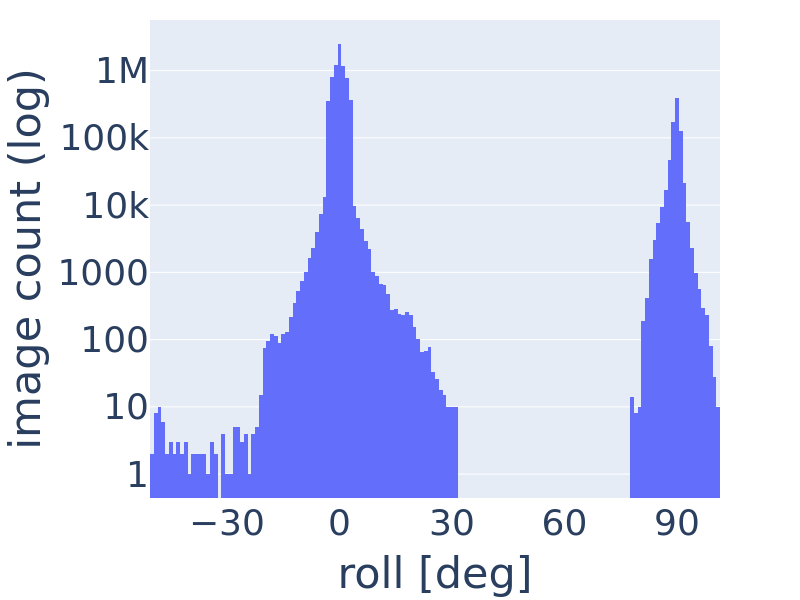}      
 }
\end{subfigure}
\quad
\begin{subfigure}{2in}
\centering 
    \includegraphics[height=1.5in]{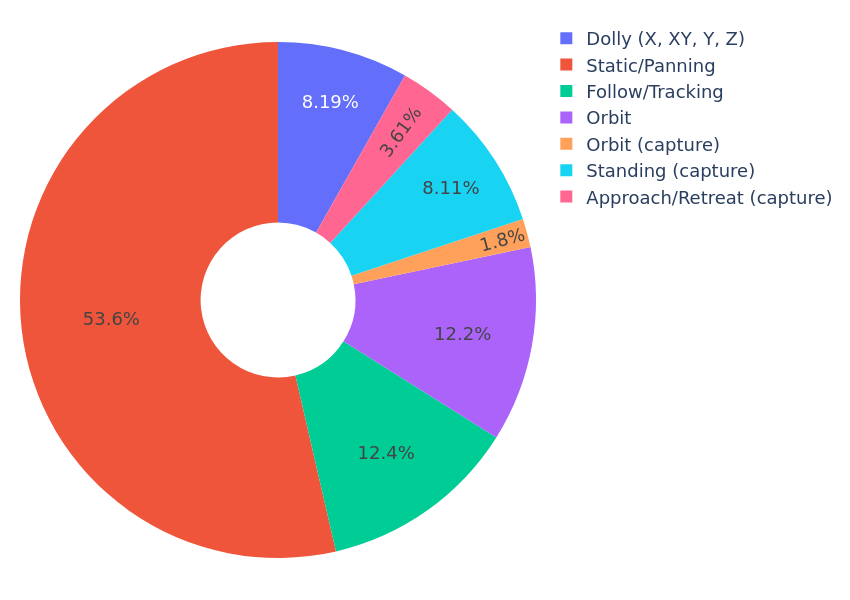}
\end{subfigure}
    \caption{{\bf Camera motion statistics.} (top) Log histogram of camera height above the ground (50-1200cm). (bottom left) Video duration (seconds). (bottom middle) Histogram of pitch. (bottom right) Log histogram of roll. (right) Distribution of different types of camera motion (see text).}
    \label{fig:cameramotionstats}
\end{figure}

\subsection{Cameras}

The recent focus of the field is on human pose and shape estimation in scenes with moving cameras. There is limited training data for such scenarios, which makes it hard to train models end-to-end on this task.
\btwo addresses this by significantly increasing the complexity and realism of the cameras and their motions.

\textbf{Focal lengths.}
\bedlam intrinsics primarily cover a small Horizontal Field Of View (HFOV) range of $52^\circ$ or $65^\circ$ and are mostly fixed during a shot. Temporal variation of camera extrinsics is very limited since most sequences use a static camera.
We address this in \btwo with a much broader focal length coverage.
Figure \ref{fig:cameraintrinsics} shows the distribution of HFOVs.
Specifically, we cover focal lengths from 14mm up to 400mm on a 16:9 DSLR sensor (36 x 20.25mm) that is designed to mimic real-world focal lengths.
This is much more diverse than the statistics of Flickr images reported in \cite{CameraHMR2025}; see \supmat Fig.~\ref{fig:hfov_plot_comp} for details and a comparison with \cite{CameraHMR2025}.

Nine percent of all generated videos have varying focal length during the shot by zooming in or out. This is important for realism as it mimics many real-world videos.
Start and end focal length values are randomized from a predefined configuration range suitable for the desired location shot and are then keyframed in Unreal Sequencer using Unreal Python automation. Indoor environment shots typically have short focal lengths, whereas long focal lengths are primarily used in outdoor environments.

\begin{figure}[t]
    \centering
        \includegraphics[trim=0 0 0 150, clip, height=1.5in]{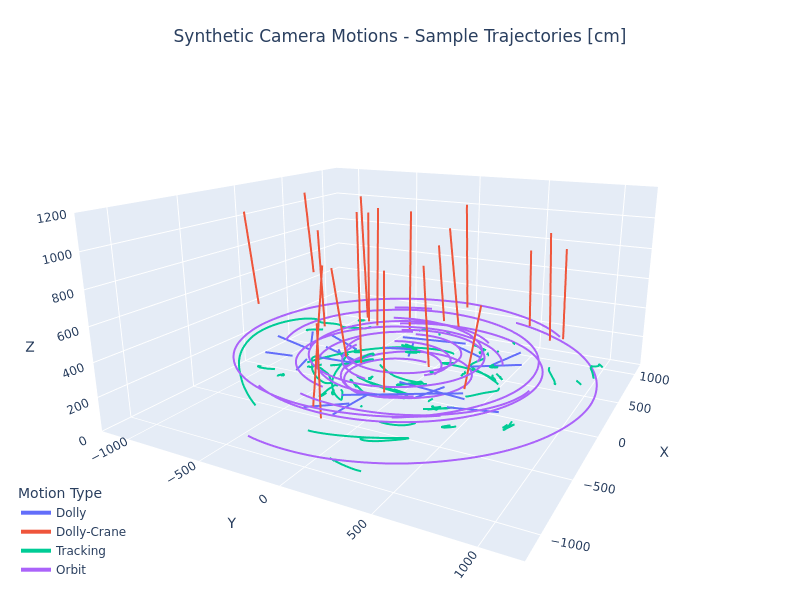}
        \includegraphics[trim=0 0 0 150, clip, height=1.5in]{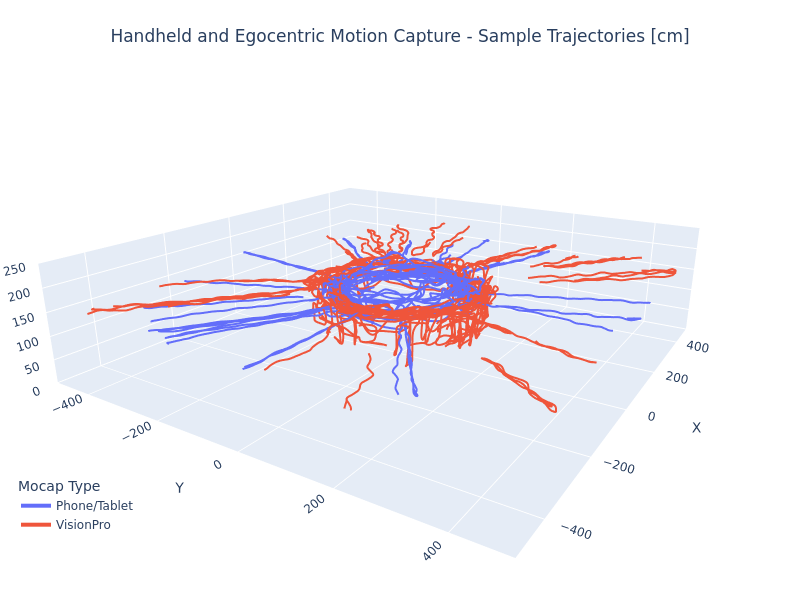}
    \caption{{\bf Sample camera motions used in dataset.} Left: Synthetic, Right: Captured.}
    \label{fig:cameramotions}
  \end{figure}

\textbf{Synthetic camera motions.}
\btwo introduces a variety of auto-generated synthetic camera motions, including static, panning, tracking, dolly, orbit, and zoom, and various combinations of these; see Fig.~\ref{fig:cameramotions} (left) and the \supmat for details.
To maximize variation in camera pose we optionally layer differentiable synthetic Perlin-noise camera shake for location and rotation on top of all these motions. The intensity of this shake effect is randomized.
We have options to track individual randomized body parts like the pelvis, spine or head with additional height offset randomization. For shot types that track moving bodies, the changes in extrinsics are keyframed in Unreal Sequencer, similar to the focal length setup. For tracking shots we utilize a custom Unreal Blueprint that uses a camera setup with the Unreal SpringArm component for smooth camera motions via low-pass filtering of the target location. For panning shots we use the default LookAt low-pass rotation filter feature of the Unreal cinematic camera. When low-pass filtering position or rotation we make sure to properly initialize the camera pose to the correct start pose for the first rendered frame before activating the filter. For the majority of shots the camera height above ground is between 0.5m and 2.5m with the exception of crane shots (dolly up/down) where the camera can move up to 12m above ground.
This gives top-down views, which are often missing from existing real-world datasets, increasing diversity.

\textbf{Captured camera motions.}
To add more realism and diversity, we capture real-world camera motions.
These include hand-held camera motions, which we capture with phone and tablet devices as well as egocentric camera motion data captured with an Apple Vision Pro headset.
The user views a scene containing 3D humans and we capture three types of camera movements: static location shots of a user standing in same location, orbit shots, and approach/retreat shots.
See  the \supmat for details.

Before rendering, we optionally randomize both handheld and egocentric camera motion data with additional height and distance offsets, pitch offsets, as well as world-space rotation offsets for viewpoint randomization.
Figure \ref{fig:cameramotions} (right) illustrates a few of these captured motions.

\textbf{Summary.} Together, the synthetic and real camera motions cover a broad range of movement types (Fig.~\ref{fig:cameramotionstats} (right)) with a diversity of camera pitches and heights (Fig.~\ref{fig:cameramotionstats} (left)).
\btwo also contains many longer sequences than B1 (Fig.~\ref{fig:cameramotionstats} (bottom left)). 
In the final dataset, 86.4\% of the motions are synthetic, while 13.6\% are captured.
See the \supmat for further details and \url{https://b2dash.is.tuebingen.mpg.de/} for detailed statistics.

\subsection{Human motions}

Our motion pool is composed of 4643 motions in SMPL-X format, as compared to 2311 in \bedlam.
The pool includes diverse motions sampled from AMASS \cite{AMASS}, in particular from the datasets CMU \cite{AMASS_CMU}, KIT \cite{AMASS_KIT-CNRS-EKUT-WEIZMANN}, BMLmovi \cite{AMASS_BMLmovi}, BMLrub \cite{AMASS_BMLrub}, HDM05 \cite{AMASS_HDM05}, ACCAD \cite{AMASS_ACCAD}, Transitions \cite{AMASS}, MoSh \cite{AMASS_MoSh}, SOMA \cite{AMASS_SOMA}, PosePrior \cite{AMASS_PosePrior} and DFaust \cite{AMASS_DFaust}. We go beyond \bedlam to sample additional motions from the MOYO dataset \cite{AMASS_MOYO}, which contributes complex yoga movements, and the BEAT2 dataset \cite{emage:cvpr:2024}, which captures conversational gestures.

\textbf{Preprocessing.} The motions are filtered to exclude actions where the balance of the body depends on an external object, such as sitting, or where the motion is not supported by a ground plane, such as going up and down stairs. Since many mocap motions start and/or end with a T-pose, we automatically identify the frames where the body is in T-pose and exclude these segments from sampling. Furthermore, we subsample the motions to 30 fps and add an offset to the translation parameters to ensure that the body is centered at the origin during the first frame of the motion.

\textbf{Sampling.} From the motion pool, we sample motion segments ranging from a minimum of 4 seconds to a maximum of 16 seconds in length, prioritizing 16-second long segments when available and shortening only when the source motion duration is not long enough. 
For comparison, the maximum length in \bedlam is 8 seconds.
From this sampling process we obtain a total of 10592 motion segments and 3,231,846 pose frames, of which 1,665,448 (51.53\%) appear only once in the dataset.

\textbf{Retargeting.}
To augment the animation data with various body shapes, we designed an automated pipeline to retarget the samples from our motion pool to the sampled body shapes. 
This is necessary because the sampled bodies may have different limb lengths than the original motion-capture subjects.
We use the IK-Retargeter tool from Unreal Engine, which retargets a source motion to a target skeleton, minimizing foot-sliding by world-space pelvis location adjustments for the new limb lengths.
This significantly increases the shape-motion diversity compared to \bedlam. Moreover, by adjusting body poses to fit different skeletons, the retargeting increases pose variety.
The code is available on the project website.

\textbf{Hand motion.} Due to the complexity of capturing hands with optical motion capture technology, a large fraction of AMASS does not contain hand motion. In order to provide hand pose variation, we augment the data by adding randomly sampled hand motions from the ARCTIC dataset \cite{fan2023arctic} to the existing AMASS motions.

\begin{figure}[t]
    \centering
        \includegraphics[height=1.25in]{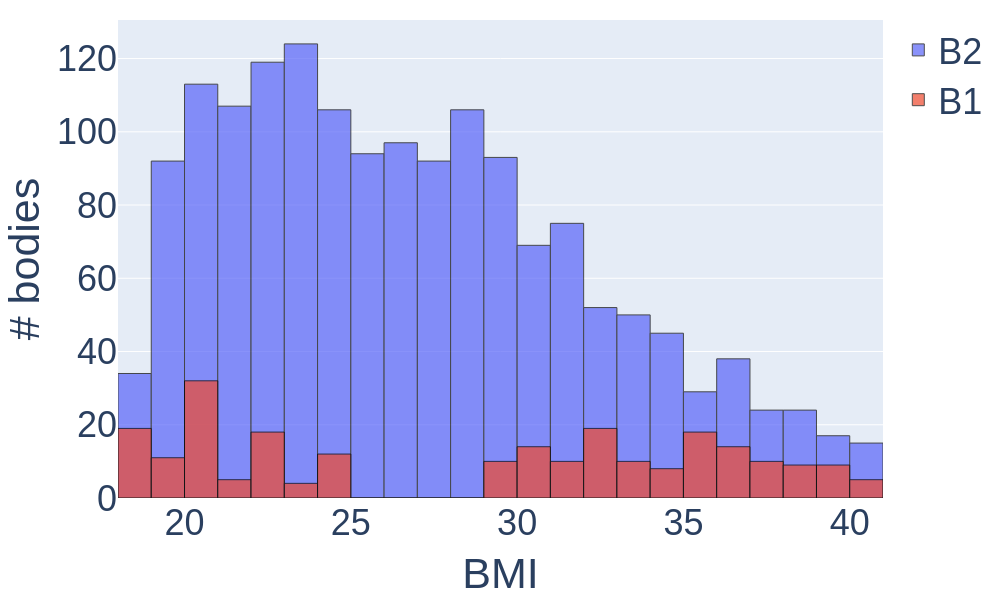}
        \includegraphics[height=1.25in]{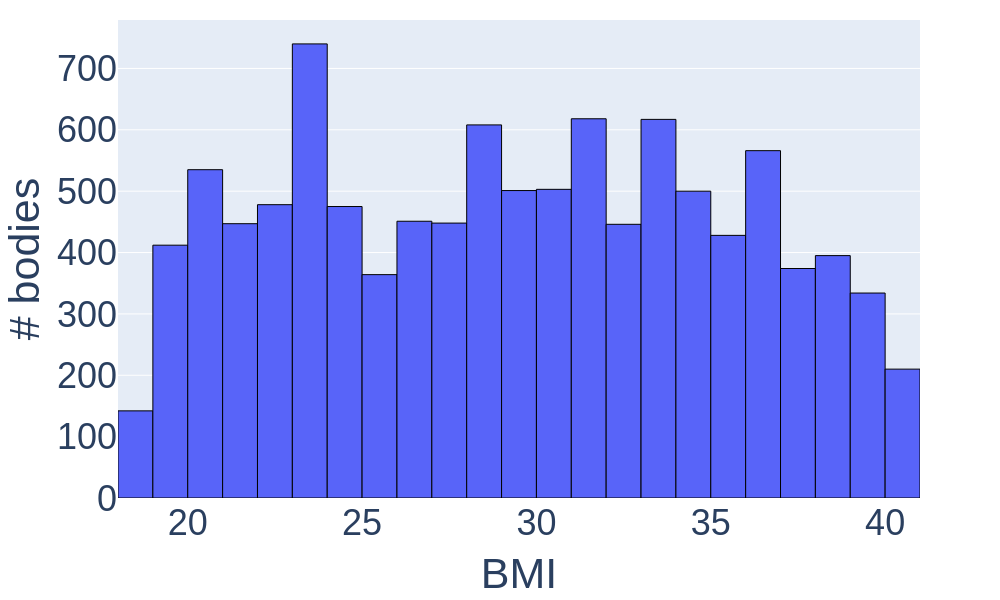}
        \caption{{\bf Body shapes.} Left: Histogram of BMIs for the body shapes in \bedlam (red) and \btwo (blue).
          Right: Resampled histogram used to generate more diverse bodies in \btwo.}
        \label{fig:bmi}
  \end{figure}
  
\subsection{Body shape}
Body shapes in \bedlam are not as diverse as in the real population. Here we increase the body-shape diversity, particularly for high-BMI bodies.
To that end, we sample SMPL-X \cite{SMPL-X:2019} with 16 shape parameters in accordance with the body mass indices (BMIs) in the CAESAR dataset \cite{CAESAR}, which includes a diverse range of male and female body shapes.
Specifically, we sample 1,615 bodies with BMIs ranging from 18 to 41, ensuring balanced representation across the entire BMI spectrum, as shown in Fig.~\ref{fig:bmi} (left).  
Note that the BMIs in CAESAR are skewed to BMIs under 30.
To provide more shape diversity, we resample this distribution to include more bodies with high BMIs as depicted in Fig.~\ref{fig:bmi} (right).

Note that B1 uses a different version of SMPL-X from B2. For B2, we use the version with a ``locked'' head, which removes the hair ``bun'' from the shape space. This is needed for realistic hair groom generation. B1 also uses fewer body shape components (11 vs 16).
To enable training and comparison using both datasets, we refit the B1 ground-truth using the B2 16-beta model. The resulting motion files are available for download from the original BEDLAM website (\url{https://bedlam.is.tuebingen.mpg.de/}).

\subsection{Hair}

Hair realism and variation in \bedlam is limited by the card-based hair models used and most subjects do not have hair. Additionally, the hair assets have a license that does not allow redistribution.
To address these issues \btwo uses higher quality strand-based 3D hair grooms.
This approach models the hair as individual hair strand 3D curves, allowing us to adapt each hair groom to the individual body head shapes. This also improves render quality and results in more accurate hair rendering under HDR image-based lighting conditions with raytraced shadows. Unlike \bedlam we use hair in all rendered sequences.

\begin{figure}[t]
    \centering
    \includegraphics[height=1.0in]{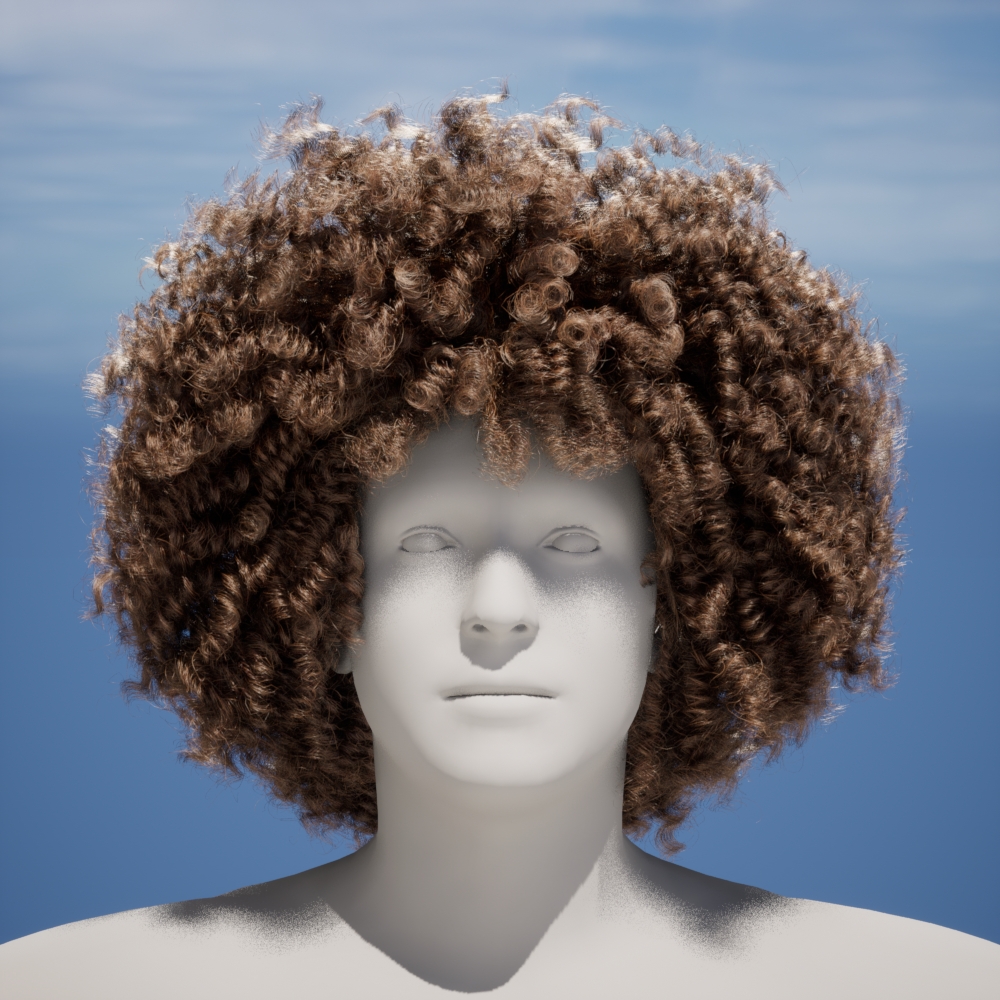}
    \includegraphics[height=1.0in]{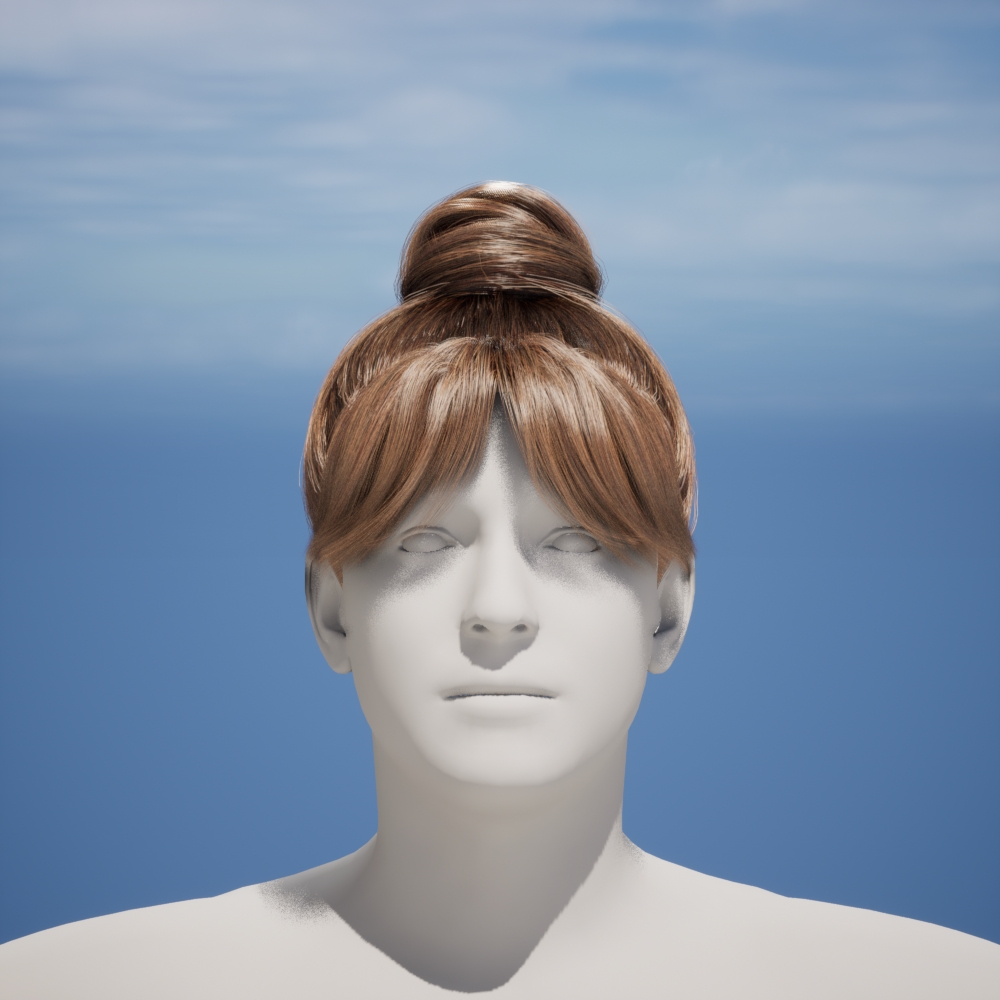}
    \includegraphics[height=1.0in]{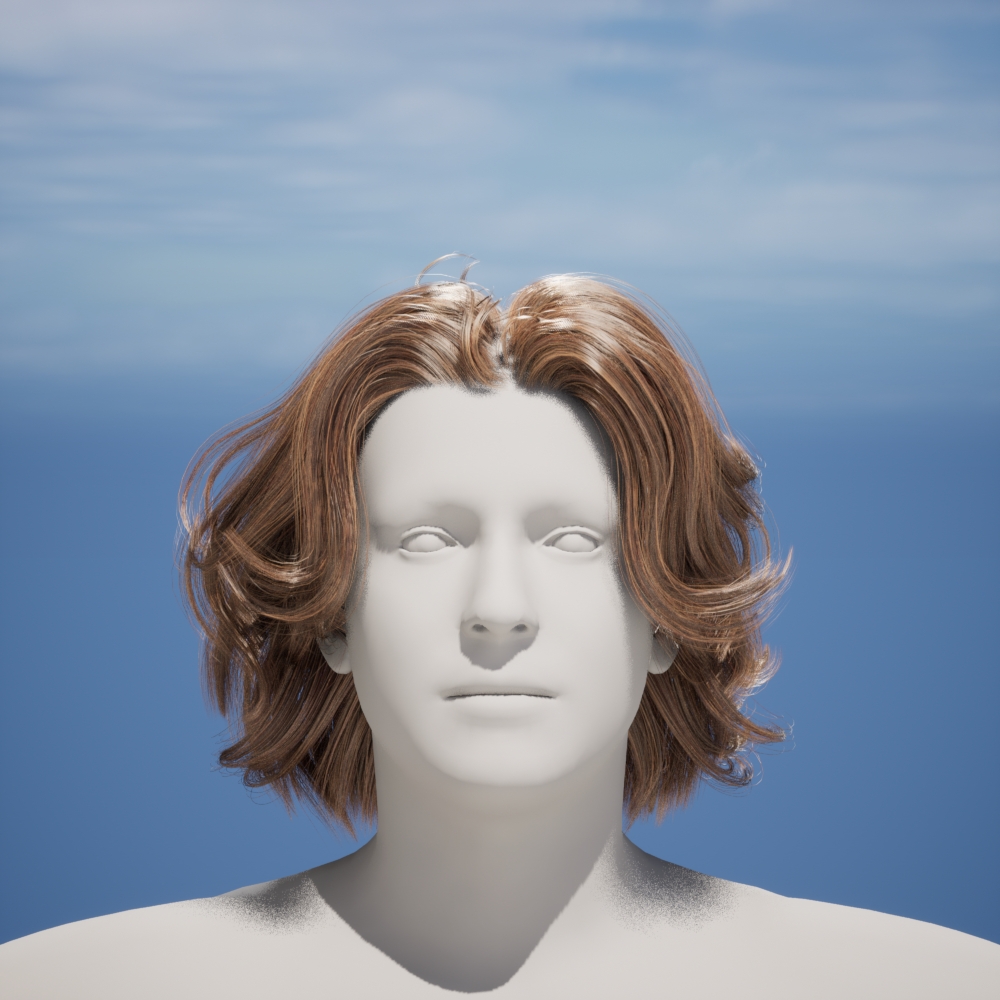}
    \includegraphics[height=1.0in]{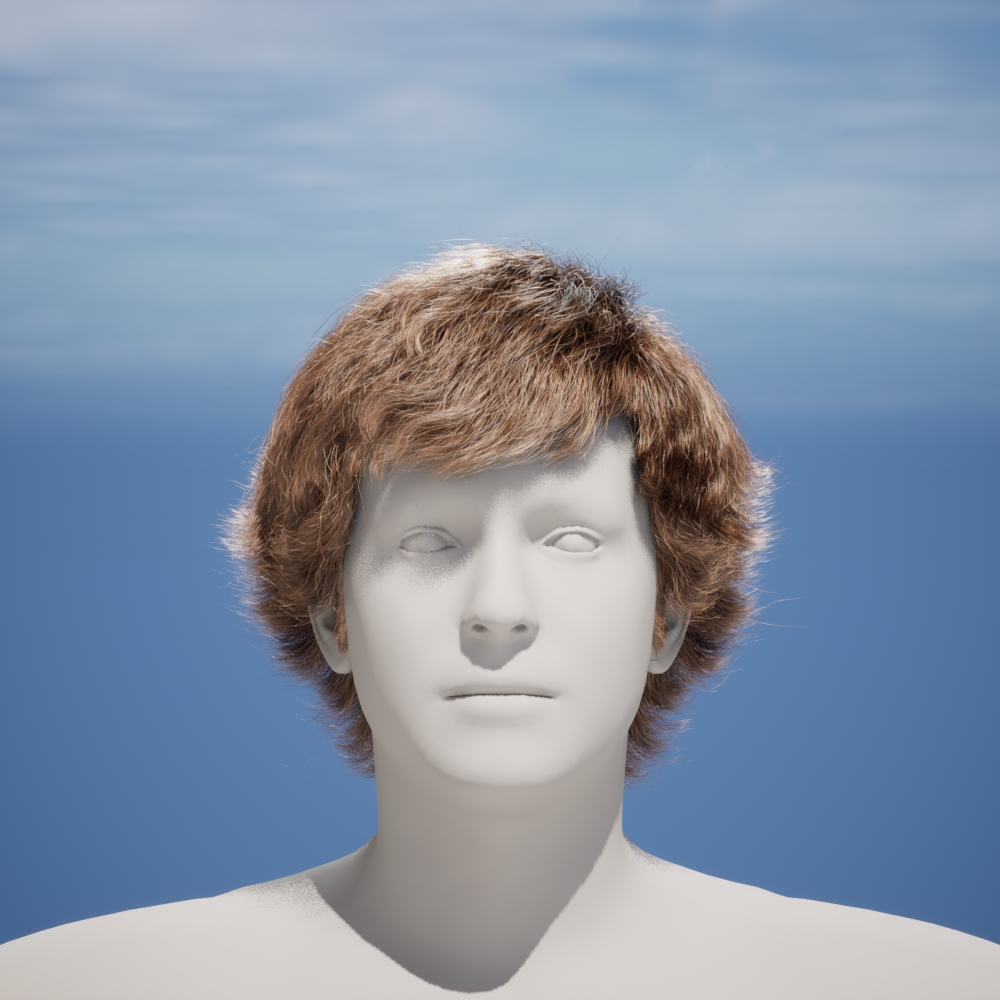}
    \includegraphics[height=1.0in]{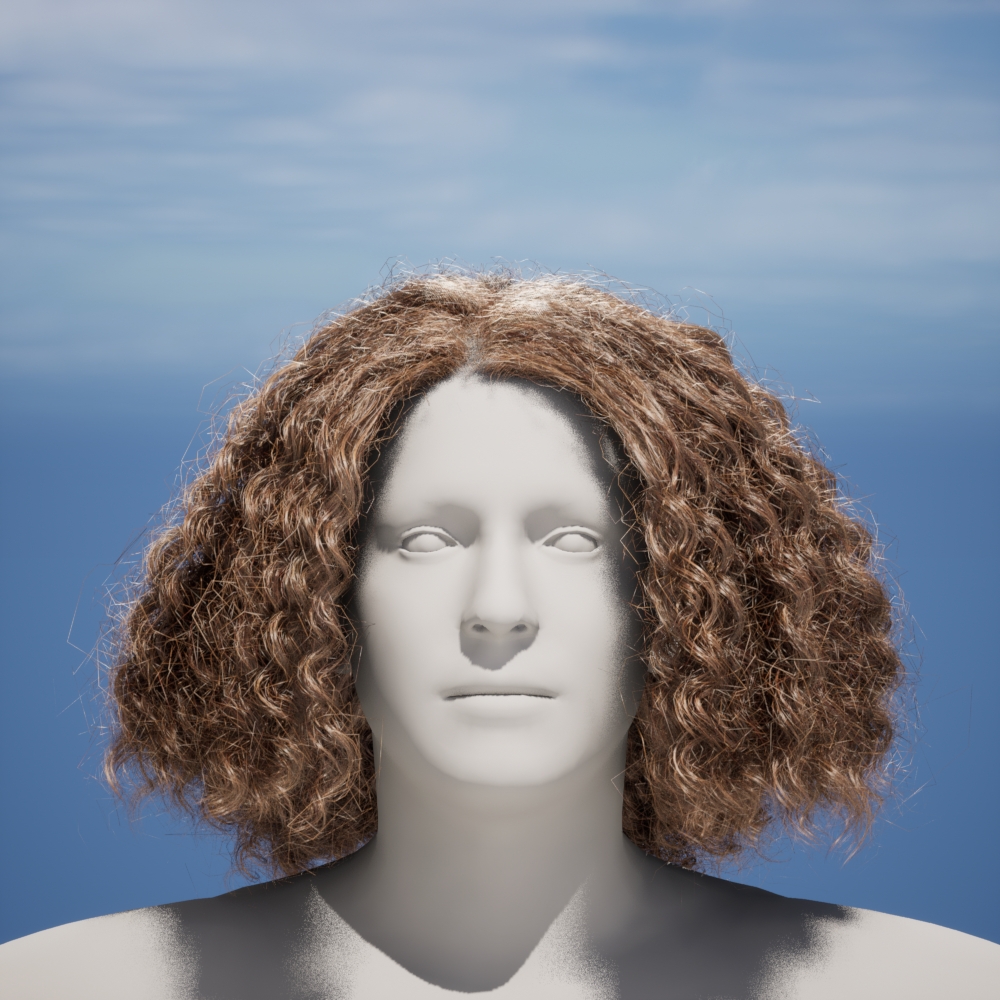}
    \caption{\textbf{Strand-based hair.} These examples illustrate the realism and diversity, which is much better than in \bedlam. }
    \label{fig:hair}
\end{figure}

We contracted a professional VFX studio with previous experience in generating realistic grooms for synthetic data generation \cite{wood2021fake} to create 40 unique hairstyles for our SMPL-X (AMASS, no head bun) neutral mesh default head shape (see Fig.~\ref{fig:hair}).
Each groom has between 50k and 100k 3D strand curves of varying length. Total vertex count for a groom starts at 1 million for straight hair and can go up to 13 million for complex curly hair styles like afro, where each strand contains about 170 vertices. Unlike previous work in this area, we release all of these grooms for non-commercial use.
In Unreal, we apply a randomized hair groom to the selected target body and use the hair binding component to adapt it automatically to the target headshape surface. Hair color is determined by a dedicated hair shader, which uses a combination of melanin and redness values to define the 9 hair material presets used for rendering.

\subsection{Shoes}
\bedlam, and similar datasets, contain SMPL or SMPL-X bodies with bare feet.
This creates a domain gap to real imagery in which people typically wear shoes.
This further creates an issue for estimating the body height and foot-ground contact because shoes introduce a displacement between the bottom of the SMPL-X foot and the ground.

\begin{figure}[t]
    \centering
    \includegraphics[height=1.6in]{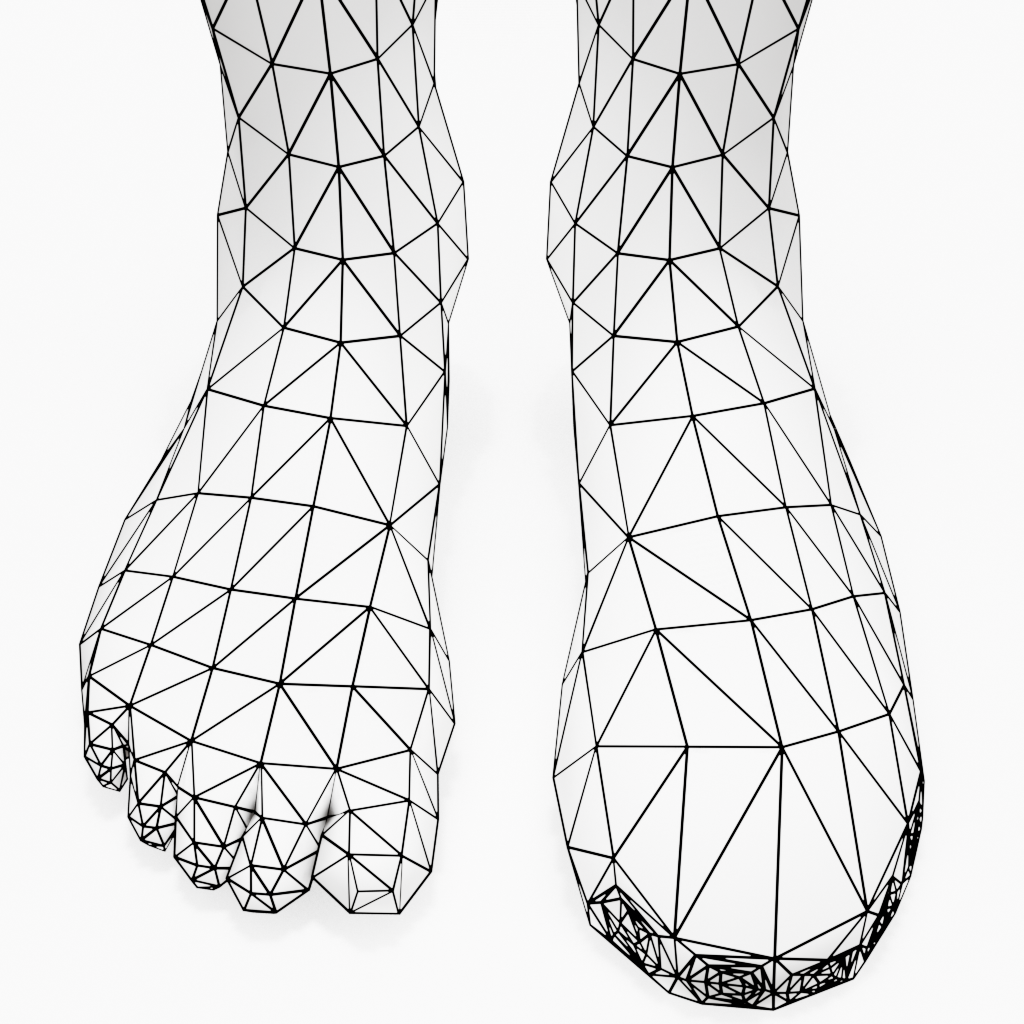}
    \includegraphics[height=1.6in]{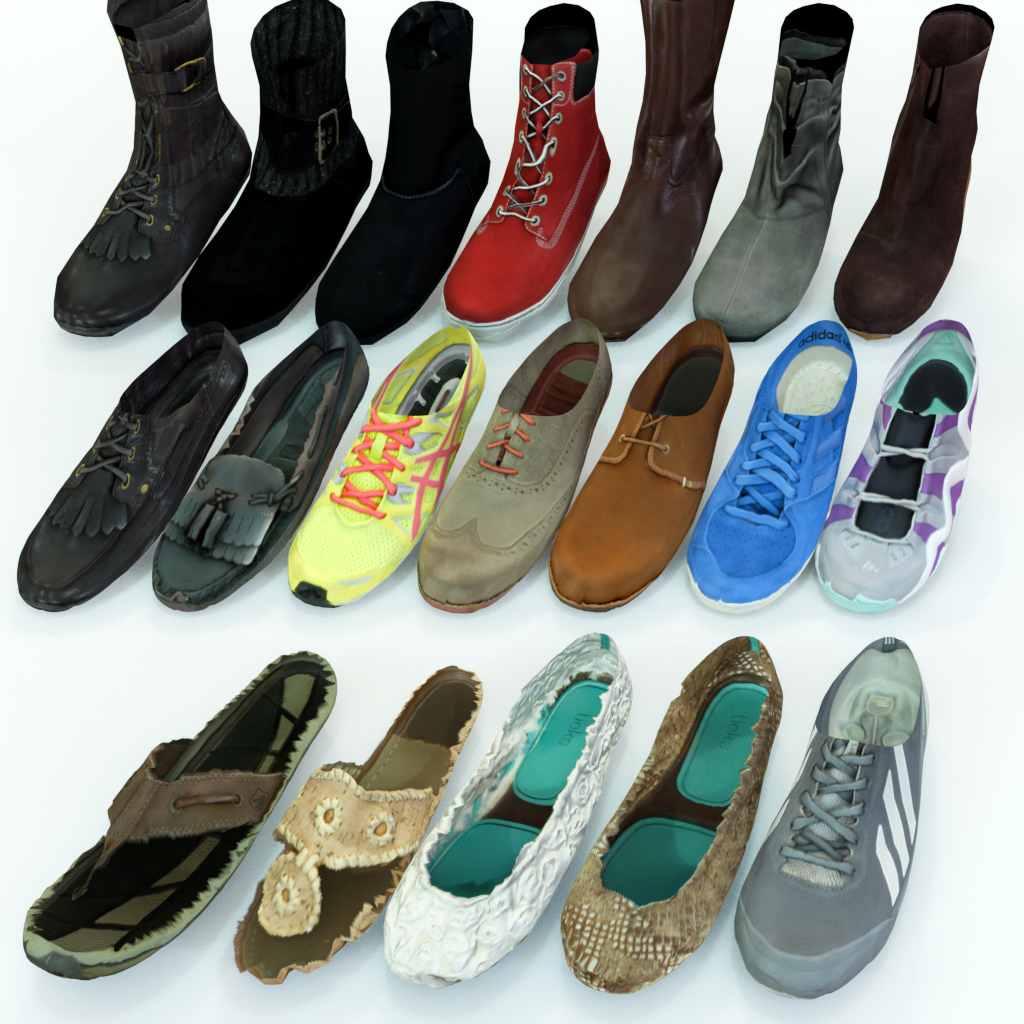}
    \includegraphics[height=1.6in]{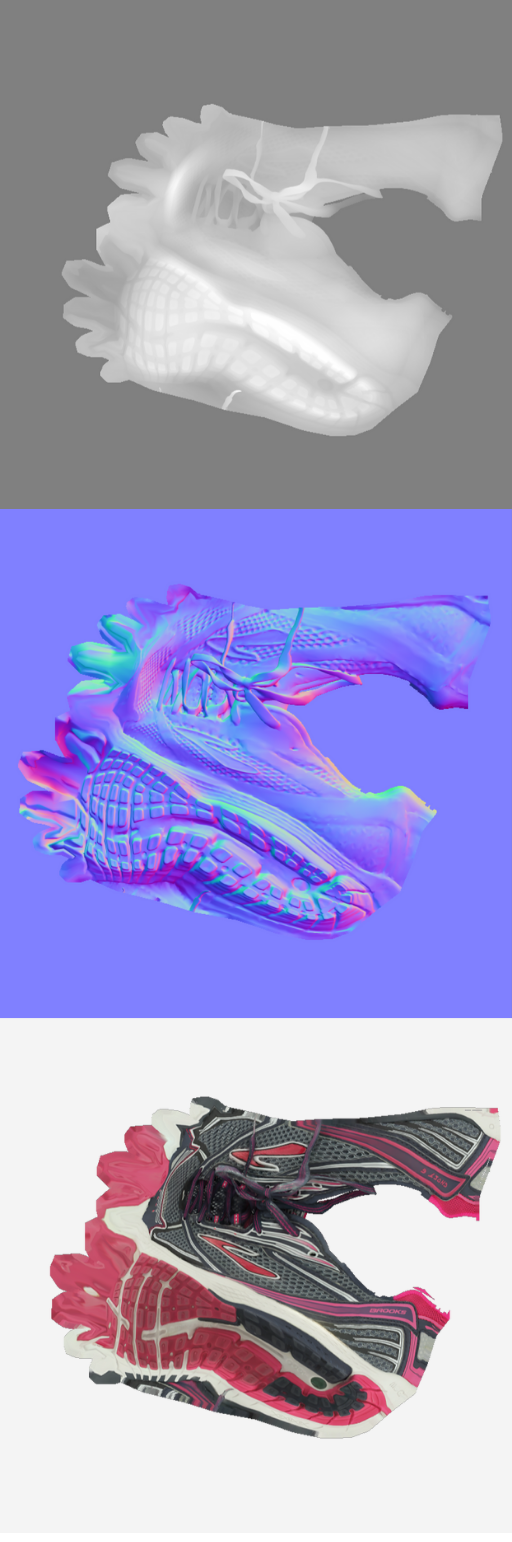}
    \includegraphics[height=1.6in]{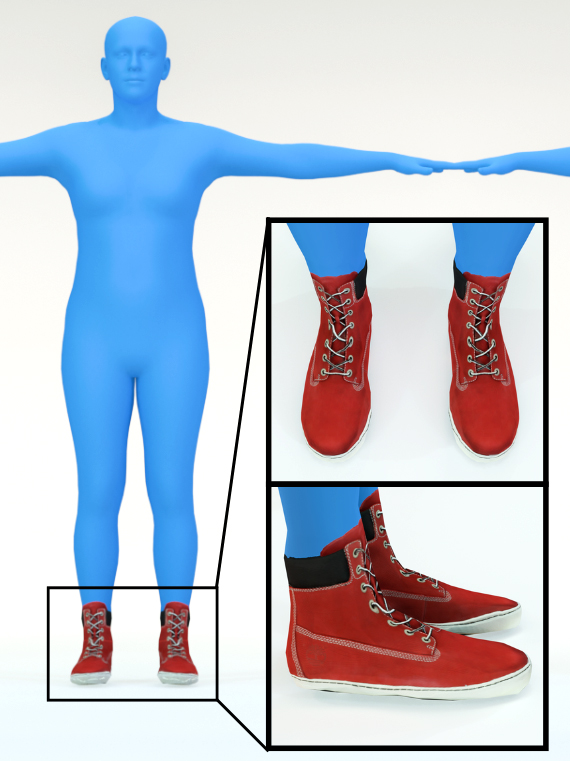}
    \centerline{\small\hspace{0.9in} a\hfill b\hfill c\hfill d\hspace{0.7in}}
    \caption{\textbf{SMPL-X Shoes.} (a) SMPL-X foot vs the ``sock'' foot. (b) Examples from the shoe database. (c) Displacement, normal and texture maps. (d) Shoes rendered on the SMPL-X body.}
    \label{fig:shoes}
\end{figure}

Adding shoes to SMPL-X is not trivial since the mesh topology represents the toes.
Consequently, as a first step, we smooth out the SMPL-X toes to create a smooth canonical sock-like foot that better matches the shape of shoes (Fig.~\ref{fig:shoes}a). 

Next, we source shoes from the Google Scanned Objects dataset \cite{googlescannedobjects},
which contains a wide variety of shoes and textures (see Fig.~\ref{fig:shoes}b for examples).
We use a subset containing 45 loafers, 6 formal shoes, 9 ballerina flats, 3 flip-flops, 18 boots, %
5 football shoes (with traction studded soles), while the remaining 96 are casual sport shoes.

We align all the shoe meshes, scaling them to create a common shoe size corresponding to the neutral SMPL-X mesh. 
We then align the shoes with the SMPL-X mesh and, for each pair of shoes, bake normal and texture maps to the SMPL-X UV space and compute a displacement map (Fig.~\ref{fig:shoes}c) that, when applied to the sock-like foot, deforms its shape to match the shape of the shoe. Subsequently, we add an appropriate upwards translation to the whole body to account for the sole thickness. 
See the \supmat for further details.

This approach provides the illusion of a shoe (Fig.~\ref{fig:shoes}d) while maintaining compatibility with AMASS motions \cite{AMASS} and leverages the SMPL-X shape space to handle shoe-size changes according to body shape. We only consider flat-soled shoes because heels require a significant change in foot topology and would influence posture and movement; this is future work.

\begin{figure}[t]
    \centering
    \includegraphics[height=1.2in]{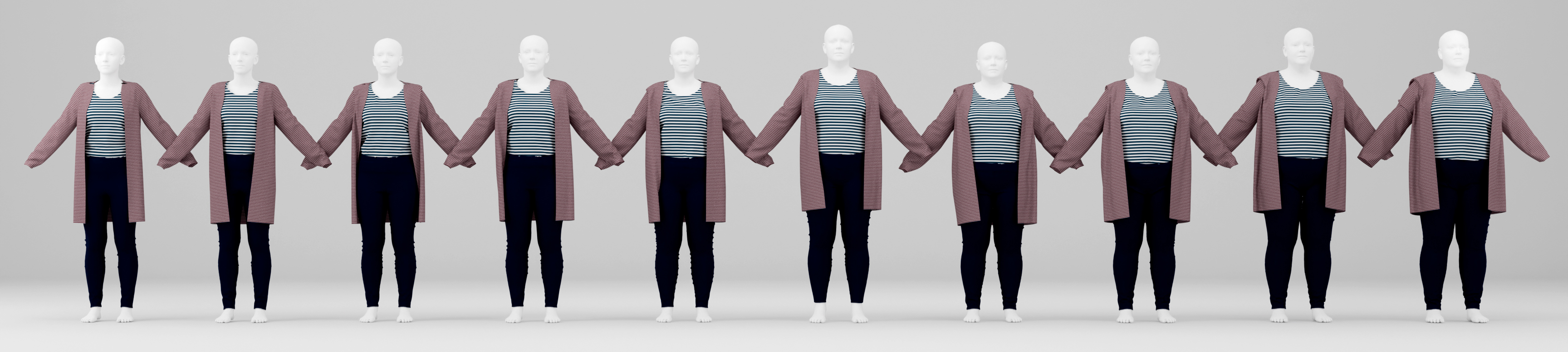}
    \caption{{\bf Graded garments.} Example of a graded outfit draped on bodies of varying BMI.}
    \label{fig:grading}
\end{figure}

\subsection{Clothing}

\bedlam animates bodies in 111 high-quality, detailed, 3D outfits.
Unfortunately, the outfits come in only one size and do not fit high-BMI bodies.
Consequently, all large bodies in \bedlam are only rendered with clothing textures on the naked body, limiting realism and making it hard to train methods that estimate body shape under clothing.

We go beyond \bedlam to add 76 new 3D outfits (187 in total) created by a professional 3D clothing designer using CLO \cite{clo}. The dataset contains a wide variety of outfits, from simple dresses to complex multi-layered outfits comprised of multiple garments, such as a man's suit.

We had the designer grade 50 of the outfits into sizes XS, S, M, L, XL, 2XL, 3XL, 4XL, 5XL, and 6XL.
We match each outfit size with a reference BMI value, and assign an outfit size to each avatar based on the BMI of the body.
This allows us to dress, animate, and render bodies of all sizes in realistic clothing (Fig.~\ref{fig:grading}).

For each outfit, we define a set of texture patterns (from \bedlam) from which we sample during rendering. Each outfit has a minimum of 6 and a maximum of 28 texture variations, while the median number of texture variations per outfit is 10.
As in \bedlam, cloth simulation is done using CLO and manually checked for quality; see the \supmat for details.

\subsection{Scenes and lighting}

Similar to \bedlam we render the animated bodies in many different environments 
including both simple scenes with high-dynamic-range (HDR) image-based lighting and full 3D environments.

\textbf{HDRI.}
We use HDRI environments when the camera is primarily panning and not translating since there is no parallax in this case.
We randomly sample from 94 preselected HDR images with varying illumination from PolyHaven \cite{polyhaven} and use these as the sole light source to light the bodies. The equirectangular HDR image is re-projected onto a virtual half dome using the Unreal HDRIBackdrop component. To increase the illumination and background variation we randomize the orientation of the rendered subjects and cameras in world space. Raytraced shadows are used to provide realistic dynamic body shadows on the ground plane for HDR images taken in direct sunlight.

\textbf{3D environments.}
\bedlam uses only 5 3D environments, whereas \btwo has 15 high quality, geographically diverse, 3D environments (from Unreal Marketplace and Fab).
Only one 3D environment is shared between \bedlam and \btwo and the other 14 are new. The number of indoor environments increases from 1 in \bedlam to 9 here.
For each environment, we sample between 3 and 25  shot locations depending on the type of environment, creating much more background variation compared to \bedlam.
We also increase lighting variation by using time-of-day randomization for selected environments with properly calculated physically correct sun light direction. We also use  different lighting setups like daylight, sunset, overcast and night time when available in the 3D environment.

\subsection{Occlusion}

The dataset includes a significant amount of occlusion of the bodies, including frame occlusion, self-occlusion, person-person occlusion, and environmental/scene occlusion, not to mention occlusion of the body caused by clothing, hair and shoes.

We performed a quantitative analysis on a randomly selected set of 41.5k images (covering 58.6k rendered bodies) to examine frame, scene, and person-person occlusion. Our results show that 12.7\%\/ of the images exhibit more than 20\%\/ occlusion, and the top 10\%\/ most occluded bodies experience an average of 61.1\%\/ occlusion.
See Fig.~\ref{fig:occlusion} in the \supmat for examples.

\subsection{Rendering: Image and Depth data}

Data is rendered in Unreal Engine 5.3 on NVIDIA RTX3090/RTX4090 GPUs using the built-in Movie Render Pipeline plugin with deferred rasterizer in cinematic quality settings. Two separate render passes are used for image and depth data generation at a resolution of 1280x720 for a target video framerate of 30 frames per second.

\textbf{Image data.} To achieve realistic motion blur, which is important for work like \cite{chen2025_imageimu}, we use the default 180-degree shutter for motion blur and always render 7 separate temporal subframe images that are combined into the final output image. Rendered image data is saved in lossless compressed PNG and EXR formats. EXR output is used to store the correct camera pose used at render time for each frame in JSON format embedded in the EXR metadata. This approach ensures correct camera pose data when camera shake modifiers are used. We observed that Blueprint-based approaches for camera pose logging like the method used in \bedlam do not capture these additional shake modifications, which would lead to inaccurate camera ground truth data.

\textbf{Depth data.} We also generate ground-truth depth maps in a separate render pass using center subframe camera pose, resulting in non-blurred depth in EXR format with 16-bit float precision.
To ensure correct camera pose consistency between the image and depth pass, we created two custom Unreal Engine C++ plugins. The first plugin ensures that Perlin noise camera shake is deterministic between re-renders with controllable variability. We achieve this by extending the existing camera shake functionality with the option to externally specify the used seed for noise randomization. The second plugin fixes the existing behavior of the Unreal Movie Render Pipeline, which only stores the last subframe camera pose and not the desired center subframe camera pose as ground truth in the EXR metadata. When rendering fast camera motions with 7 temporal samples there are noticeable differences in camera pose between the last subframe and the center subframe.
We extend the functionality to log the camera pose for all subframes in EXR metadata with center subframe as the ground truth camera pose reference.
See the \supmat for more details.

\section{Dataset Statistics}

\btwo contains 27480 video image sequences with a resolution of 1280x720 at 30 frames per second resulting in total 8,048,411 PNG images. This results in 12.5M and 862K bounding boxes containing humans with ground truth SMPL-X parameters for training and test set, respectively.
These images are generated from 56,338,877 rendered temporal subframes for realistic motion blur in every image of the dataset. The average video length is 10s. We also provide compressed H.264 encoded MP4 videos for all sequences as well as overview images for first, middle and last image of all sequences.
For each image we provide world-space ground truth for camera extrinsics and intrinsics and ground truth 3D bodies and their appearance randomization parameters. 
For 44\% of the images we also provide depth images (16-bit float) and a corresponding center subframe render without motion blur in EXR multilayer format. All image renders are organized by camera motion type to facilitate selecting desired camera motion subsets.
See the \supmat for a numerical comparison with other datasets.
Also please see the project website (\url{https://bedlam2.is.tuebingen.mpg.de/}) and video (\url{https://youtu.be/ylyqHnwhpsY})
example sequences illustrating the dataset.

To construct the training and test splits, we held out (i) a subset of 161 body shapes, and (ii) a subset of 597 motions, ensuring that the test set contains exclusively unseen pose and shape parameters; we then render 1824 new sequences in 5 new environments, containing only test bodies and motions, resulting in 449061 test images.
Please refer to the \supmat for details on the data split and its usage.

\section{Experiments}

Since \bedlam is already widely adopted in the field, we focus on comparing B2 with B1.
We use standard metrics for camera-space methods (MPJPE, PA-MPJPE, and PVE); see \cite{bedlam}.
For methods that estimate bodies in world space, we use WA-MPJPE$_{100}$, W-MPJPE$_{100}$, RTE, Jitter, and Foot-Sliding; see \cite{pace2024kocabas,SLAHMR,wham:cvpr:2024} for definitions.
We evaluate on 3DPW \cite{vonMarcard18ECCV}, RICH \cite{Huang:CVPR:2022}, and EMDB \cite{emdb}. These are all real-image datasets with high-quality pseudo ground truth.

\textbf{Image-based methods.} 
For single-image methods the diversity of the training data (poses, scenes, cameras, etc.) is key to accuracy.
While camera motion, per se, is irrelevant for single-image methods, B2 still has a wider range of camera focal lengths and camera poses than B1.
We take the current most accurate single-frame method at time of writing, CameraHMR \cite{CameraHMR2025}, as representative of such methods.
As shown in Tab.~\ref{tab:hps} training on B2 alone produces significantly lower error than training on B1. 
Training on both B1 and B2 does not provide a clear advantage. Training on B2 also results in a 20\% improvement in shape accuracy compared to training on B1 (see Table \ref{tab:image_shape_accuracy} in the \supmat).
\begin{table*}
   \caption{Single-frame pose estimation using CameraHMR \cite{CameraHMR2025}. See text.}
   \centering 
\resizebox{\textwidth}{!}{
  \begin{tabular}{ll|ccc|ccc|ccc}
    \toprule
    & \multirow{2}{*}{Dataset} & \multicolumn{3}{c}{3DPW~\cite{vonMarcard18ECCV}} & \multicolumn{3}{c}{EMDB~\cite{emdb}} & \multicolumn{3}{c}{RICH~\cite{Huang:CVPR:2022}} \\
    \cmidrule(lr){3-5}
    \cmidrule(lr){6-8}
    \cmidrule(lr){9-11}
    && PA-MPJPE $\downarrow$ & MPJPE $\downarrow$ & PVE $\downarrow$ & PA-MPJPE $\downarrow$ & MPJPE $\downarrow$ & PVE $\downarrow$ & PA-MPJPE $\downarrow$ & MPJPE $\downarrow$ & PVE $\downarrow$ \\
    \midrule
        & BEDLAM1 & 43.2 & 68.0 & 80.7 & 50.0 & 88.7 & 101.6 & 42.1 & 75.2 & 83.2 \\
        & BEDLAM2 & \textbf{41.1} & \textbf{64.8} & \textbf{76.3} & 46.5 & \textbf{74.6} & \textbf{86.2} & 36.8 & 70.8 & 79.4 \\
        & BEDLAM1+2 & \textbf{41.0} & 65.2 & 77.7 & \textbf{46.4} & 75.5 & 87.3 & \textbf{36.4} & \textbf{68.0} & \textbf{75.7} \\
    \bottomrule
  \end{tabular}
  }
\label{tab:hps}
\end{table*}

\textbf{Video-based methods.}
We use B1 and B2 to train two recent SOTA methods that estimate human motion from video, GVHMR \cite{shen2024gvhmr} and PromptHMR \cite{promptHMR}.
For the image-space evaluation of these video-based methods, see the \supmat. 
Here we focus on the world-space evaluation in Tab.~\ref{tab:video_world}.
The top half of the table reports the accuracy of other existing methods (from their respective papers).
Note that these methods are typically trained using a variety of data including real and synthetic (including B1). 

The lower half shows GVHMR and PromptHMR (abbreviated PHMR here) trained on B1, B2 or both. 
Overall, the combination of B1+B2 offers the best results. 
This makes sense, since datasets like EMDB contain activities that are present in B1 but not in B2.
\bedlam contains several motions like sitting and climbing stairs for which the rendered videos do not contain supporting objects. Hence, these motions are non-physical given the scene. 
In \btwo, we remove these to focus on physical plausibility in the 3D scene.
This actually reduces accuracy on scenarios like stair climbing in EMDB relative to B1.
Thus the two datasets are complimentary and users can select whether to include B1 or not, depending on the kinds of motions they anticipate.

What is important to note is that GVHMR and PromptHMR, trained  using {\em only synthetic data}, are more accurate the the originally published versions.
Note that the original versions train using B1 together with real sequences; both are improved by adding B2, even when no real data is used.
For visualizations and more results, including an evaluation of video-based methods on the B2 test set, see the \supmat.

\begin{table*}
    \caption{World-space evaluation of video-based methods.}
    \centering
    \resizebox{\textwidth}{!}{
        
    \begin{tabular}{l|ccccc|ccccc}
        \cmidrule[0.75pt]{1-11} & \multicolumn{5}{c}{RICH (24)} & \multicolumn{5}{c}{EMDB (24)}  \\
        \cmidrule(lr){2-6} \cmidrule(lr){7-11}
        
        Models & WA-MPJPE$_{100}$ & W-MPJPE$_{100}$  & RTE & Jitter & Foot-Sliding & WA-MPJPE$_{100}$ & W-MPJPE$_{100}$ & RTE & Jitter & Foot-Sliding \\
        \cmidrule{1-11}
        DPVO\cite{dpvo} + HMR2.0\cite{hmr2} & 184.3 & 338.3 & 7.7 & 255.0 & 38.7 & 647.8 & 2231.4 & 15.8 & 537.3 & 107.6 \\
        GLAMR \cite{yuan2022glamr} & 129.4 & 236.2 & 3.8 & 49.7 & 18.1 & 280.8 & 726.6 & 11.4 & 46.3 & 20.7 \\
        TRACE \cite{sun2023trace} & 238.1 & 925.4 & 610.4 & 1578.6 & 230.7 & 529.0 & 1702.3 & 17.7 & 2987.6 & 370.7 \\
        SLAHMR \cite{SLAHMR} & 98.1 & 186.4 & 28.9 & 34.3 & 5.1 & 326.9 & 776.1 & 10.2 & 31.3 & 14.5 \\
        WHAM \cite{wham:cvpr:2024} & 109.9 & 184.6 & 4.1 & 19.7 & 3.3 & 135.6 & 354.8 & 6.0 & 22.5 & 4.4 \\
        GVHMR  \cite{shen2024gvhmr} & 78.8 & 126.3 & 2.4 & 12.8 & 3.0 & 111.0 & 276.5 & 2.0 & 16.7 & 3.5 \\
        PHMR~\cite{promptHMR} &  &   &   &   &   & 71.0 & 216.5 & 1.4 & 16.3 & 3.5  \\
        \cmidrule[0.75pt]{1-11}
        GVHMR~\cite{shen2024gvhmr} - B1 & 87.3 & 140.0 & 2.6 & 13.5 & 2.9 & 112.4 & 284.6 & 1.8 & 17.1 & 3.5 \\
        GVHMR~\cite{shen2024gvhmr} - B2 & 75.5 & 120.6 & 2.4 & 12.3 & 2.7 & 113.7 & 284.4 & 2.1 & 15.9 & 3.4 \\
        GVHMR~\cite{shen2024gvhmr} - B1 + B2 & 75.8 & 121.3 & 2.3 & 11.3 & \textbf{2.6} & 109.7  & 273.1 & 1.7 & 15.0 & 3.4 \\

        \cmidrule[0.75pt]{1-11}
        PHMR~\cite{promptHMR} - B1 & 85.7 & 139.4 & 2.9 & 12.7 & 4.0 & 77.6 & 211.1 & 1.4 & 14.9 & 3.4 \\
        PHMR~\cite{promptHMR} - B2 & 75.3 & 122.4 & 2.5 & 11.7 & 2.8 & 71.9 & 197.7 & 1.4 & 12.2 & 3.4 \\
        PHMR~\cite{promptHMR} - B1 + B2 & \textbf{72.5} & \textbf{116.6} & \textbf{2.3} & \textbf{10.2} & \textbf{2.6} & \textbf{70.5} & \textbf{193.7} & \textbf{1.4} & \textbf{11.3} & \textbf{3.2} \\

        \bottomrule
    \end{tabular}
    }
    \label{tab:video_world}
\end{table*}

\section{Conclusions and future work}

\btwo addresses a key need in the community for an extensive ground truth dataset for training methods to estimate 3D human motion in world coordinates, particularly in sequences with moving cameras and changing focal lengths.
This is currently a critical topic for the field.
\btwo provides diverse ground-truth camera motions not present in any other dataset while improving on the original \bedlam dataset in every aspect (body shape, clothing, hair, scenes, shoes).
The results with SOTA methods suggest that training on B2 (or B1+B2), with no real data, achieves world-space accuracy exceeding the recent SOTA.
We release the rendered video sequences, the ground-truth 3D humans, as well as the 3D clothing, hair and shoes; these are all available at \url{https://bedlam2.is.tuebingen.mpg.de/}.
The only assets that we cannot release are the 3D environments and, for these, we provide a ``shopping list'' in Table  \ref{tab:used-assets} of the \supmat explaining how people can obtain the assets.
Code for training, evaluation, rendering are provided, as well as the code we use to retarget AMASS motions to new body shapes.
We also provide the trained model checkpoints for CameraHMR, PromptHMR and GVHMR.

\subsection{Limitations.} A key limitation of B1 and B2 is that they only support people interacting with the ground and not with other objects.
B1 contains motions like sitting or climbing that are not grounded in the 3D scene with object contact. We removed such motions from B2 so that the movements are grounded.
Even so, other than foot-ground contact, other body-scene contacts may not be accurate; for example, during a cartwheel, the hands may not properly interact with the ground.
Generating realistic synthetic sequences of general human-object and human-human interaction remains an open research problem but is clearly the next step for the field.
Such data would support inference of human-object and human-human contact; cf.~\cite{yin2024whac}.
As in B1, the motions in B2 are not semantically meaningful in the context of the scene or the motions of other humans in the scene. 
This limits use of the dataset for some semantic tasks. 
Like previous models, the bodies do not include children, amputees, or people whose body morphology deviates significantly from the mean (e.g.~scoliosis).
Similarly, the movements are from healthy individuals, lacking physical impairments, motor disorders, or supportive equipment like canes or walkers.
And, of course, there is a still a visual domain gap between B2 and real videos.
Despite this, as evidenced by our experiments, B2 is sufficiently realistic to produce SOTA results.
An important direction for future work is to add facial motions and audio, which are completely lacking in B1 and B2. This is necessary to develop synthetic data that supports reasoning about direct human-human communication.

\subsection{Broader impacts.} Synthetic data of people has a huge advantage over real data, which is typically scraped from the Internet without consent. 
In contrast, synthetic data reduces privacy concerns.
The primary use case of \btwo is to train methods to estimate 3D humans from video; this has positive and negative use cases. 
We use a custom license that prohibits use of the data for ``pornographic, military, or surveillance, purposes," as well as to ``create fake, libelous, misleading, or defamatory content."

In addition to supporting work on human motion estimation, \btwo is useful for training and evaluating methods for 3D/4D point tracking, structure from motion estimation with non-rigid motions, depth estimation, optical flow, and dynamic scene reconstruction.
Recent work on these topics \cite{dust3r,monst3r} is limited due to a lack of ground truth sequences for end-to-end training.

\textbf{Acknowledgments.} We thank STUDIO LUPAS GbR for creating the 3D clothing and  Meshcapade GmbH for the skin textures. We thank T. Alexiadis, T. Obersat, C. Gallatz, A. Bertler, A. Cseke, A. Kuznetcova, F. Doll, S. Bhor, T. Rakshit, T. Niewiadomsky and  V. Fourel for 3D outfit texturing and quality evaluation of the clothing simulations. We thank the Software Workshop at MPI-IS for deploying the dataset-statistics web app.

\textbf{Disclosure:} While MJB is a co-founder and Chief Scientist at Meshcapade, his research in this project was performed solely at, and funded solely by, the Max Planck Society.

\bibliographystyle{plain}
\bibliography{B2,bedlam,datasets}

\newpage
\section{Appendix}

In this Appendix, we provide additional details in roughly the order they are mentioned in the main paper.
Please see \url{https://b2dash.is.tuebingen.mpg.de/} for a wide range of dataset statistics, presented graphically.

Note: We maintain a list of known issues on the project website at \url{https://bedlam2.is.tuebingen.mpg.de/}. 
If you identify a problem with any part of the dataset please send a report to \texttt{bedlam@tue.mpg.de} and include a ``[BEDLAM2]'' tag in the subject line to indicate that your report is specific to \btwo.

\subsection{Details of the camera motions}

\paragraph{Synthetic camera motion generation.}
We extend the limited camera motions and focal lengths of \bedlam by adding a new camera movement component to the render pipeline pre-processing stage. It uses randomizable camera movement configurations and outputs the desired movement definitions as camera pose keyframes for the Unreal Sequencer in JSON format. This is then used with Unreal Python to fully automate the generation of needed Unreal Level Sequences for rendering.

Synthetic camera shot types:
\begin{itemize}
    \item Panning shots at static location, optionally augmented by tracking of target body parts
    \item Tracking camera shots for a moving target body. Maintains distance and optionally also viewpoint to the target body. Target location and rotations are low-pass filtered to ensure smooth camera motions.
    \item Dolly shots (left-right, forward-backward, diagonal, up-down/crane)
    \item Orbit shots (tracking fixed target location or moving target body)
    \item Zoom shots with varying focal length, can be combined with other shot types
    \item Camera shake with Perlin noise with randomized intensity, can be combined with other shot types
\end{itemize}

\paragraph{Captured camera motions.}

All motion capture shots are captured for a target stimulus at the origin which allows us to later randomize the viewpoint onto the subject by rotating the mocap data. We also vary the strength of the capture device shake between the various capture sessions. 

Handheld footage is captured in landscape or portrait mode with Apple iPhone Pro 14 and Google Pixel 4a phone devices and an iPad Pro 11 tablet device running the Unreal VCam application. This application captures camera extrinsics with the ARKit/ARCore tracking components which estimate 6DOF device pose through visual odometry by sensor fusion of device camera and IMU data, similar to device pose capture approaches in \cite{chen2025_imageimu}. The camera pose is sent from the handheld device to a Windows PC running a custom Unreal Engine scene with target stimulus. A real-time render of the scene is streamed back from the PC to the handheld device to provide immediate feedback on the current camera pose values, helping the operator to properly frame the stimulus. The PC is also recording the camera pose which we later auto-export from its Unreal-specific binary format into a reusable JSON representation compatible with the synthetic camera motion generation pipeline.

\begin{figure}[ht]
    \centering
    \includegraphics[height=2.00in]{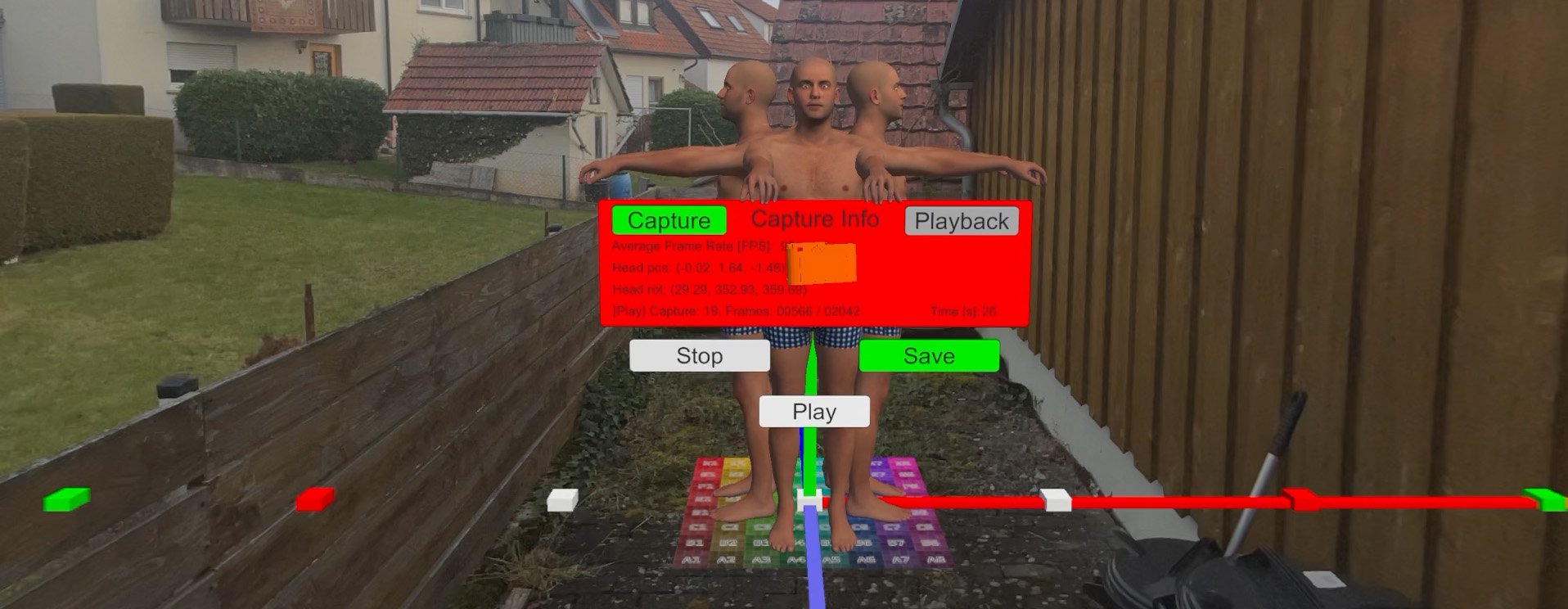}
    \caption{\textbf{Egocentric Camera Motion Capture} In-the-wild 90Hz camera pose capture setup on Apple Vision Pro, where our custom Unity 6 app renders a real-time 3D reference stimulus. The user can start, stop and review captures by interacting with the provided 3D user interface.}
    \label{fig:new_outfits_supmat}
\end{figure}

Egocentric footage is captured on Apple Vision Pro mixed-reality headset running a custom Unity 6 3D head pose capture application. It provides an egocentric 90fps real-time render of the stimulus scene based on the head pose of the user. We record the user head pose at 90Hz. To ensure consistent data capture timestamps we first record head pose into a large pre-allocated data array in memory and only save to local device storage at the end of the capture session. Captured data can optionally be played back on device and visualized with a virtual camera 3D model for initial quality assessment. This capture setup is completely self-contained without external device dependencies and can be used wherever there is enough walking space for the desired camera motion. This approach allows us to capture camera motions that go beyond the space limitation of typical optical motion capture studios.

\subsection{Dataset statistics}

\btwo includes:
\begin{itemize}
    \item 8M images with realistic motion blur using 7 temporal subframes
    \item 27480 video sequences at 1280x720 resolution, 30fps
    \item 74.52 hours of video
    \item ground truth SMPL-X bodies with 16 shape coefficients
    \item 1,615 diverse body shapes including high and low BMI
    \item 4,643 motions sampled from AMASS, MOYO and BEAT2
    \item 187 unique clothing outfits
    \item 182 unique shoes
    \item 40 unique hairstyles with 50-100K 3D strands
    \item 10,592 unique combinations of body shape, motion, and clothing
    \item widely varied focal lengths and camera motions with realistic noise and ground truth
    \item 94 HDRI environments and 15 detailed 3D environments
    \item ground truth depth maps 
    \item 26TB of data
\end{itemize}

Note that the core dataset distribution includes the images together with the associated SMPL-X ground truth and camera ground truth. 
We also provide the 3D clothing, shoes, and hair; Subsection \ref{sec:assets} describes the assets in detail and how to obtain them.

Detailed interactive plots with comprehensive statistics that cover the rendered images, cameras and camera motions as well as bodies and animations can be found at this website: \url{https://b2dash.is.tuebingen.mpg.de/}

\subsection{Shoes}

In contrast to previous datasets based on SMPL and SMPL-X, here we add shoes to the model.
Roughly 40\%\/ of the rendered images contain bodies with shoes.
It is easy for users of the dataset to sample sequences with our without shoes as needed.

Shoes are represented using the SMPL-X UV space by defining texture, displacement, and normal maps, which we obtain from the Google Scanned Objects dataset.
Here we provide details of how we transform shoes from the dataset and apply them to SMPL-X.

\paragraph{Stocking feet.}
We first modify the foot of the SMPL-X template mesh to remove the shape of the toes, while keeping the mesh topology the same (see Fig.~6a in the main paper).
We find that applying the standard shape parameters ($\beta$) to this modified template mesh produces natural looking variations in foot shape.
We  make this modified SMPL-X template, with stocking feet, available for research.

\paragraph{Texture Extraction.}
We align, rotate %
and scale the raw scans from the dataset to create a common shoe size that corresponds to the neutral SMPL-X base foot (zero-pose, zero-shape). 
We then translate all shoes %
so that they align with the SMPL-X base feet. 

Textures are extracted from the shoe models using the xNormal software (\url{https://xnormal.net}). It works by raycasting from the right foot to the right shoe to get color, normal and displacement values. These values are mirrored for the left foot. 

We make the code for putting shoes on SMPL-X available and this contains all the details.

\paragraph{Sole thickness and ground contact.} Sole thickness %
for each shoe is calculated as the mean displacement value in the displacement map corresponding to the sole of that shoe. The body is then translated vertically by the mean amount.

\subsection{Hair}

Note that we cannot say that having realistic hair actually improves HPS accuracy as we have not performed an ablation study that removes hair. 
While this remains unclear, realistic hair may serve other uses like training generative models of 3D hair and hair motion.
We hope that the 3D assets will find uses beyond our present application.

\subsection{Clothing}
More examples of new complex outfits and graded outfits are shown in Figure \ref{fig:new_outfits_supmat} and Figure \ref{fig:grading_supmat} respectively, and in the supplemental video (\url{https://youtu.be/ylyqHnwhpsY}).
Note that an ``outfit" may include multiple pieces of clothing; e.g.~a man's suit includes the pants, dress shirt, and jacket.  These individual pieces can be rendered individually or in new combinations. This gives users of the dataset flexibility in generating new data.

\begin{figure}[t]
    \centering
    \includegraphics[height=1.05in]{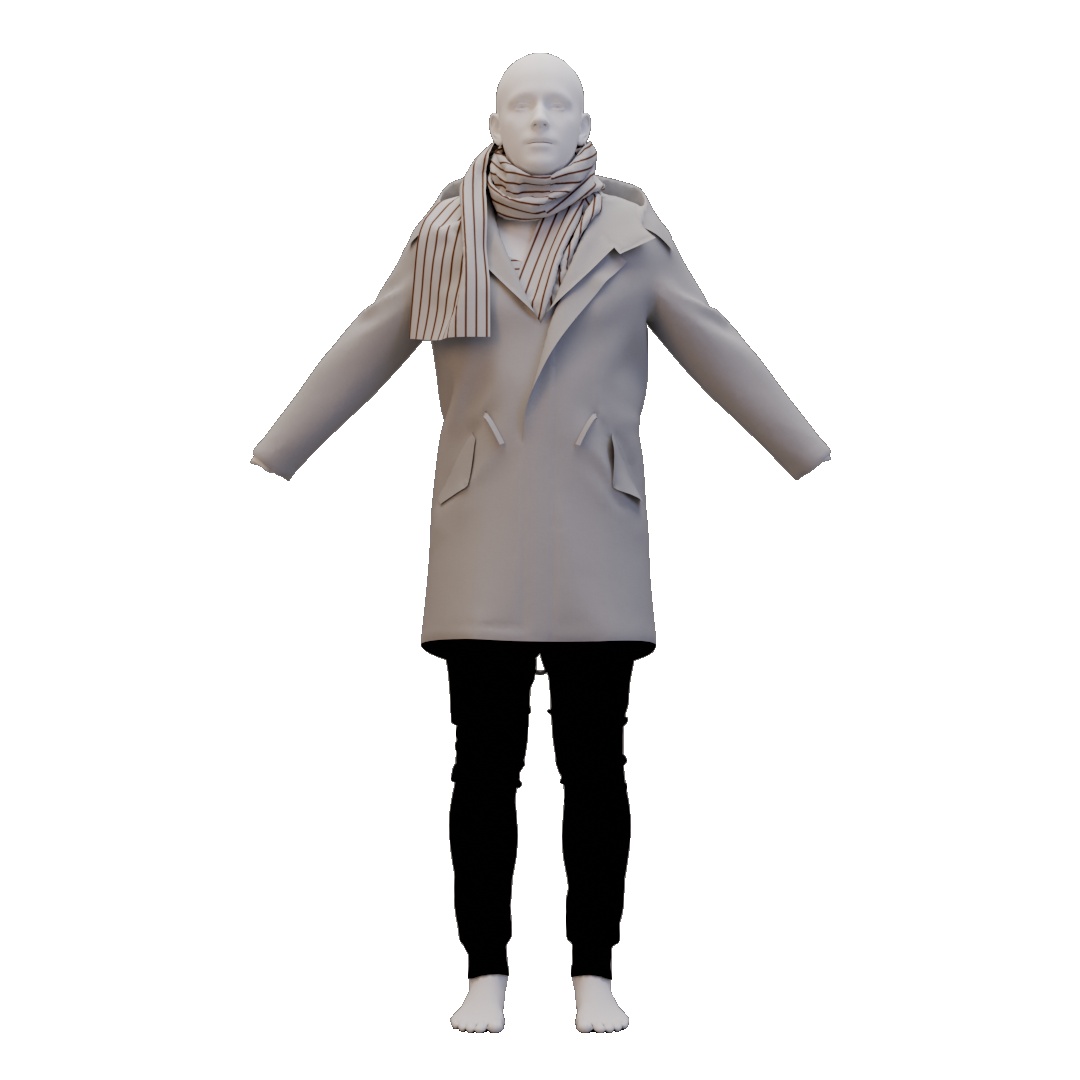}
    \includegraphics[height=1.05in]{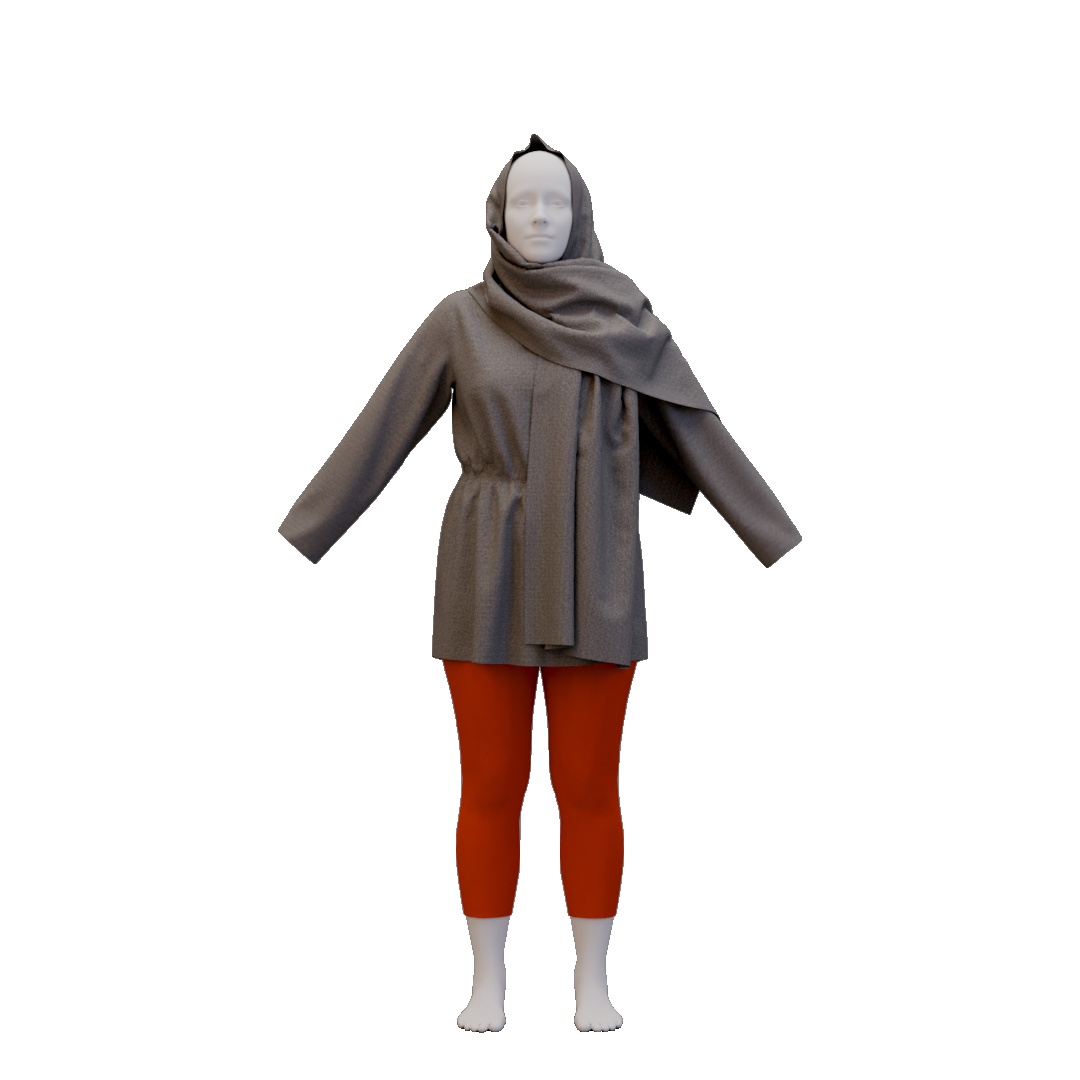}
    \includegraphics[height=1.05in]{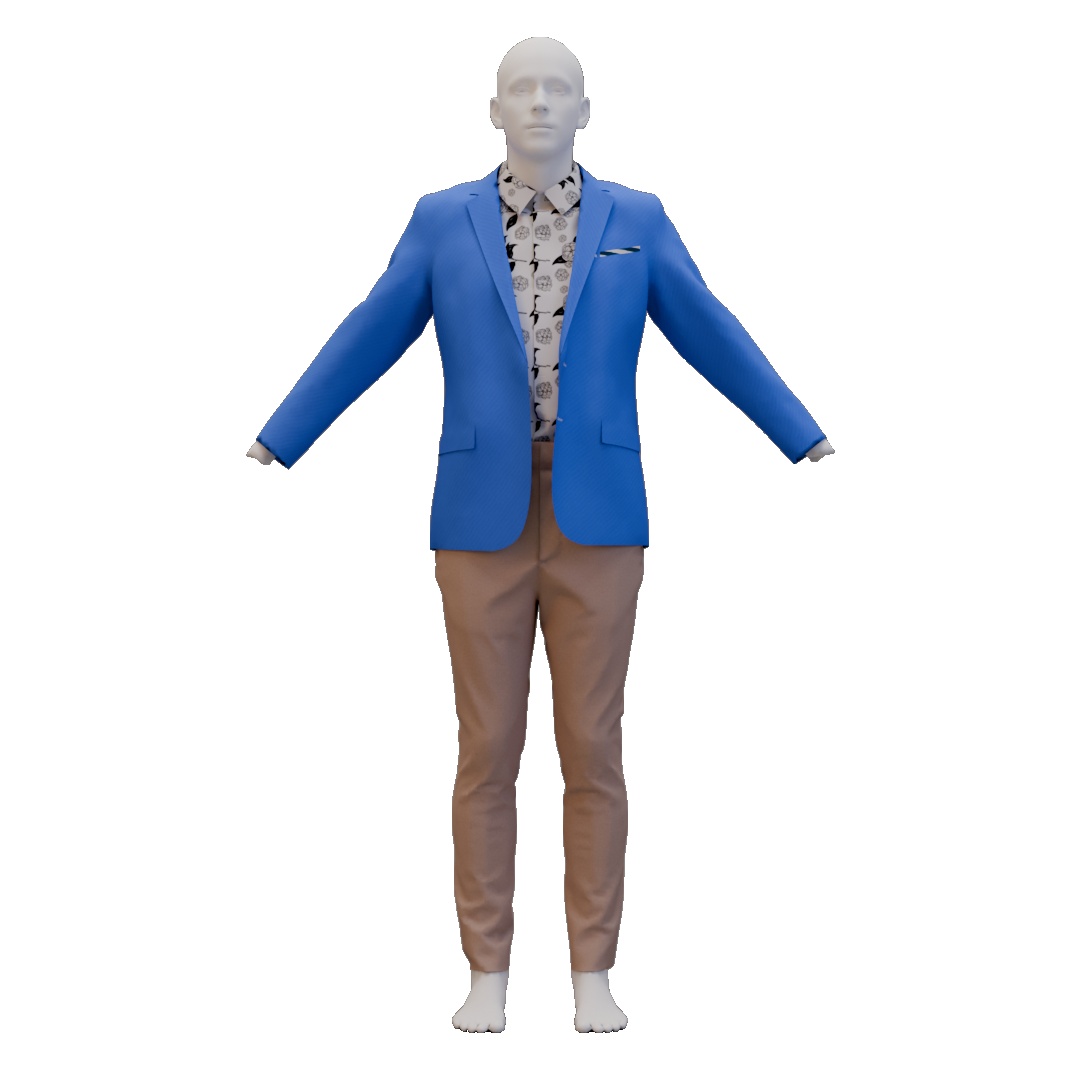}
    \includegraphics[height=1.05in]{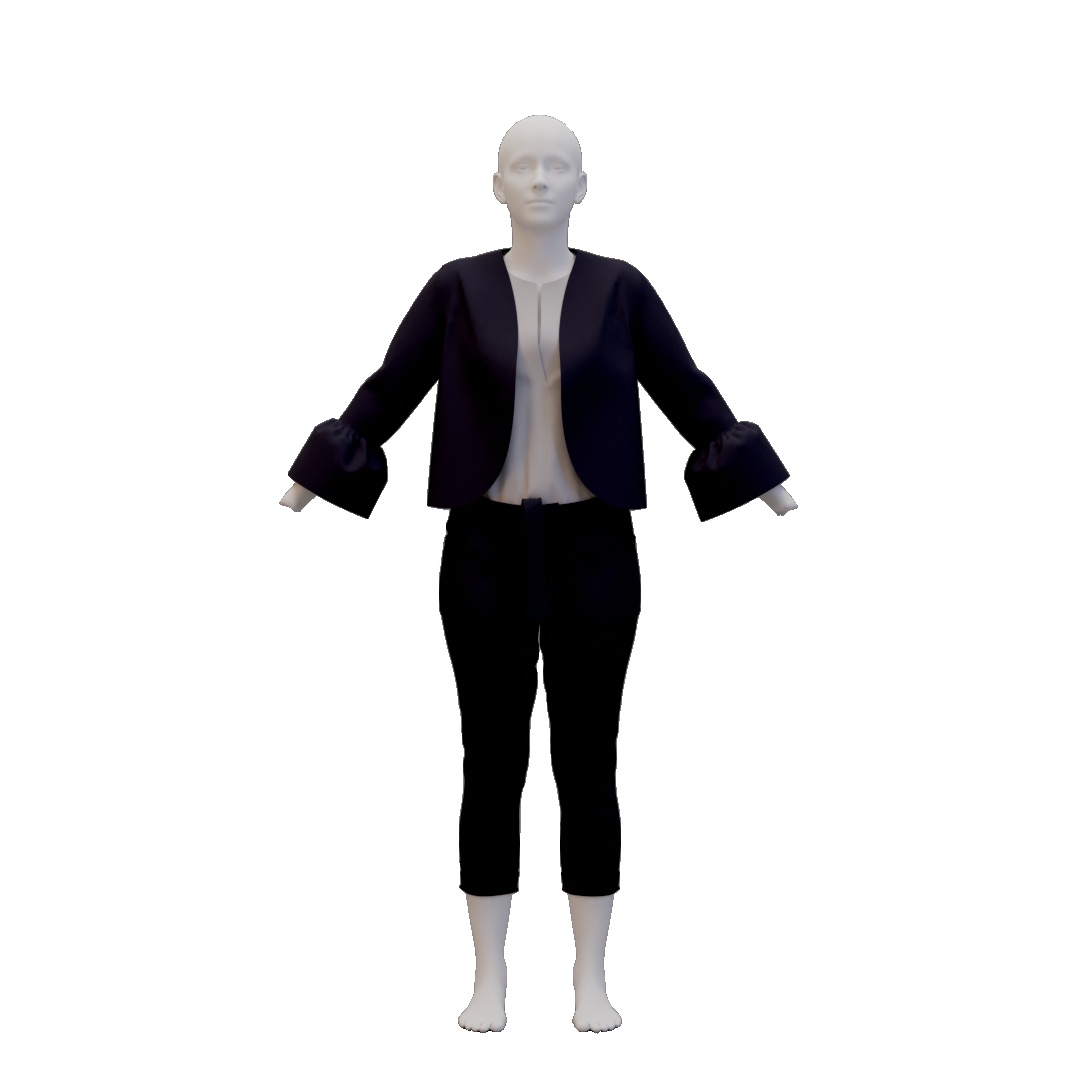}
    \includegraphics[height=1.05in]{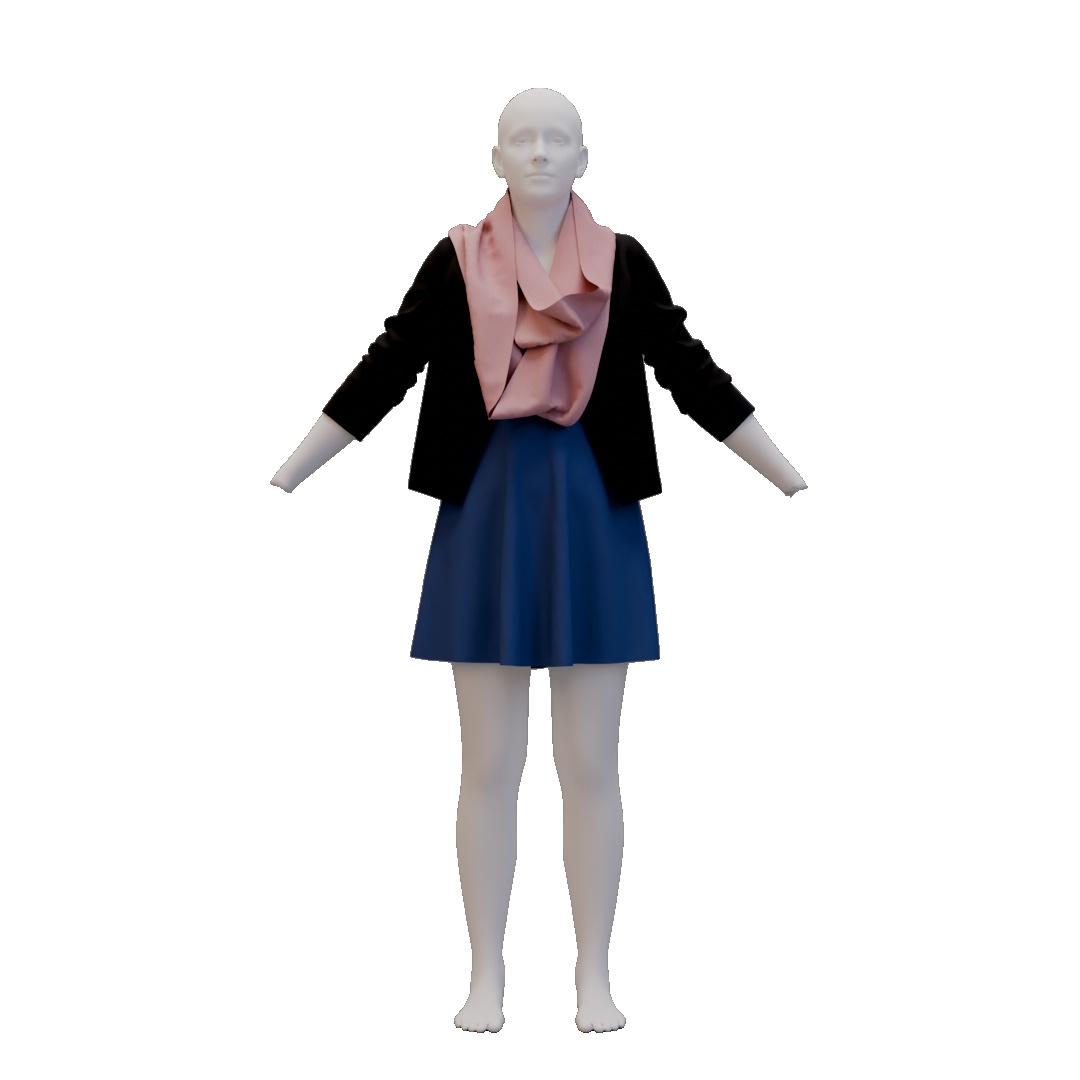}
    \\
    \includegraphics[height=1.05in]{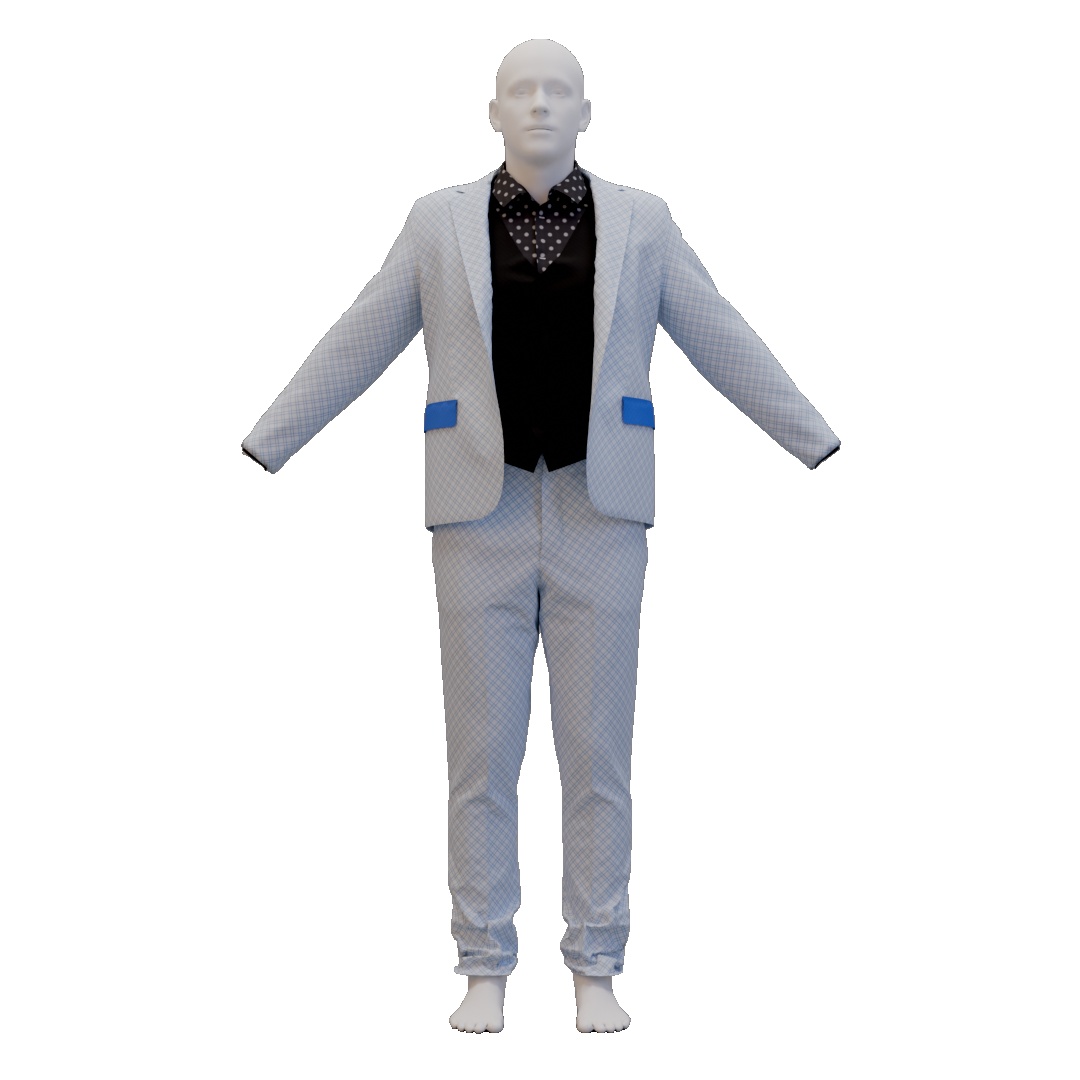}
    \includegraphics[height=1.05in]{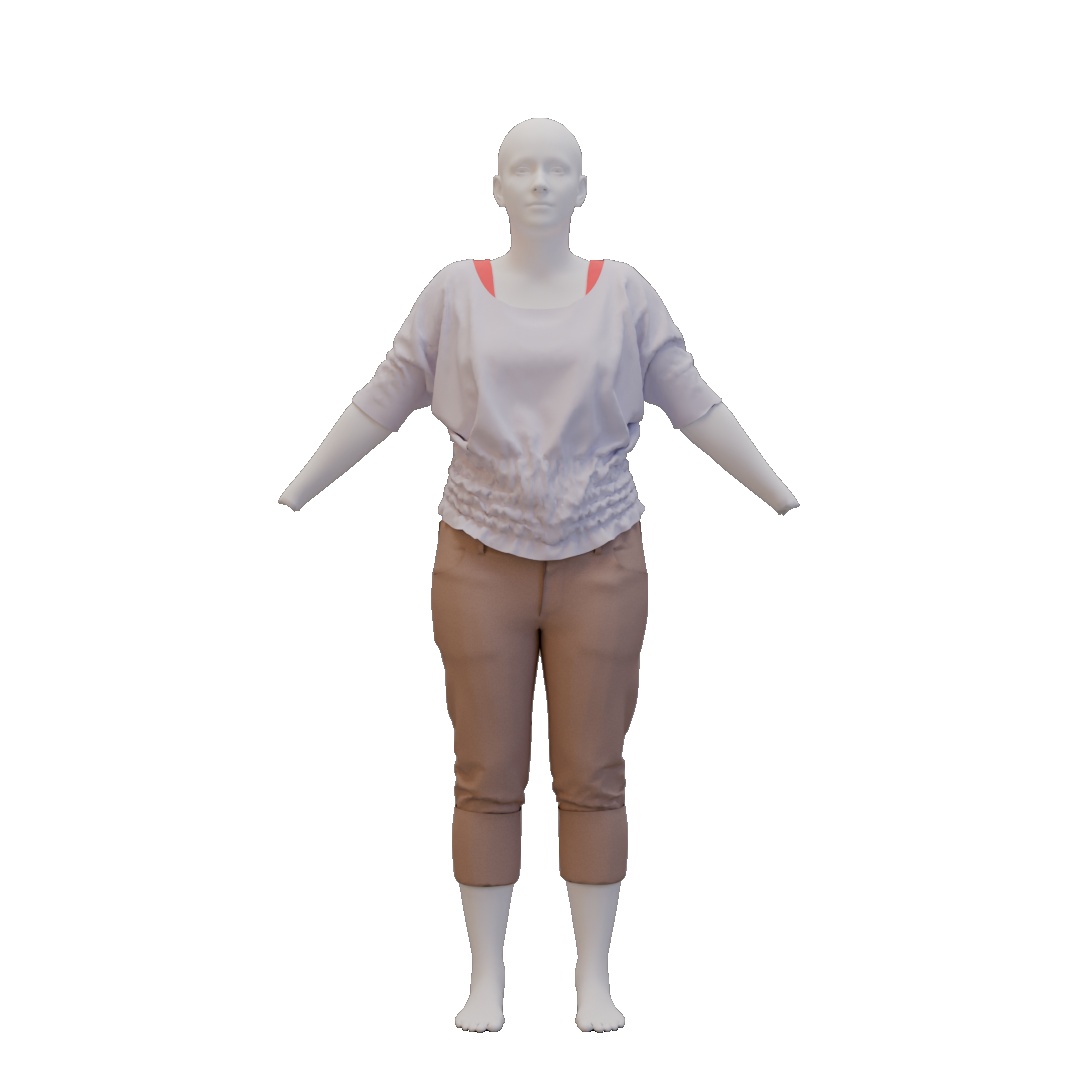}
    \includegraphics[height=1.05in]{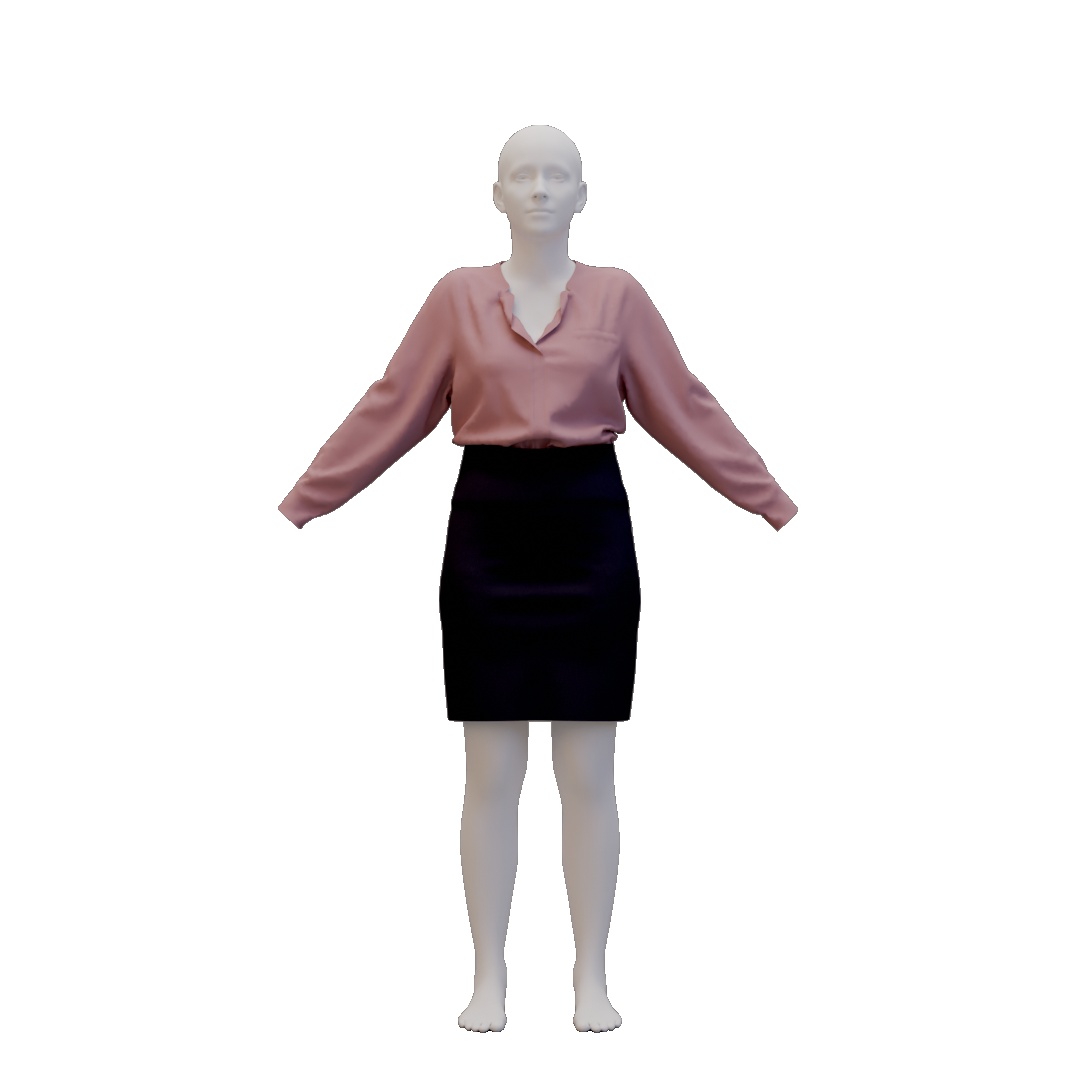}
    \includegraphics[height=1.05in]{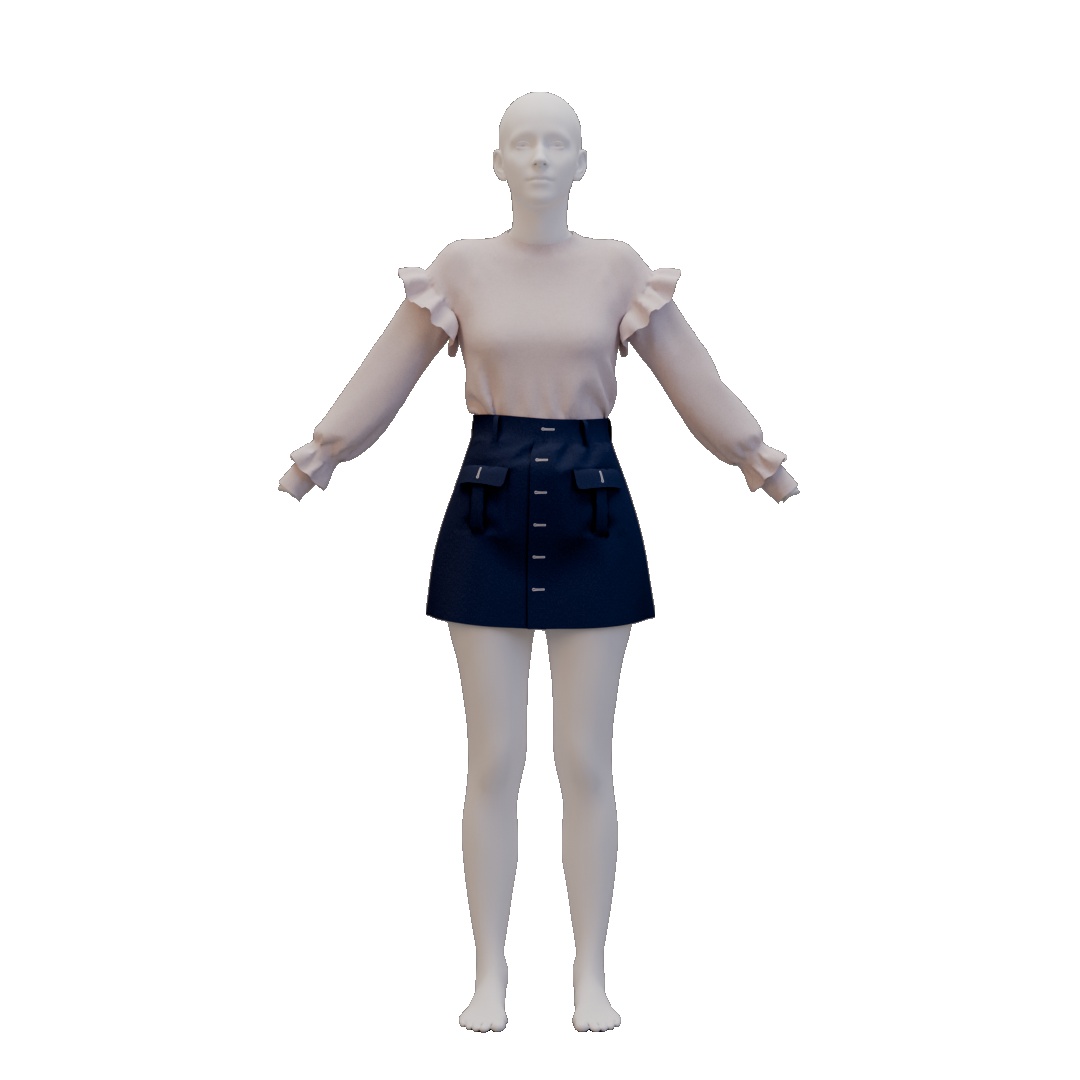}
    \includegraphics[height=1.05in]{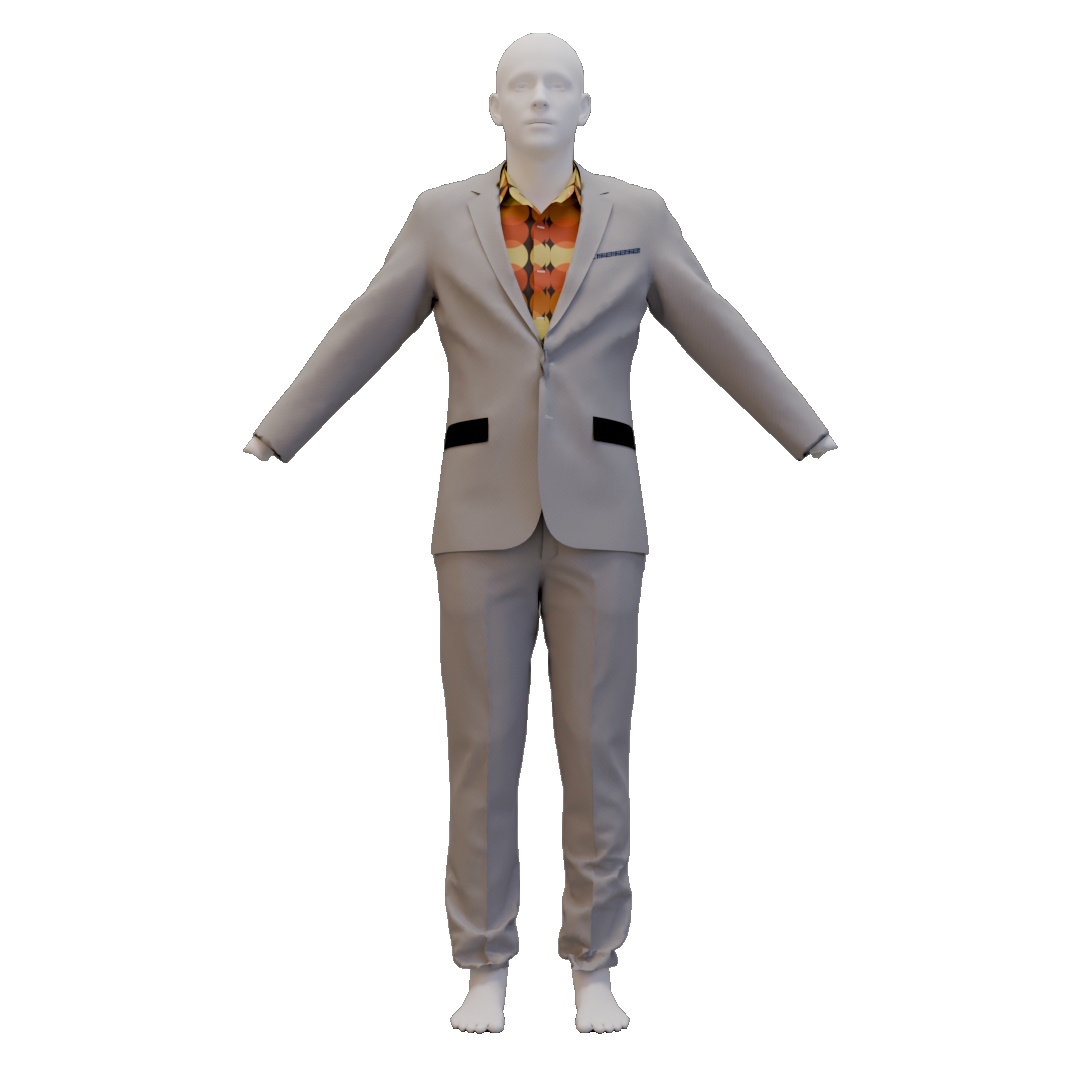}
    \caption{\textbf{New outfits.} We added 76 new and more complex clothing outfits to \btwo. }
    \label{fig:new_outfits_supmat}
\end{figure}

\begin{figure}[t]
    \centering
    \includegraphics[height=1.05in]{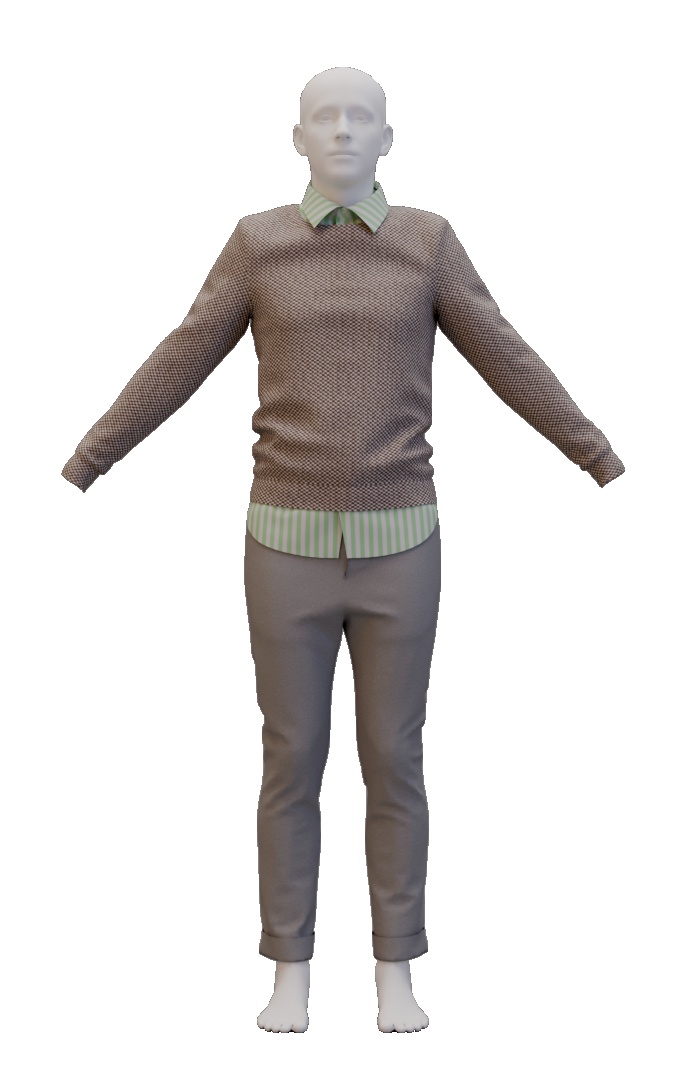}
    \includegraphics[height=1.05in]{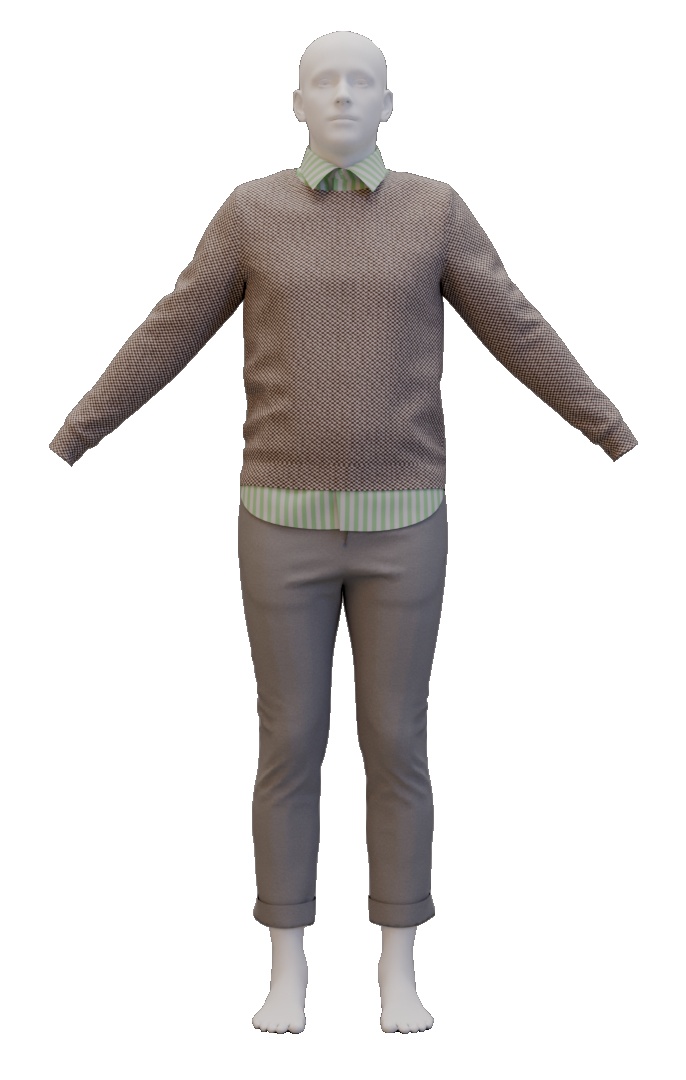}
    \includegraphics[height=1.05in]{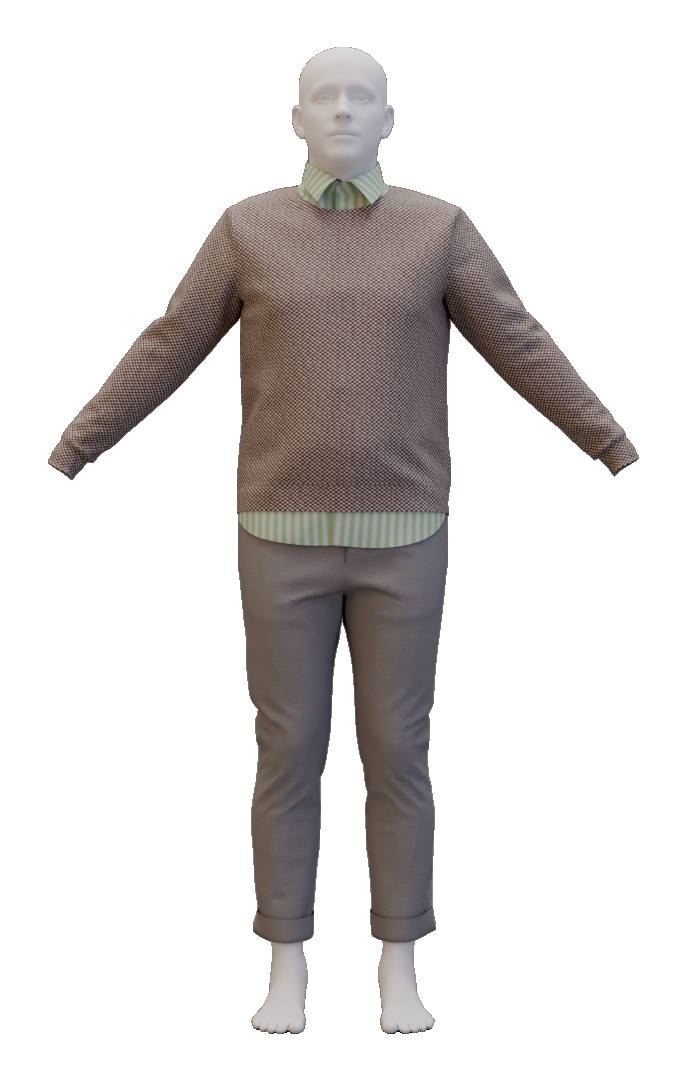}
    \includegraphics[height=1.05in]{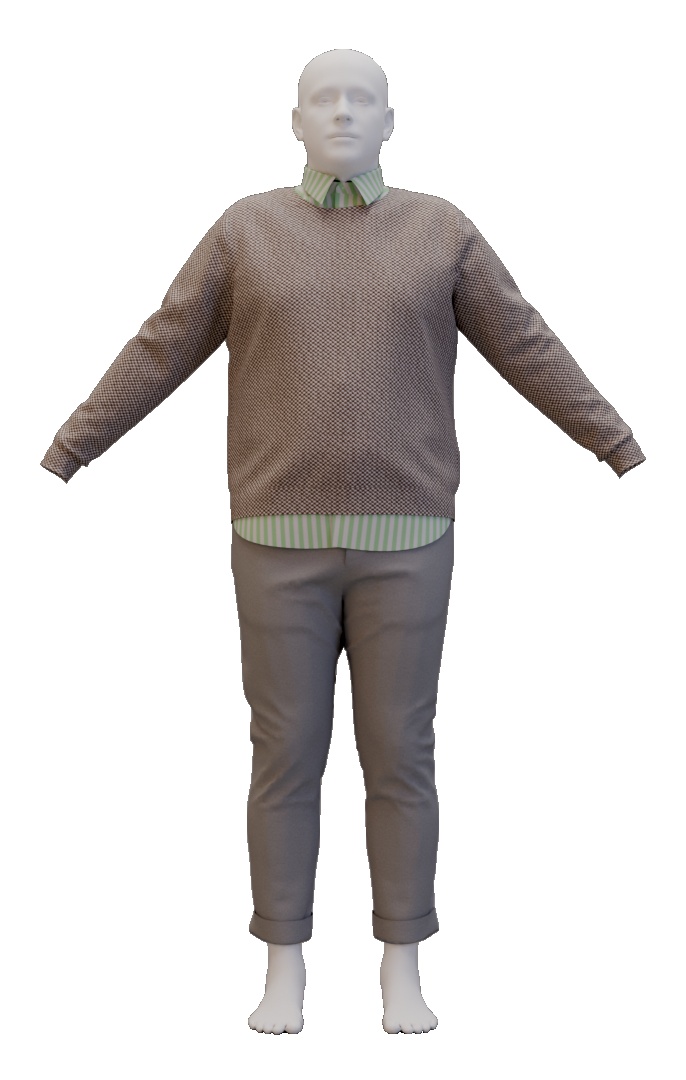}
    \includegraphics[height=1.05in]{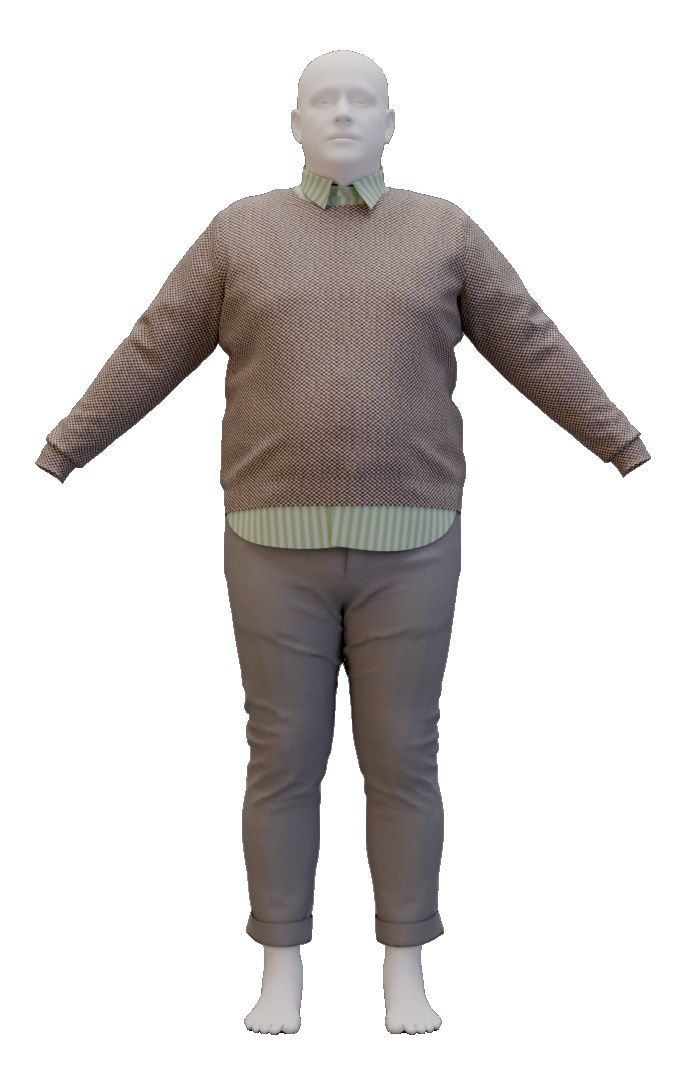}
    \includegraphics[height=1.05in]{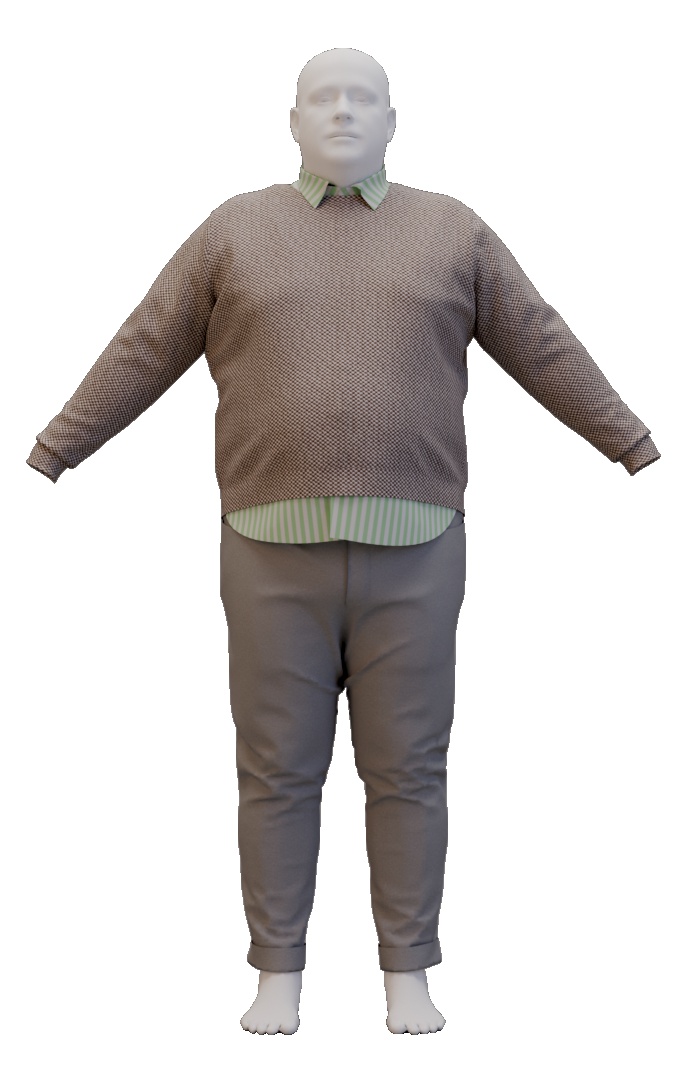}
    \includegraphics[height=1.05in]{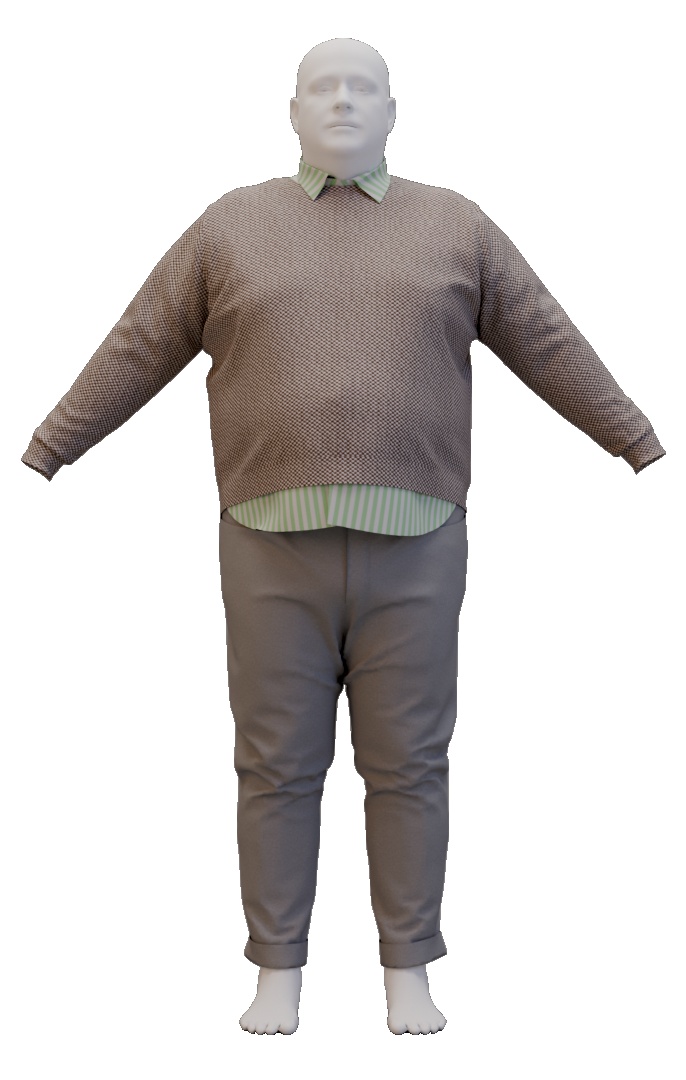}
    \\
    \includegraphics[height=1.05in]{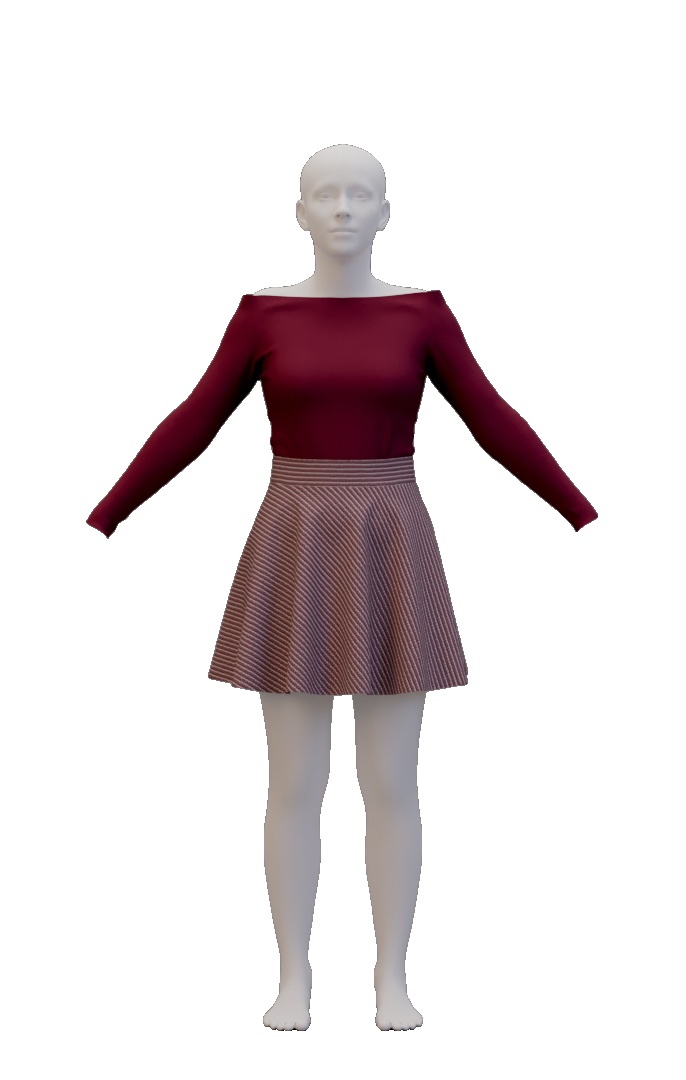}
    \includegraphics[height=1.05in]{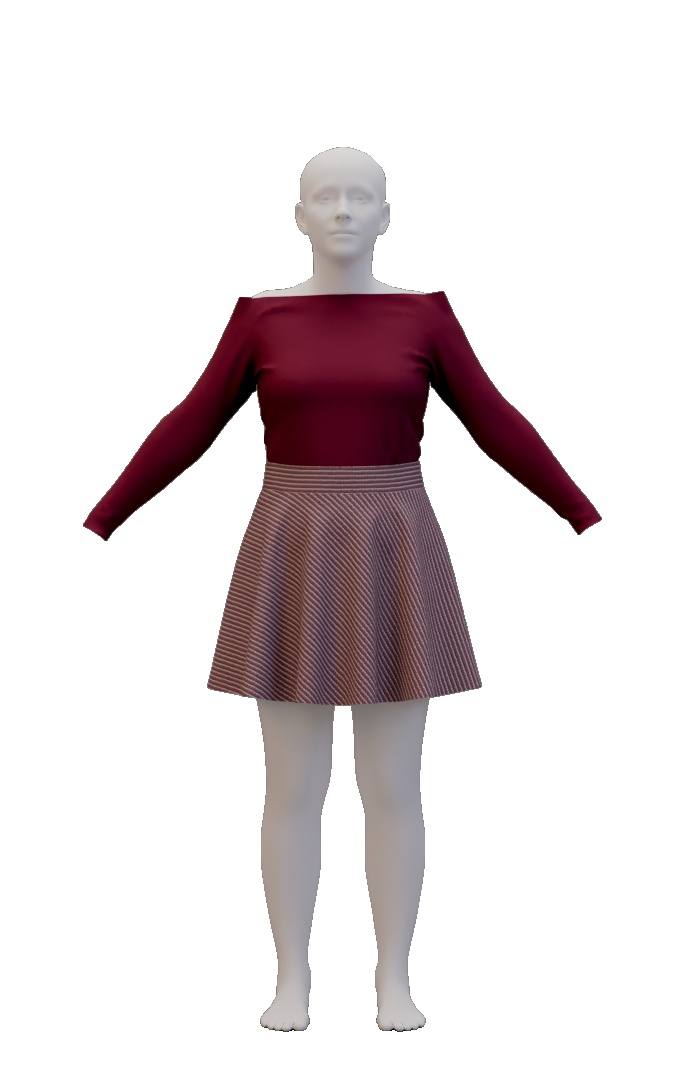}
    \includegraphics[height=1.05in]{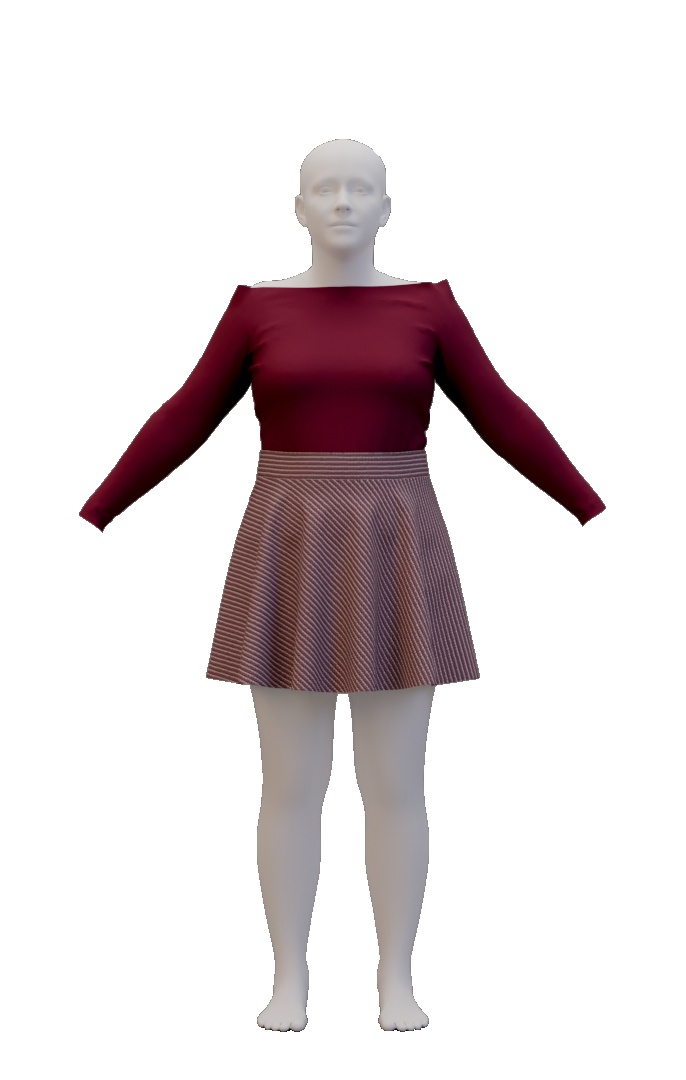}
    \includegraphics[height=1.05in]{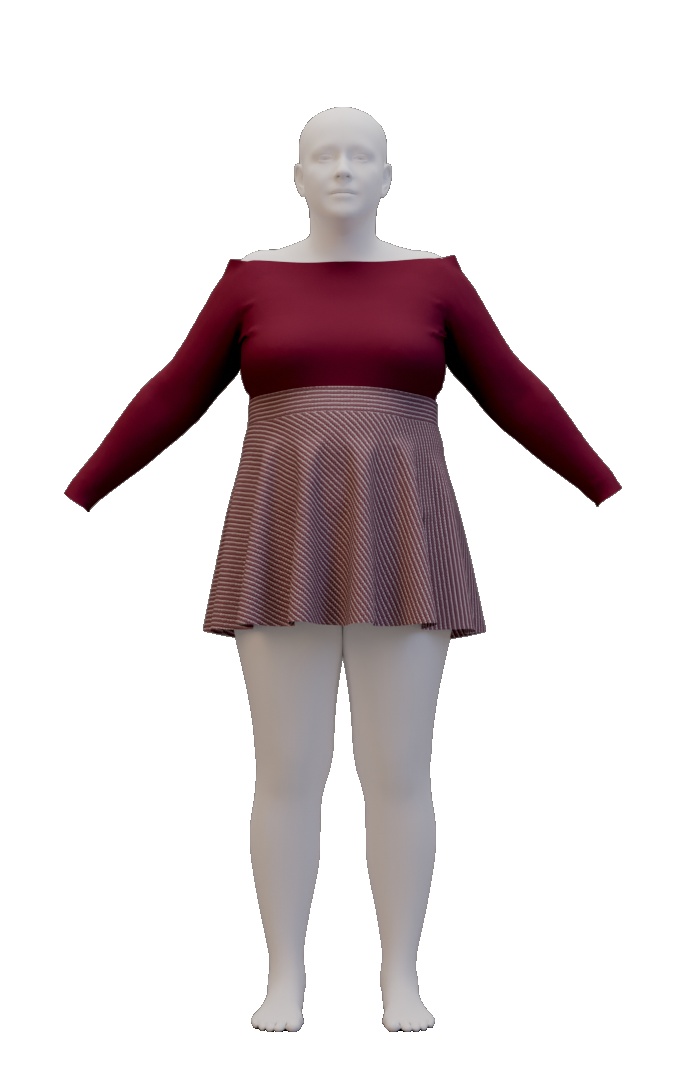}
    \includegraphics[height=1.05in]{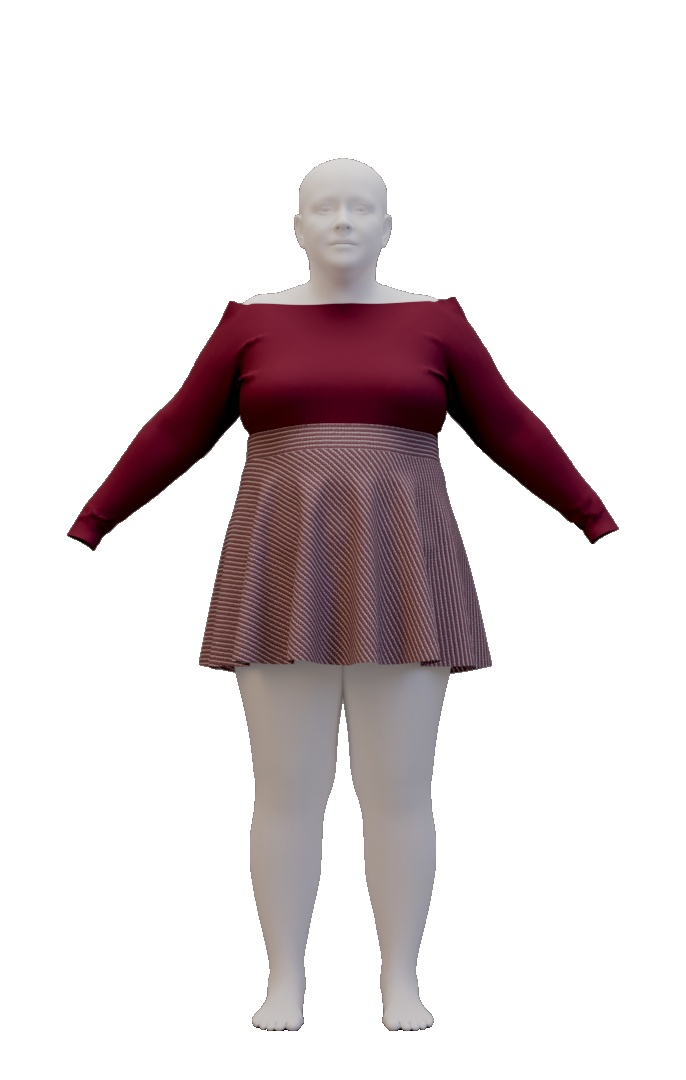}
    \includegraphics[height=1.05in]{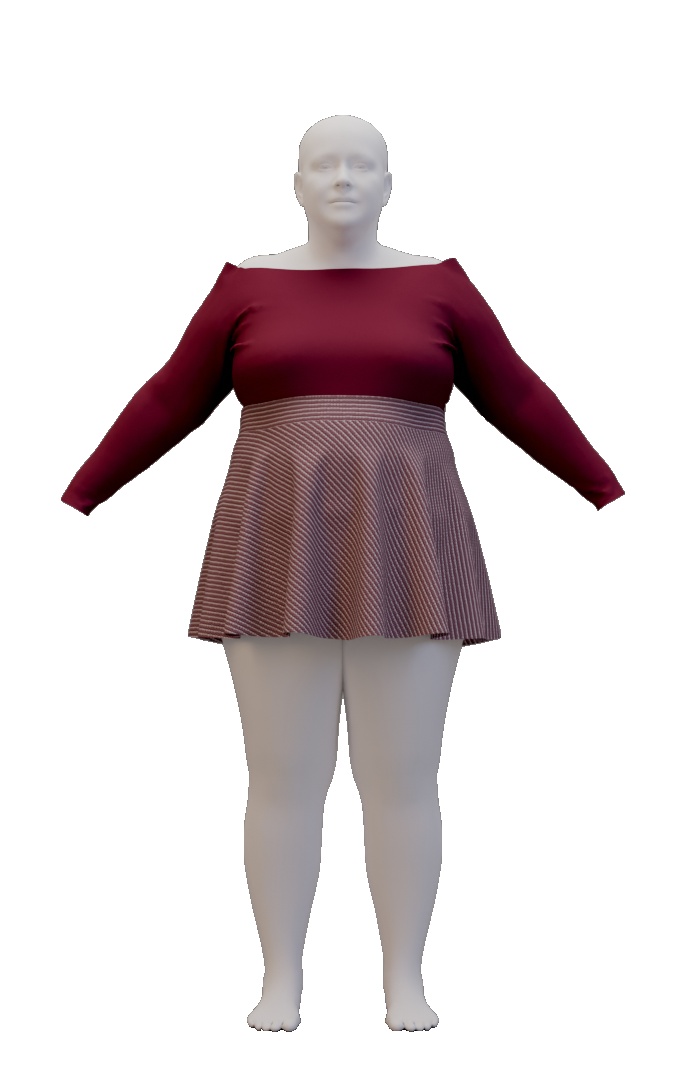}
    \includegraphics[height=1.05in]{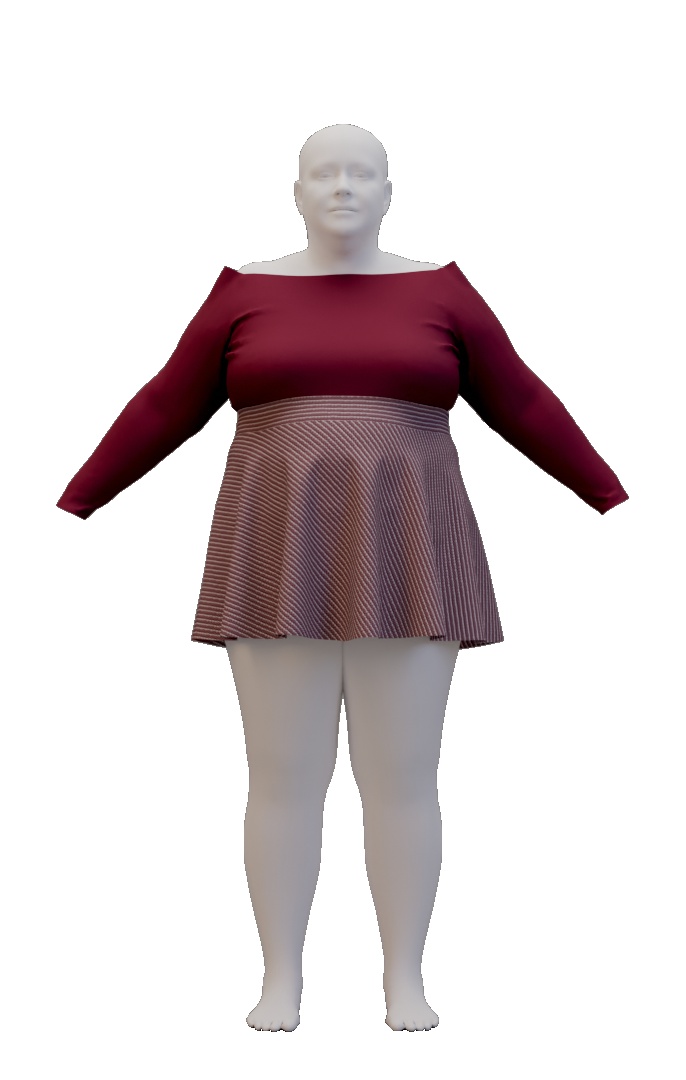}
    \caption{\textbf{Grading.} More examples of graded outfits. Here are shown, from left to right, sizes XS, M, L, 2XL, 3XL, 5XL and 6XL. }
    \label{fig:grading_supmat}
\end{figure}

\paragraph{Simulation.} \label{subsection_clothingsim} The garments in motion are physically simulated using CLO. We pin some of the vertices of the outfit mesh to the body to avoid garments sliding and falling. Clothing simulation is extremely sensitive to mesh interpenetration; garments can get stuck to body parts and be pulled and deformed in extreme and unnatural ways, causing the simulation to break. Due to the fact that the SMPL-X model does not have the ability to deform the body based on contact, body self-interpenetration is quite common. 
Additionally, when we apply 3D motions captured on a slim subject to a high-BMI body shape, this can cause self-penetration of the body parts.
In order to reduce the number of simulation failures, we temporarily remove the hands of the body mesh during simulation because they are very likely to interpenetrate the body, especially in the thigh region. A correct outfit size assignment (described in Section~3.6 of the main paper) is also crucial to reducing the number of failures, allowing us to successfully simulate people with a wide variety of body shapes and complex outfits. The simulation results are visually checked and rated in order to exclude failures.

Due to their complexity of the yoga motions in the MOYO dataset, we do not use clothing simulation for these.
Instead we use texture maps, which look like tight-fitting yoga clothing. This type of clothing is appropriate for these motions.

\subsection{Depth data}

Since we are rendering the dataset, we can also render depth maps.
While we do not use these for training HPS methods, there are many applications where it is useful to have the depth maps associated with the video sequences.
We render approximately 44\%\/ of the images with the associated depth map; see \url{https://b2dash.is.tuebingen.mpg.de/}.

Note that, to create realistic motion blur, each RGB frame of the video is generated from 7 rendered subframes. 
Note also that the depth data is not blurred.  Thus it is critical that the depth data correspond to a specific point in time, which we chose to the center subframe in the set of 7.

By default, Unreal saves the last subframe camera information and not the center one in the generated EXR files. 
To address this we extended the existing MovieRenderQueue C++ plugin with functionality to store camera ground-truth values for every subframe. This allows us to then properly store the ground truth for the center subframe in the EXR image metadata section.

Data generated before 10/2024 used the last subframe and we fixed this with post processing.
Specifically, we first determined the fixed time delta between the last and center subframe by modifying the Unreal C++ source code to obtain this information. 
For our use case with 7 temporal samples at 30fps render rate, we obtained a subframe delta time of 0.002375s which results in 0.007125s time difference between the last subframe and the center subframe when using 7 temporal samples.
We then used that information to resample the camera pose ground truth from the last subframe to the center subframe using Piecewise Cubic Hermite Interpolating Polynomials for a tight fit to original data without overshooting. 
This allowed us to re-render the depth at the center subframe with high accuracy.
We also log this information in the camera ground truth JSON files so that it is clear if the camera pose data was resampled in post or correctly obtained from our custom center subframe ground truth render plugin.

\subsection{Normal data}
\btwo does not contain normal image data. To help researchers who are interested in this type of data we are releasing the \btwo render pipeline code for Unreal Engine 5.3, similar to our previous \bedlam render code release for Unreal Engine 5.0. Access details are provided on our project website (\url{https://bedlam2.is.tuebingen.mpg.de/}).
The new 5.3 code includes the functionality to also output camera-space or world-space normal images in EXR format. This functionality can be useful for downstream tasks like normal estimation or avatar creation from a single image.

\subsection{Comparisons with other datasets}

It is hard to fully compare synthetic datasets numerically since they rarely report all the details and the details that are reported often differ.
Here we try to bring together the relevant, and known, details to enable numerical comparison.

There have been several static image datasets like SynthBody \cite{hewitt2024look}, BEDLAM-CC \cite{wang2024blade}, and PDHuman \cite{wang2023zolly} that are all similar to BEDLAM but address different issues. SynthBody adds more facial detail than is present in BEDLAM, while the other two datasets focus on close-up shots of the body with wide-angle lenses.
We focus on the datasets that introduce camera motion.

\textbf{SynBody \cite{yang2023synbody}:}
This dataset is very similar to BEDLAM in terms of goals and approach. 
They sample 10K body shapes from the SMPL-X shape space and generate 6,960 sequences with 1.2M images and ground truth SMPL-X. These are generated in 6 3D scenes.
They use 1,187 motions from AMASS.
The dataset is generated from 4 fixed camera viewpoints; it does not include camera motion.
As reported in \cite{cai2023smplerx}, SynBody is not as effective for training as BEDLAM. 
The reasons are not completely clear but it likely has to do with the visual quality, which is lower for SynBody.

\textbf{WHAC-A-Mole \cite{yin2024whac}: }
This dataset is built on SynBody and inherits its limited realism.
What it adds, however, are sequences with human-human interactions, including dancing, together with diverse camera motions. 
Like \btwo, they use a range of standard camera motions but, in addition, they use combinations of these.
Overall, the dataset has 1.46M crops and 2434 sequences.

\textbf{PACE \cite{pace2024kocabas}:}
Camera trajectories use heuristics and include dolly zoom, arc motions, tracking shots, and motions from the MannequinChallenge dataset \cite{mannequin}. 
The bodies are scans from RenderPeople (see \cite{Patel21CVPR}), the motions are from AMASS \cite{AMASS}, and the 3D scenes are from Unreal Marketplace. 
The dataset contains only 25 video sequences with 1-8 people per sequence.

\textbf{HumanVid \cite{HumanVidSyn}:}
HumanVid generates synthetic sequences with either SMPL-X bodies or anime characters; we focus on the former here. For this, they use BEDLAM bodies, motions and clothing. They appear to use only one body per sequence.
Where they go beyond BEDLAM is in designing a rule-based camera motion generation pipeline.
Given a body, they sample camera locations in a semicircle in front of the body, point the camera at the person, and randomly sample the camera roll.
They form a camera path between these camera keyframes using an interpolating spline.
With this, they generate 50K clips with a total of 8M frames.

\textbf{EgoGen \cite{egogen}:} Goes further to use generated human motions in varied environments, effectively enabling the capture of an infinite amount of synthetic egocentric video.
There have been several prior efforts \cite{xregopose,wang2024egocentric} to create egocentric data but EgoGen is more realistic in that it is built on BEDLAM assets. Rather than use physics simulation for the clothing as done here, they use a neural garment simulation method \cite{grigorev2022hood} to make the process more automatic and scalable.
They generate two datasets. The depth dataset has 105,000 depth images, while the RGB dataset has 301,073  images.
The key property of this dataset is that the generated motion of the body drives the camera, producing diverse and natural looking camera motions. We get similar motions for \btwo by directly capturing them from an Apple Vision Pro or handheld device.  

\subsubsection{Focal Length Distribution Comparison}
\begin{figure}[ht]
    \centering
    \includegraphics[height=2.2in]{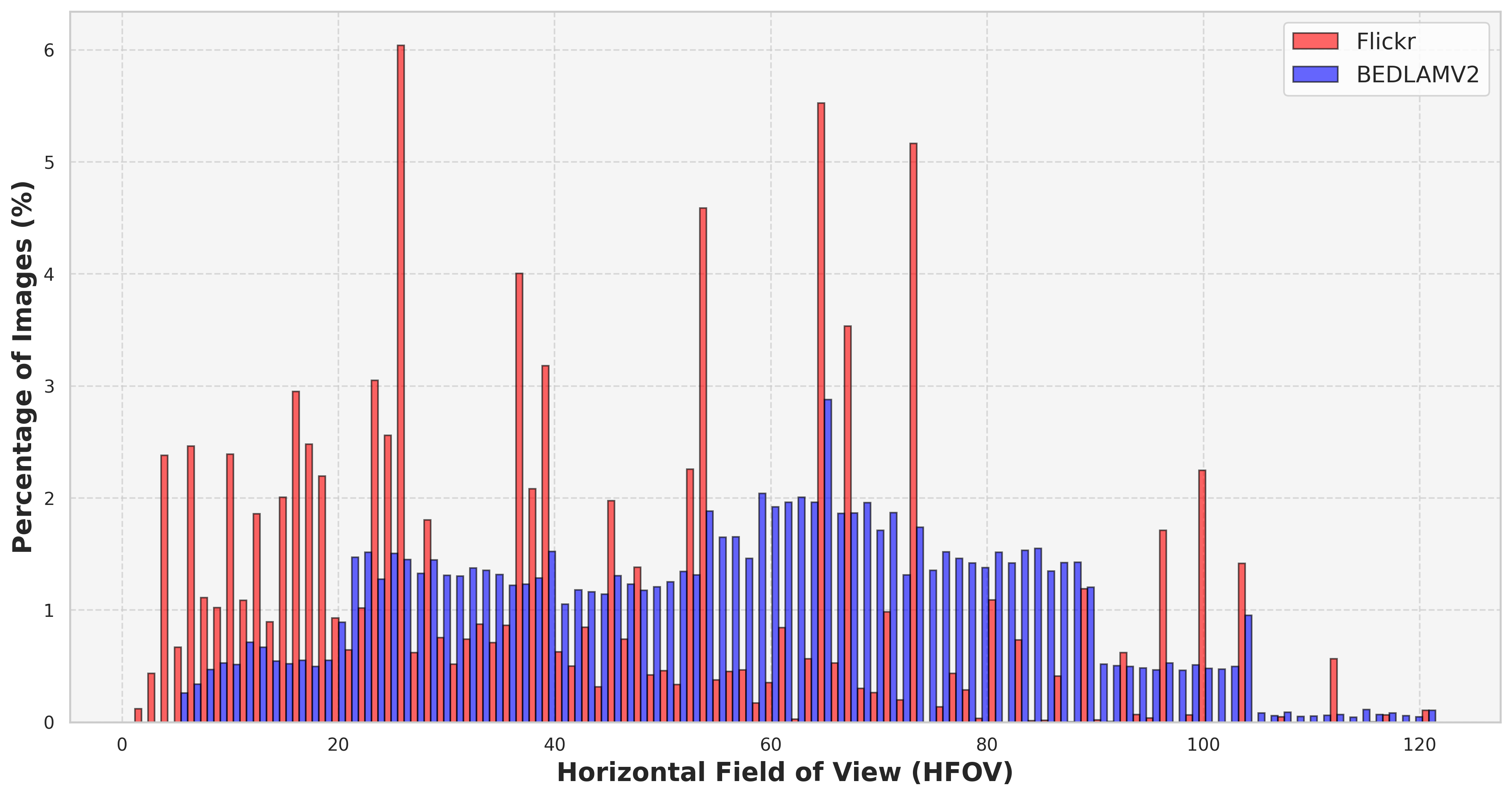}
    \includegraphics[height=2.2in]{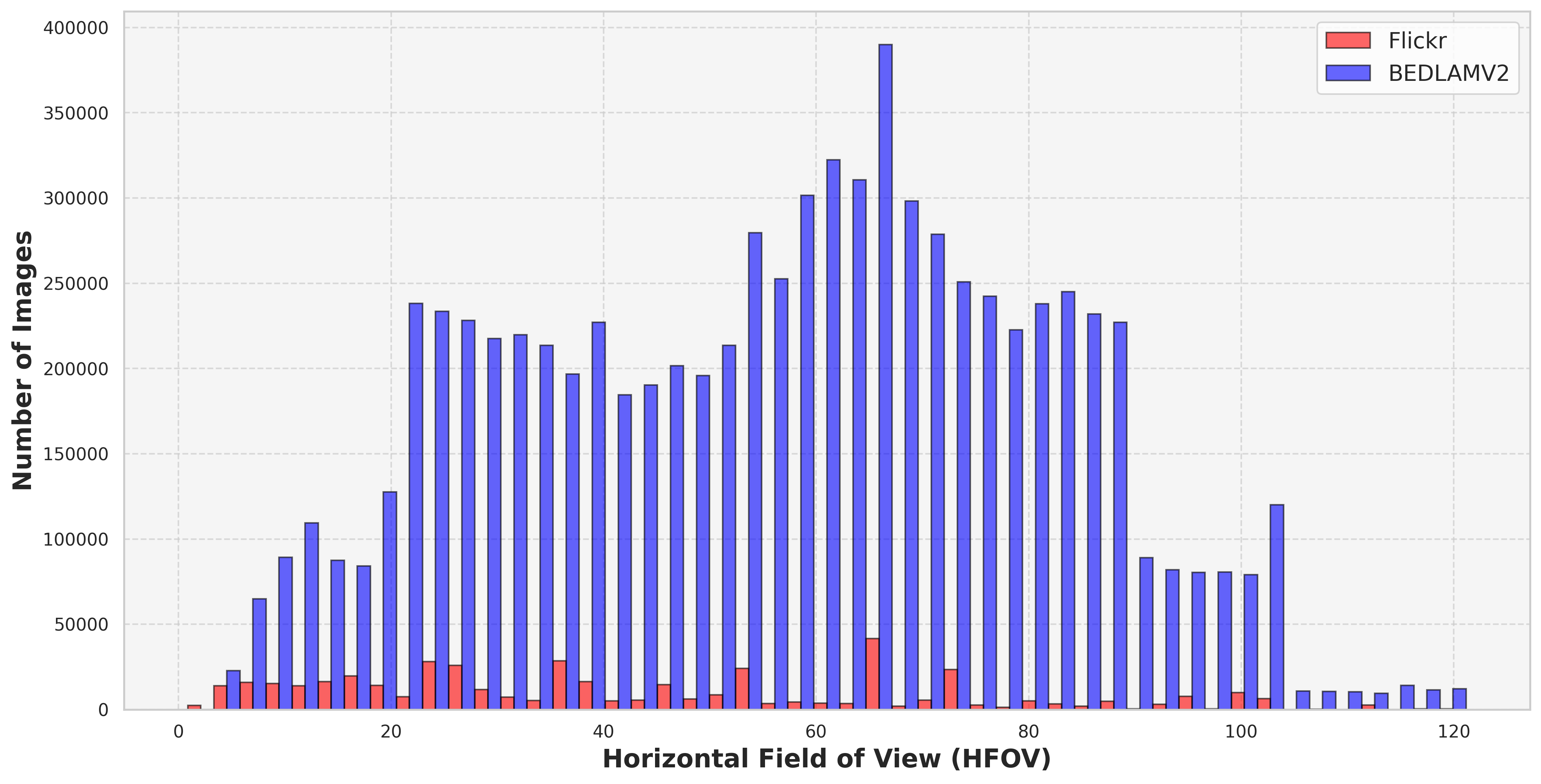}
    \caption{HFOV comparison between \btwo and real Flickr data from \cite{CameraHMR2025}.
    (top} percentage of HFOVs.  (bottom) absolute number of HFOVs.
    \label{fig:hfov_plot_comp}
\end{figure}

In addition to synthetic datasets, we compare the distribution of camera intrinsics in \btwo against real-world images from the Flickr HumanFOV dataset introduced in CameraHR \cite{CameraHMR2025}.
To minimize the domain gap between synthetic and real world cameras, \btwo is designed to cover the space of FOVs observed in the Flickr dataset (see Fig.~\ref{fig:hfov_plot_comp}). While the Flickr HFOV (Horizontal Field of View) values are derived from real EXIF metadata and capture natural camera usage patterns, \btwo samples HFOV values synthetically. 

Note that the Flickr data has distinct ``spikes'' for common focal lengths. 
\btwo is quite different in that it is a video dataset and many sequences include changing focal lengths, i.e.~zooming during the camera motion. This results in a more even distribution of HFOVs.

\subsection{Human motions and bodies}
\begin{figure}[t]
    \centering
    \includegraphics[trim=000mm 000mm 000mm 000mm, clip=False, width=10cm]{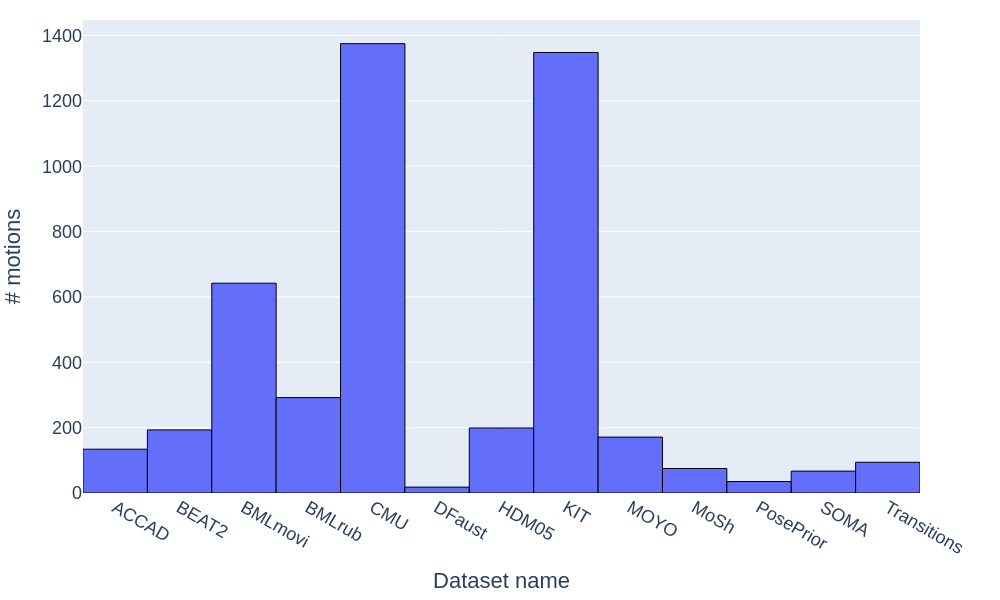}
    \caption{{\bf Motion pool} Number of motions sampled from each dataset, including BEAT2, MOYO and a subset of the datasets in AMASS. }
    \label{fig:motion_usage_all}
\end{figure}

To construct our pool of moving bodies, we repeatedly sample bodies from a set of 1,615 body shapes and assign them a motion from our set of 4,643 motions, whose composition is shown in Figure \ref{fig:motion_usage_all}. As mentioned in subsection \ref{subsection_clothingsim}, we do not run clothing simulation for the 171 motions in the MOYO dataset. We obtain a total of 10,592 motions with clothing simulation that are used multiple times across different render sequences with varied clothing textures.

We use the SMPL-X neutral body with the locked head -- no hair bun and 16 $\beta$ shape parameters.
This differs from the version in BEDLAM.
The original BEDLAM uses a version of SMPL-X without the locked head, which can produce a hair ``bun''. 
This bun produces a bump at the back of the head with female body shapes, and this makes it difficult to simulate strand-based hair realistically; for hair simulation, we need a proper scalp shape.
Moreover, the bodies in BEDLAM are represented with only 10 $\beta$ shape parameters.

We provide a version of the original BEDLAM ground truth in the format of B2, making it easy for people to train HPS models using a combination of B1 and B2 data.

\paragraph{Train-test split. } We reserve a holdout test set of body shapes and motions. 
The test set is composed by 161 body shapes (10\% of the total) and 597 motions (13\% of the total).
Note that these bodies and motions may appear in the training set images, but their ground truth SMPL-X parameters are not provided. This prevents people from training on these characters.

Note also that the motions come from AMASS so, theoretically, people could exploit this to gain an advantage in accuracy. 
However, since we retarget these motions to new body shapes, they are not identical to the AMASS motions.

\begin{figure}[t]
    \centering
    \includegraphics[height=1.6in]{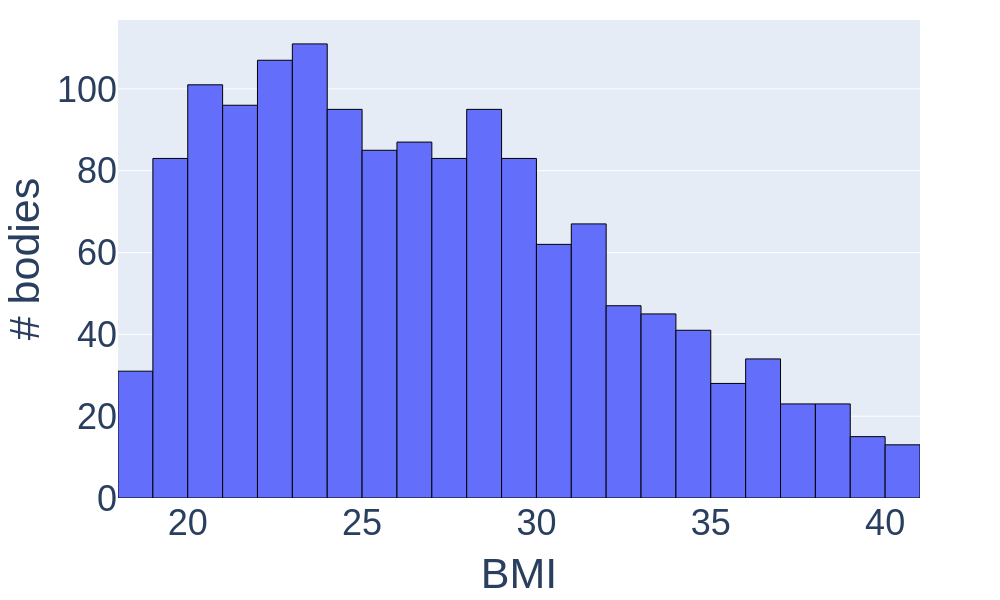}
    \includegraphics[height=1.6in]{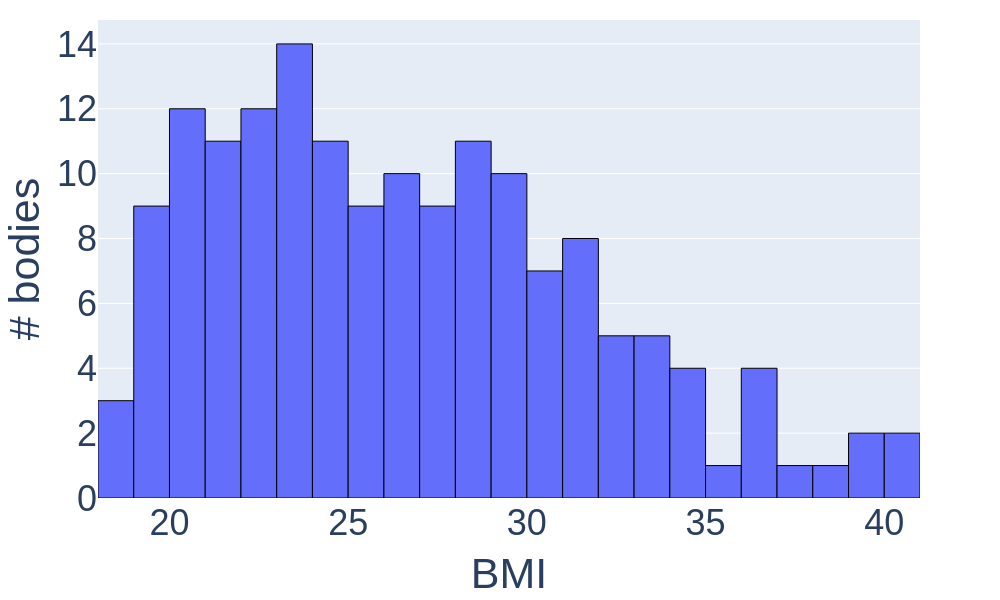}
    \\
    \includegraphics[height=1.6in]{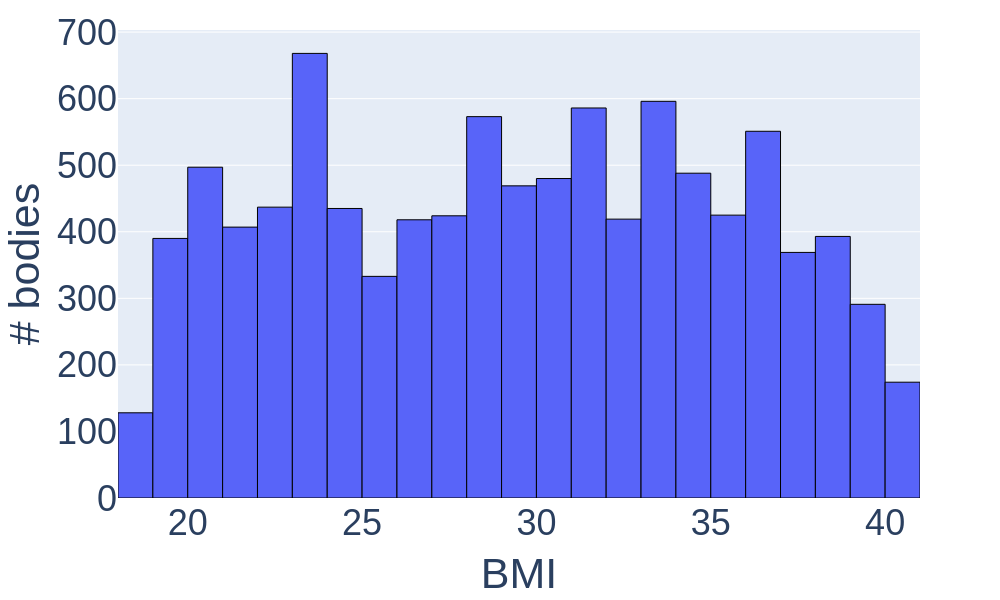}
    \includegraphics[height=1.6in]{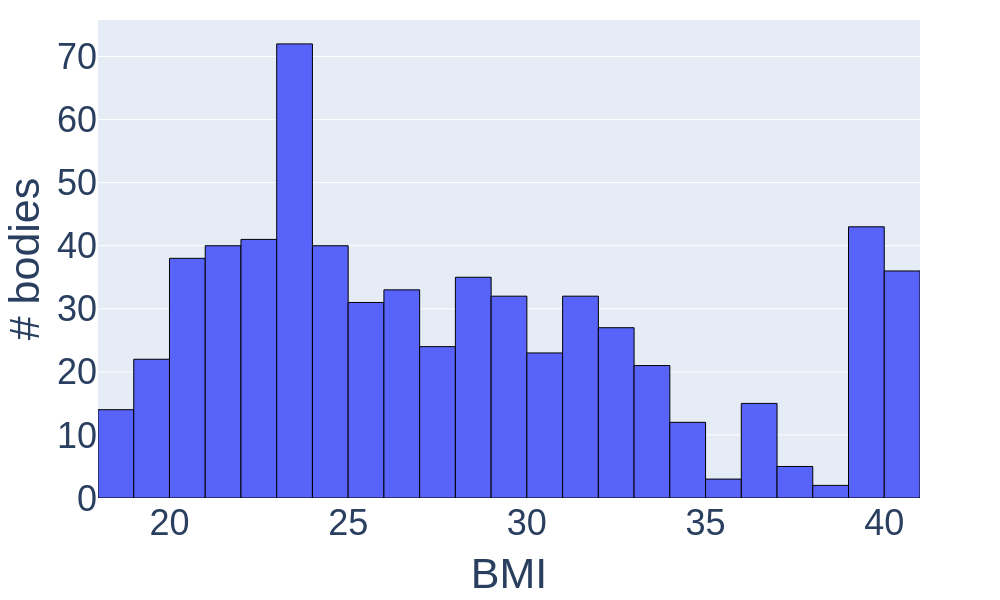}
    \caption{\textbf{Train-test body split.} Left: BMI distribution of body samples in the training set before (top) and after (bottom) oversampling high-BMI examples to balance the dataset. Right: Corresponding BMI distribution in the test set, shown for comparison.}
    \label{fig:bmi_split}
\end{figure}

\begin{figure}[t]
    \centering
    \includegraphics[height=1.6in]{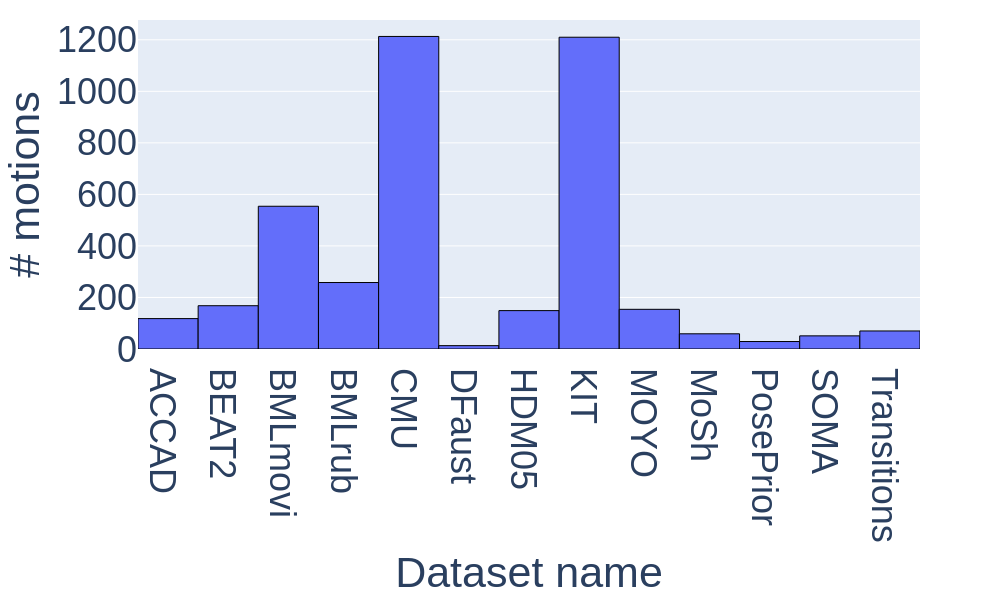}
    \includegraphics[height=1.6in]{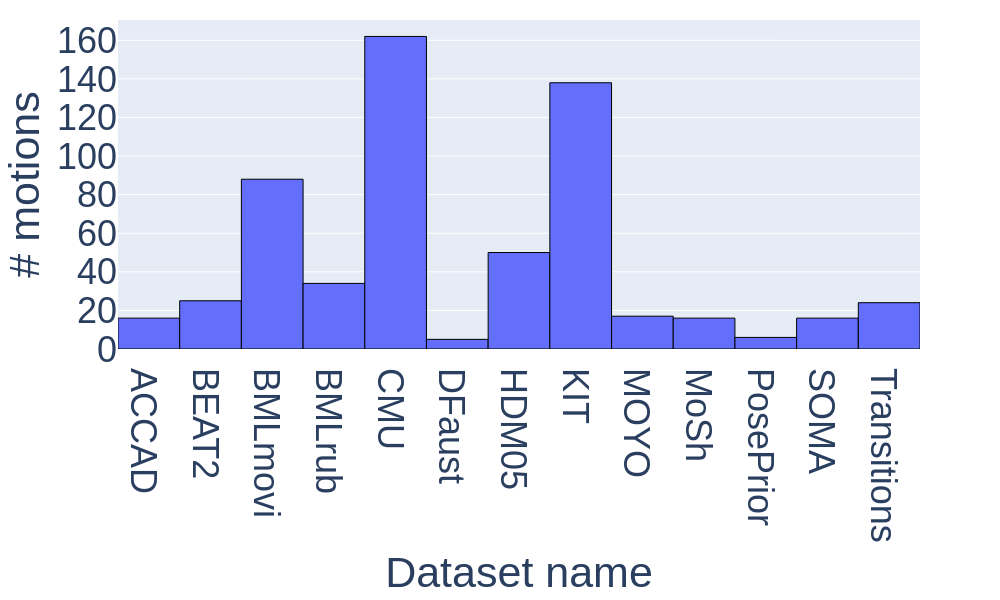}
    \caption{\textbf{Train-test motion split.} Left: number of motions from each motion dataset in the training set. Right: Corresponding numbers of motions sampled for the test set.}
    \label{fig:motion_usage_split}
\end{figure}

The BMI distributions for the training and test bodies can be seen in Figure \ref{fig:bmi_split}, while the numbers of motions sampled from each dataset is shown in Figure \ref{fig:motion_usage_split}. The number of unique motion sequences with clothing simulation that are part of the test set amounts to 641 (6.61\% of the total); these are used to generate 1,824 new render sequences across five novel environments, featuring only test bodies and motions, resulting in a total of 449,061 test images.

\begin{figure}[t]
    \centering
    \includegraphics[height=1.5in]{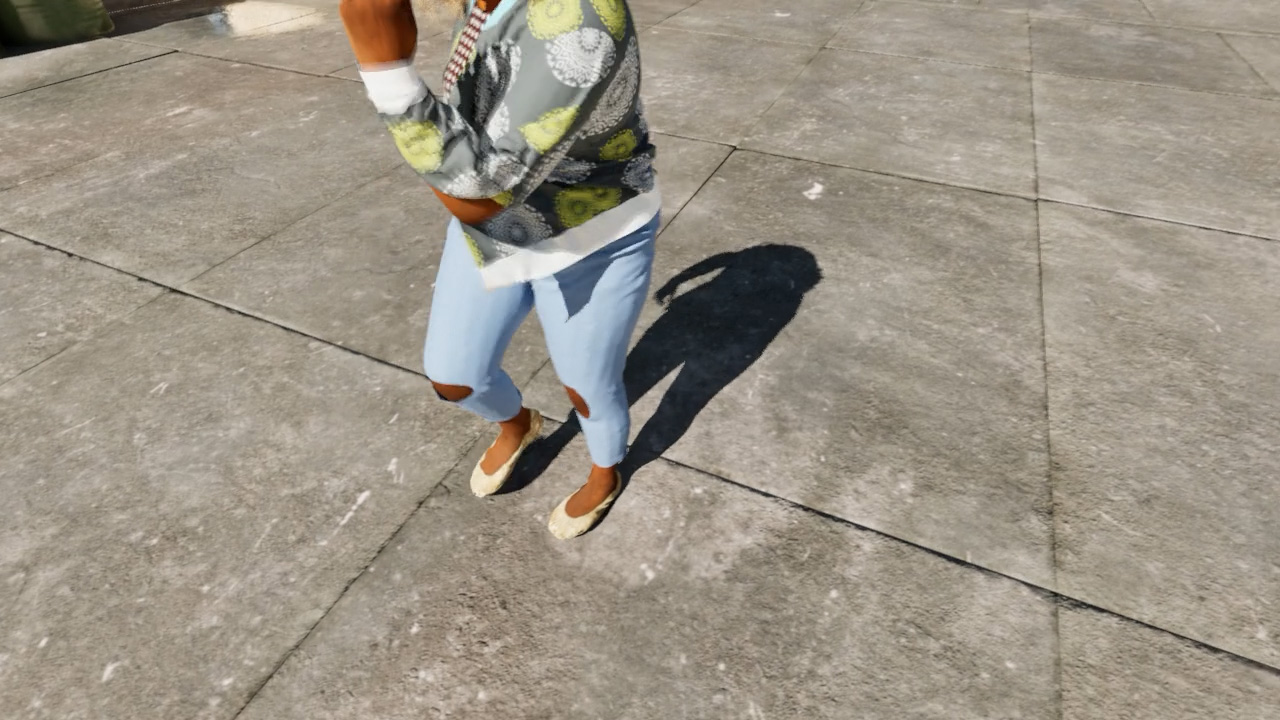}
    \includegraphics[height=1.5in]{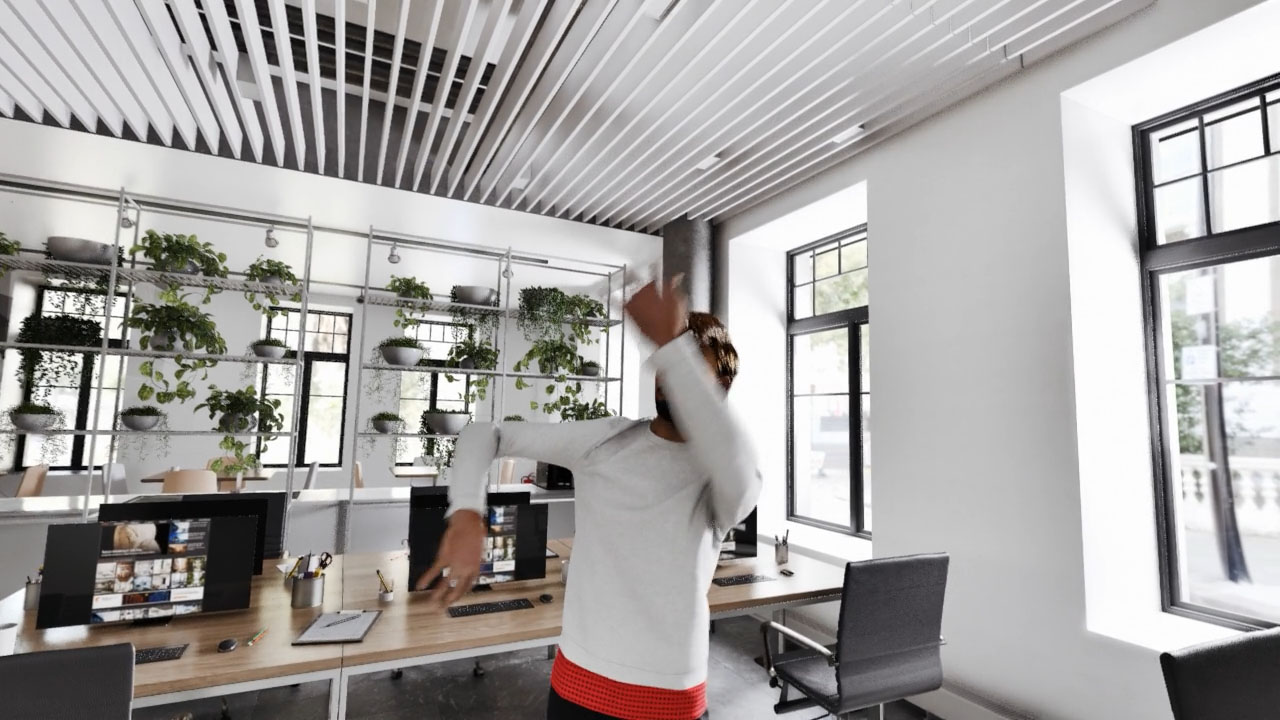}
    \\
    \includegraphics[height=1.5in]{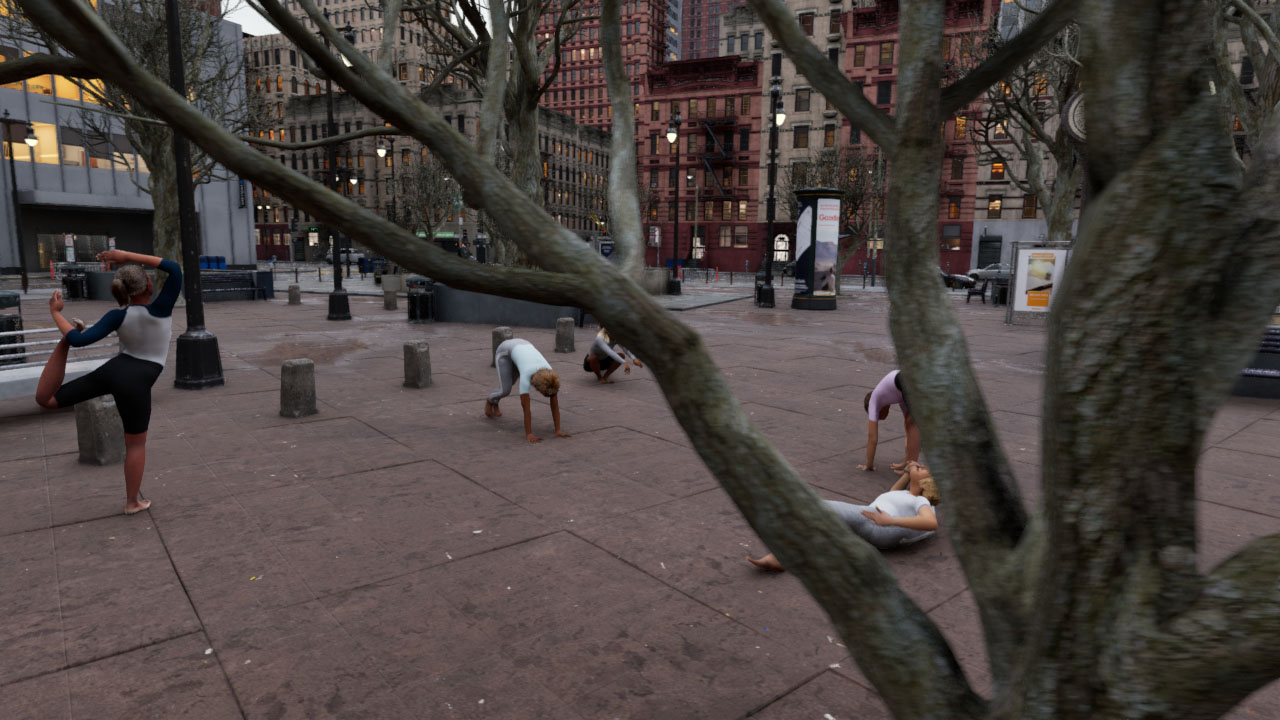}
    \includegraphics[height=1.5in]{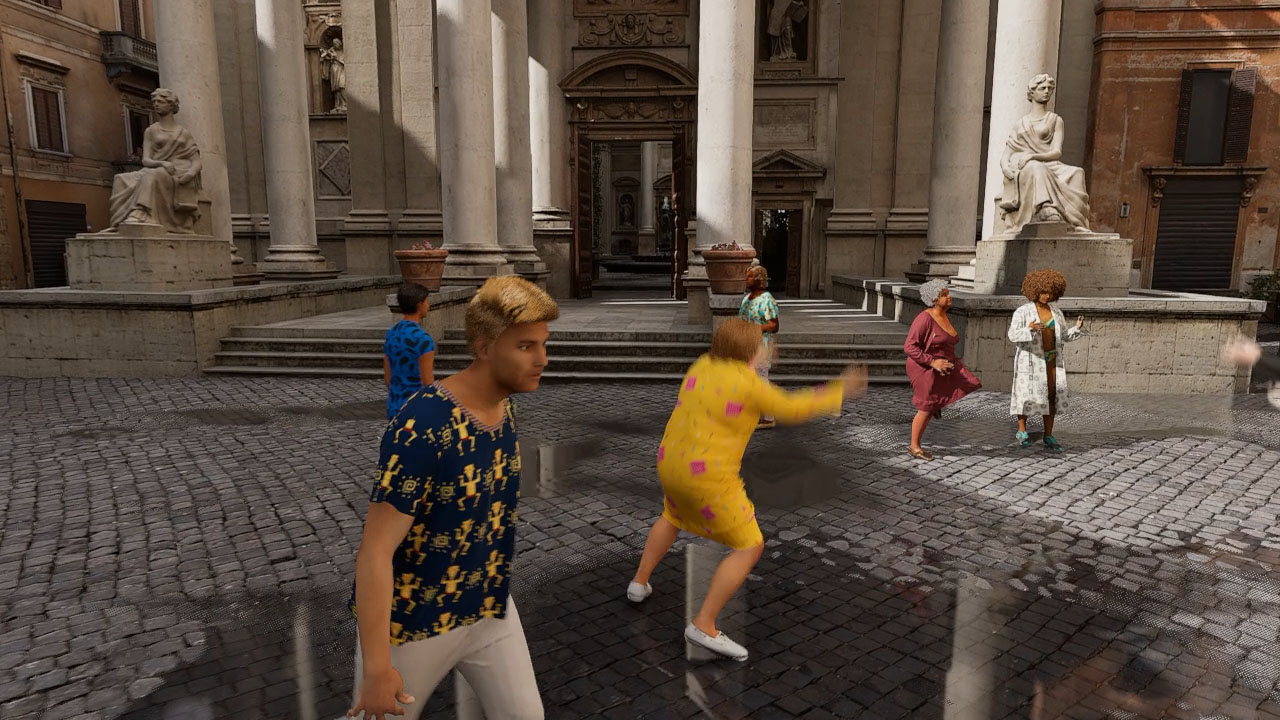}
    \caption{\textbf{Body occlusion examples.} Top row: camera frame occlusion, top right: self occlusion, bottom left: scene occlusion, bottom right: person-person occlusion. All images also show occlusion caused by clothing.}
    \label{fig:occlusion}
\end{figure}

\subsection{Body occlusion}
Our dataset covers the following common occlusion phenomena frequently observed in real-world scenarios: frame occlusion, scene occlusion, self occlusion, person-person occlusion and occlusion caused by 3D clothing. See Figure \ref{fig:occlusion} for some examples.

\begin{figure}[t]
    \centering
    \includegraphics[height=1.5in]{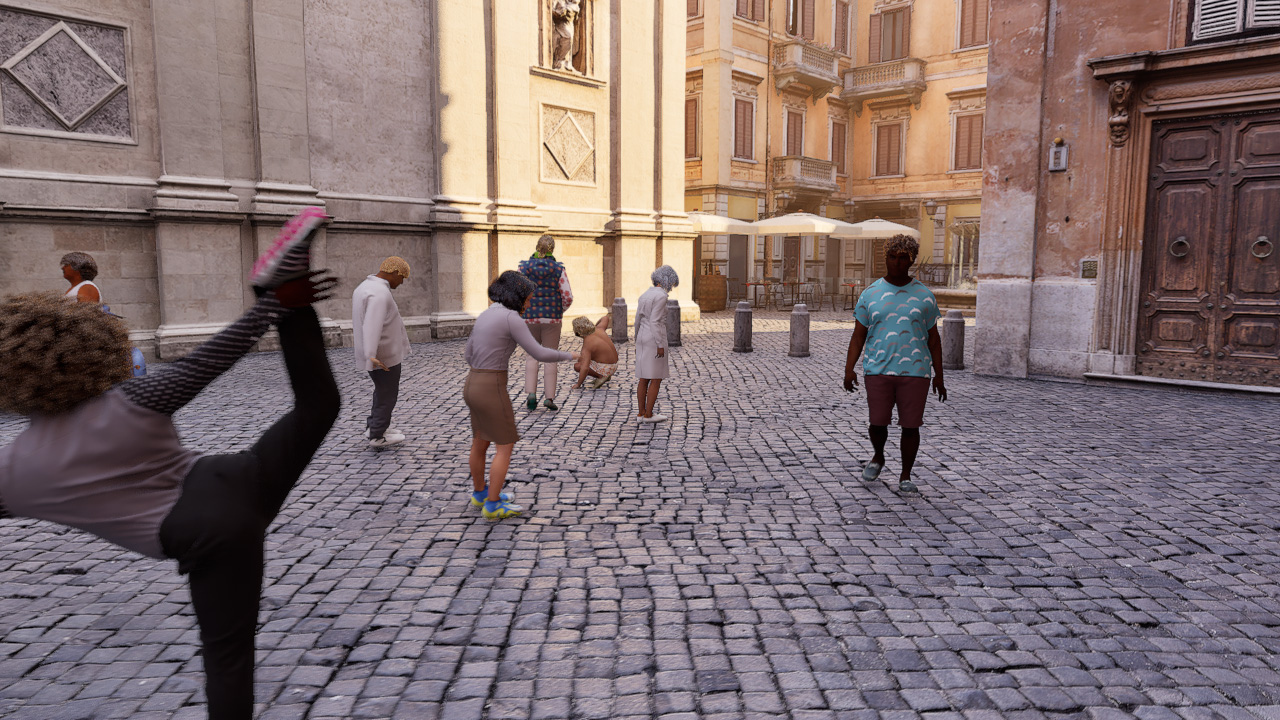}
    \includegraphics[height=1.5in]{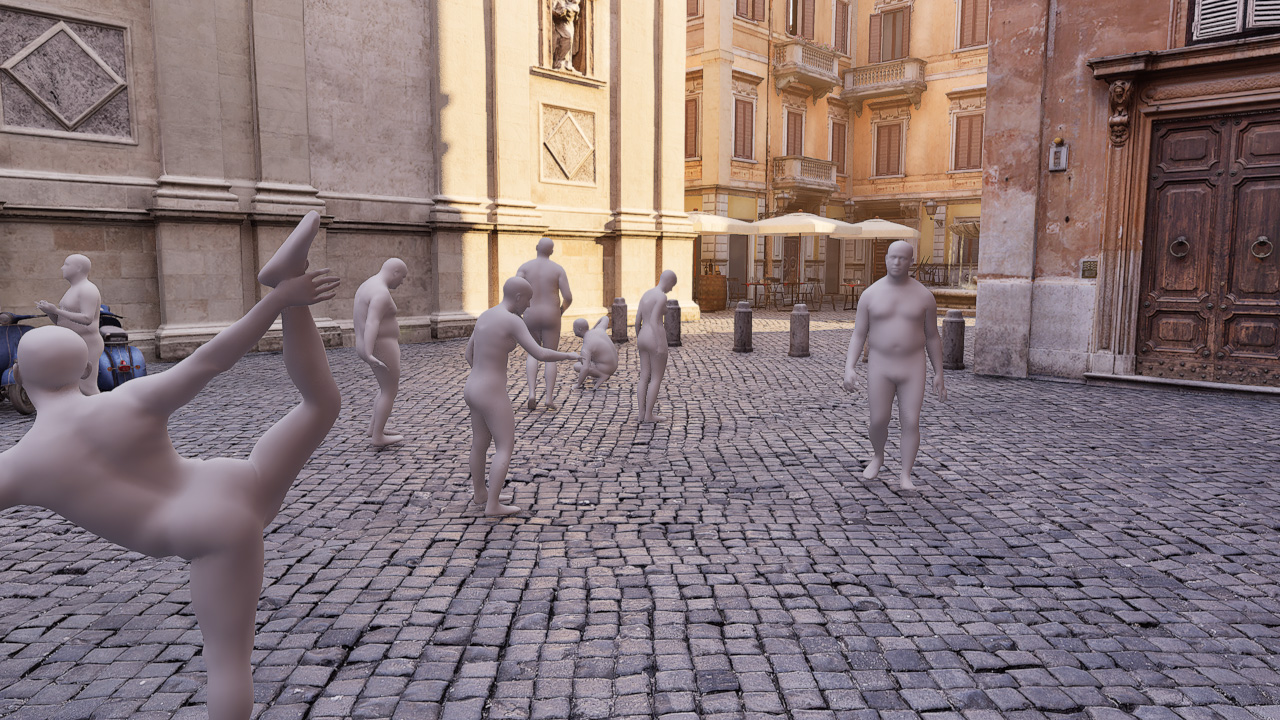}
    \\
    \includegraphics[height=1.5in]{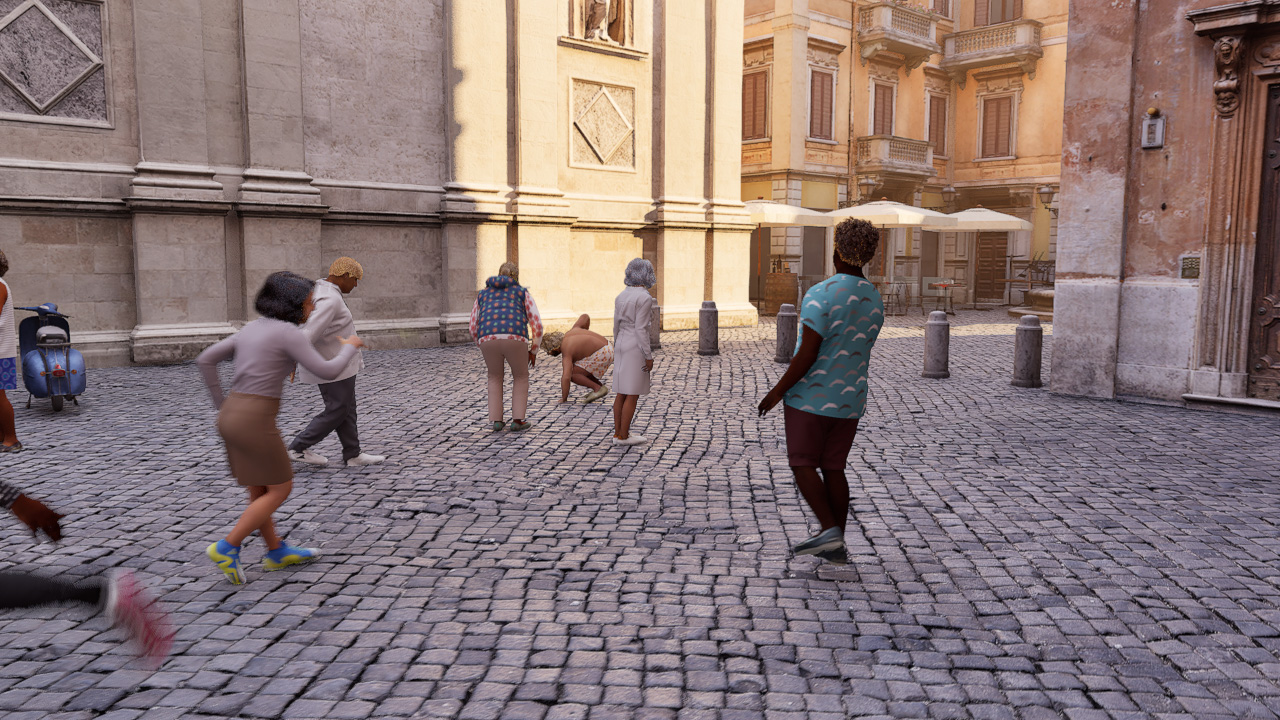}
    \includegraphics[height=1.5in]{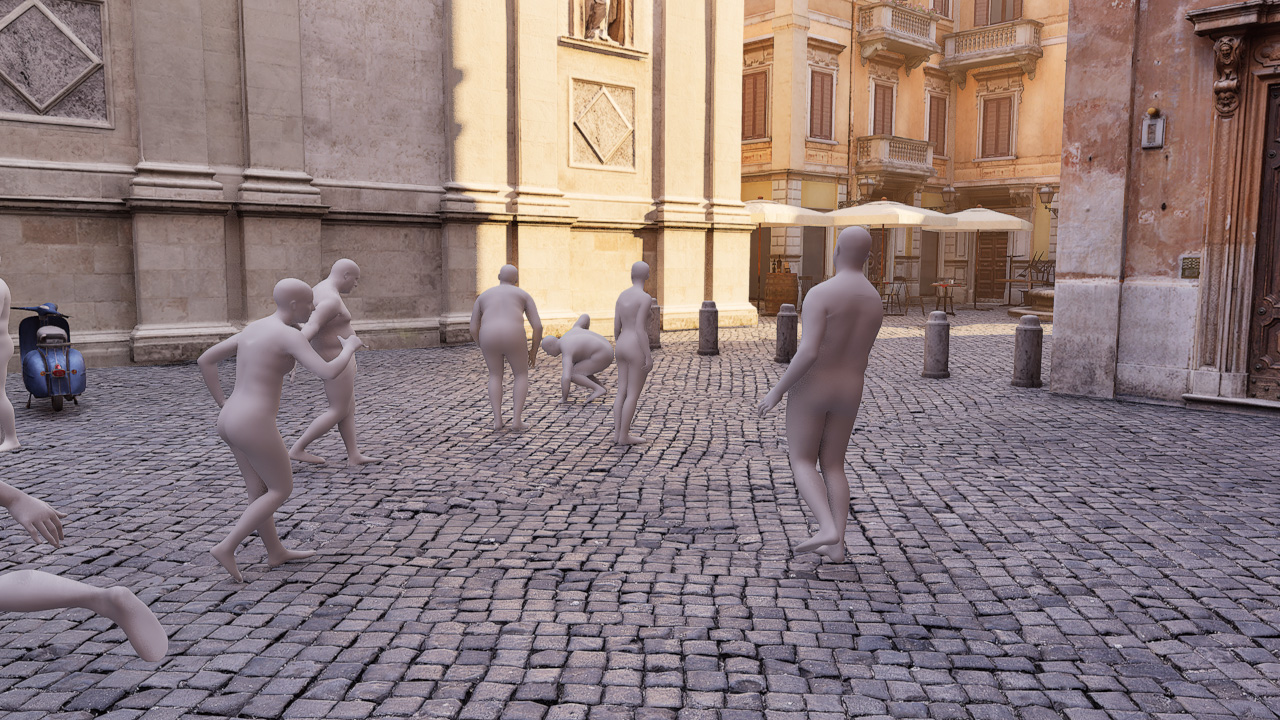}
    \caption{\textbf{Ground truth bodies.} Example frames from the dataset along with the ground truth bodies projected into the frame.}
    \label{fig:datasetsamples}
\end{figure}

\subsection{Visualizing the dataset}

Figure \ref{fig:datasetsamples} shows a few example frames from the dataset together with ground truth bodies. 

The \textbf{Supplemental Video} provides a quick overview of all the dataset components and camera motions: \url{https://youtu.be/ylyqHnwhpsY}.

Besides PNG image sequences, we also provide all 27480 dataset video sequences in much smaller MP4 video format on our project website (\url{https://bedlam2.is.tuebingen.mpg.de/}). Small dataset samples for various camera motion types are also available for download.

\subsection{Experiments}
\paragraph{Shape Evaluation of image-based methods.}
We train the image-based human pose and shape estimation method CameraHMR \cite{CameraHMR2025} on the B1, B2, and B1+B2 datasets. Since standard image-based human pose and shape (HPS) benchmarks exhibit limited shape variability, we use the B2 test set to evaluate shape accuracy. To isolate shape error from pose error, we report results using the PVE-T-SC metric \cite{STRAPS2020BMVC}, which computes the Per-Vertex Error (PVE) after bringing the predicted and ground-truth meshes to a T-pose and performing scale alignment. As shown in Table~\ref{tab:image_shape_accuracy}, training on \btwo reduces the PVE-T-SC error by approximately 20\%, demonstrating a substantial improvement in shape estimation accuracy.
\begin{table}[t]
\centering
\caption{{Shape accuracy comparison using PVE-T-SC.}}
\vspace{0.5em}
\begin{tabular}{lcc}
\toprule
{Model} & {PVE-T-SC} $\downarrow$ & {Improvement (\%)} \\
\midrule
BEDLAM1   & 8.85 & -- \\
BEDLAM2             & \textbf{7.20} & 18.6 \\
BEDLAM1+2   & 7.44 & 15.9 \\
\bottomrule
\end{tabular}
\label{tab:image_shape_accuracy}
\end{table}

\paragraph{Training of video-based methods.}
We train GVHMR~\cite{shen2024gvhmr} and PromptHMR~\cite{promptHMR} using only the AMASS~\cite{AMASS}, B1, and B2 datasets during the video training phase. Both models are trained for 500 epochs, and we report results from the final checkpoint. To improve global motion stability, we apply foot skating post-processing to both methods, which leverages foot contact predictions to reduce unrealistic sliding artifacts.

\paragraph{Image-space evaluation of video-based methods.}
We use B1 and B2 to train two of the recent SOTA methods that estimate human motion from video, GVHMR \cite{shen2024gvhmr} and PromptHMR \cite{promptHMR}.
Table~\ref{tab:video_cam} shows the results using the camera-space metrics (PA-MPJPE, MPJPE, and PVE). 
These errors are lower than for the single-image method except for PA-MPJPE on RICH.
B2 or a combination B1 and B2 provides the best results for all but the MPJPE and PVE on the RICH dataset where B1 is best.

\paragraph{Evaluation of video-based methods on \btwo test set.}

We evaluate GVHMR\cite{shen2024gvhmr} and PromptHMR~\cite{promptHMR} on the B2 test set. The world coordinate metrics show that B2 is more challenging than existing benchmarks like EMDB~\cite{emdb} and RICH~\cite{Huang:CVPR:2022}.

See the main paper, Table 2, for the evaluation of the video-based methods on real video benchmarks.

A key property of the B2 test set is varying focal lengths within sequences; e.g.~dolly zoom. 
This creates difficulties for SLAM-based approaches, which typically assume static camera intrinsics. PromptHMR~\cite{promptHMR} relies on a metric SLAM method~\cite{teed2021droid,hu2024metric3dv2} to transform human motion from camera to world coordinates. Therefore, PromptHMR fails more often than GVHMR on sequences with focal length variations. GVHMR is more robust because it only uses the angular velocity of camera motion, making it less sensitive to focal length changes.
This highlights that \btwo is sufficiently challenging to drive the field to develop robust methods that cope with natural camera movements.

\subsection{Computational costs}

\paragraph{Rendering.} 
To render the entire training dataset, it took approximately 467 hours (383 hours for the image (PNG) files and 84 hours for the depth maps).
We initially rendered with an NVIDIA RTX 3090 GPU (60\%\/ of generated PNG images) and later switched to a new PC with an RTX 4090 GPU (40\%\/ of generated PNG images).
Consequently, the overall time needed to re-render the dataset on a RTX 4090 GPU would be less than reported here.
The observed GPU utilization varied from 30\%\/ to 100\%, with HDRI renders in combination with complex hair grooms having the highest GPU utilization and benefiting the most from the RTX 4090 upgrade.

\paragraph{Clothing simulation.}
The clothing simulations were performed using CLO fashion design software on machines equipped with Intel Xeon CPUs operating at frequencies between 2.0 and 2.5 GHz. Each simulation utilized 12 CPU cores, as using a higher number of cores was found to degrade performance. Under this configuration, simulations—ranging from 120 to 480 frames in length—required an average of 0.8 hours to complete. A total of 10,592 simulations were generated for the training and test datasets, resulting in approximately 8,579 CPU hours on 12-core, 2.0–2.5 GHz processors.

\begin{table*}[t]
  \centering 
  \caption{Camera-space evaluation of video-based methods.}
\resizebox{\textwidth}{!}{
  \begin{tabular}{ll|ccc|ccc|ccc}
    \toprule
 &\multirow{2}{*}{Dataset}& \multicolumn{3}{c}{3DPW~\cite{vonMarcard18ECCV}} & \multicolumn{3}{c}{EMDB~\cite{emdb}} & \multicolumn{3}{c}{RICH~\cite{Huang:CVPR:2022}} \\
    \cmidrule(lr){3-5}
    \cmidrule(lr){6-8}
    \cmidrule(lr){9-11}
     && PA-MPJPE $\downarrow$ & MPJPE $\downarrow$ & PVE$\downarrow$ & PA-MPJPE$\downarrow$  &MPJPE$\downarrow$ & PVE$\downarrow$  & PA-MPJPE $\downarrow$ & MPJPE$\downarrow$ & PVE$\downarrow$  \\
     \midrule
         &GVHMR~\cite{shen2024gvhmr} - B1& 39.8 & 60.7 & 73.9 & 45.5 & 75.3 & 87.7 & 42.0 & 70.1 & 78.7 \\
        & GVHMR~\cite{shen2024gvhmr} - B2& 38.4 & 58.6 & 70.4 & 44.0 & 72.2 & 84.1 & 37.9 & 67.4 & 76.5 \\        
         &GVHMR~\cite{shen2024gvhmr} - B1+B2 & 37.6 & \textbf{57.2} & 70.4 & 44.0 & 74.8 & 87.0 & \textbf{37.2} & 65.3 & 74.3 \\
    \midrule
     & PHMR~\cite{promptHMR} - B1& 38.2 & 57.5 & 69.2 & 42.1 & 80.5 & 92.7 & 38.5 & \textbf{62.1} & \textbf{70.4}\\
     & PHMR~\cite{promptHMR} - B2& \textbf{37.0} & 57.3& \textbf{67.5} & \textbf{40.2} & 73.5 & 84.2 & 37.8 & 69.0 & 78.4 \\
     & PHMR~\cite{promptHMR} - B1+B2& 37.2 & 57.6 & 68.3 & \textbf{40.2} & \textbf{69.2} & \textbf{80.2} & 37.3 & 66.2 & 75.2 \\
     \bottomrule
  \end{tabular}
  }
\label{tab:video_cam}
\end{table*}

\begin{table*}[t]
    \centering
    \caption{World and camera space evaluation of video-based methods on B2 test set.}
    \resizebox{0.8\textwidth}{!}{
        
    \begin{tabular}{l|cccccccc}
        \cmidrule[0.75pt]{1-9} & \multicolumn{8}{c}{BEDLAM2}  \\
        \cmidrule(lr){2-9}
        
        Models & PA-MPJPE & MPJPE & PVE & WA-MPJPE & W-MPJPE  & RTE & Jitter & Foot-Sliding \\
        \cmidrule{1-9}
        GVHMR~\cite{shen2024gvhmr} - B1 & 57.2 & 98.6 & 83.3 & 240.6 & 502.2 & 6.3 & 16.1 & 3.0  \\
        GVHMR~\cite{shen2024gvhmr} - B2 & 38.8 & 66.1 & 55.7 & 203.9 & 444.1 & 5.3 & 16.0 & 2.5 \\
        GVHMR~\cite{shen2024gvhmr} - B1 + B2 & 36.4 & 62.0 & 52.0 & \textbf{195.3} & \textbf{440.3} & \textbf{4.8} & \textbf{14.1} & \textbf{2.4} \\
        \cmidrule[0.75pt]{1-9}
        PHMR~\cite{promptHMR} - B1 & 35.1 & 63.0 & 54.7 & 220.8 & 781.6 & 6.1 & 16.1 & 3.0 \\
        PHMR~\cite{promptHMR} - B2 & 33.5 & 57.2 & 48.6 & 230.3 & 801.5 & 6.4 & 16.4 & 3.3 \\
        PHMR~\cite{promptHMR} - B1 + B2 & \textbf{32.0} & \textbf{55.7} & \textbf{47.5} & 223.2 & 781.2 & 6.0 & 16.0 & 3.0 \\
        
        \bottomrule
    \end{tabular}
    }
    \label{tab:video_world_b2}
\end{table*}

\begin{table*}[t]
	\centering
        	\caption{Third-party assets used for rendering \btwo. Most 3D environments were purchased from Unreal Marketplace and its successor fab.com. Evermotion indoor assets were purchased directly from the vendor as a bundle. Please check individual vendor licenses for further details on Generative AI usage permissions.}
	\resizebox{\linewidth}{!}{%
	\begin{tabular}{l|l|l}
        \toprule
		Asset Type & Name & Source \\
		\midrule
		Body Textures & Bald Head Versions & Meshcapade GmbH, https://meshcapade.com, CC BY-NC 4.0\\
        Clothing Textures & B1 WowPatterns & \bedlam, https://bedlam.is.tuebingen.mpg.de/\\
		Environment - HDRI & Various HDRIs & Poly Haven, https://polyhaven.com/hdris, CC0 Public Domain\\
		Environment - 3D & ai0805 &  Archinteriors for UE, https://evermotion.org/\\
		Environment - 3D & ai0901 &  Archinteriors for UE, https://evermotion.org/\\
		Environment - 3D & ai1004 &  Archinteriors for UE, https://evermotion.org/\\
		Environment - 3D & ai1101 &  Archinteriors for UE, https://evermotion.org/\\
		Environment - 3D & ai1102 &  Archinteriors for UE, https://evermotion.org/\\
		Environment - 3D & ai1105 &  Archinteriors for UE, https://evermotion.org/\\
		Environment - 3D & archmodelsvol8 &  https://www.fab.com/listings/910a05ca-4f7a-4aac-9c1b-c0bf7aabfbd8\\
		Environment - 3D & busstation &  https://www.fab.com/listings/55c97991-d732-4f63-a831-d38843fb5fb0\\
		Environment - 3D & chemicalplant &  https://www.fab.com/listings/a70632d1-f2d2-4b4d-a621-0dc5c3b259fd\\
		Environment - 3D & citysample &  https://www.fab.com/listings/4898e707-7855-404b-af0e-a505ee690e68\\
        Environment - 3D & middleeast &  https://www.fab.com/listings/b46926e1-fe3c-4c20-83f0-8be8ee5e8de5\\        
        Environment - 3D & rome &  https://www.fab.com/listings/12ccee26-1515-4ae9-80a2-cd6402346447\\        
        Environment - 3D & stadium &  https://www.fab.com/listings/d7cfc283-bf41-46a0-b7cd-789476c3263d\\        
        Environment - 3D & yakohama &  https://www.fab.com/listings/b46926e1-fe3c-4c20-83f0-8be8ee5e8de5\\        
        Environment - 3D & yogastudio &  https://www.fab.com/listings/ba2d42c5-c9de-49f4-a681-9d0c8970a670\\        
        \bottomrule
	\end{tabular}}
    	\label{tab:used-assets}
\end{table*}

\subsection{Assets}
\label{sec:assets}

Table \ref{tab:used-assets} describes all third-party assets used in making the dataset.  
Note that some of the 3D environment assets may have a ``no-GenAI'' restriction. 
\btwo does not use GenAI in its creation and is designed to further research on human pose and shape estimation. 
We imagine that there will be other uses (e.g.~generative ones) of the dataset beyond what we designed it for. 
Users are responsible for ensuring that their use case aligns with the asset licensees.

\end{document}